\documentclass[acmsmall]{acmart}

\usepackage{tikz}
\usepackage{multirow}
\usepackage[english=american]{csquotes}
\usepackage{hyperref}
\usepackage{tcolorbox}
\usepackage{subcaption}
\usepackage[ruled, linesnumbered]{algorithm2e}

\AtBeginDocument{%
  \providecommand\BibTeX{{%
    \normalfont B\kern-0.5em{\scshape i\kern-0.25em b}\kern-0.8em\TeX}}}

\setcopyright{acmcopyright}
\copyrightyear{2018}
\acmYear{2018}
\acmDOI{XXXXXXX.XXXXXXX}

\acmJournal{TOSEM}
\acmVolume{0}
\acmNumber{0}
\acmArticle{00}
\acmMonth{0}

\definecolor{codegreen}{rgb}{0,0.6,0}
\definecolor{codepurple}{rgb}{0.8,0.2,1}

\newcommand{\eg}{\textit{e}.\textit{g}., }
\newcommand{\ie}{\textit{i}.\textit{e}., }

\newcommand{\rqone}{How do similarities and properties relate to each other in terms of DNN model types?}
\newcommand{\rqtwo}{Can the introduced similarities be used as a proxy for the defined properties?}
\newcommand{\rqthree}{Can similarities be used to operationalize test transfer?}
\newcommand{\rqfour}{Is transfer with \textit{GIST} more effective than generating test cases from scratch from a trade-off property covered/execution time point of view?}

\begin{document}

\title{\textit{GIST}: Generated Inputs Sets Transferability in Deep Learning}

\author{Florian Tambon}
\email{florian-2.tambon@polymtl.ca}
\orcid{0000-0001-5593-9400}
%\email{webmaster@marysville-ohio.com}
\affiliation{%
  \institution{Polytechnique Montreal}
  %\streetaddress{P.O. Box 1212}
  \city{Montreal}
  \state{Quebec}
  \country{Canada}
}

\author{Foutse Khomh}
\email{foutse.khomh@polymtl.ca}
%\email{webmaster@marysville-ohio.com}
\affiliation{%
  \institution{Polytechnique Montreal}
  %\streetaddress{P.O. Box 1212}
  \city{Montreal}
  \state{Quebec}
  \country{Canada}
}

\author{Giuliano Antoniol}
\email{giuliano.antoniol@polymtl.ca}
%\email{webmaster@marysville-ohio.com}
\affiliation{%
  \institution{Polytechnique Montreal}
  %\streetaddress{P.O. Box 1212}
  \city{Montreal}
  \state{Quebec}
  \country{Canada}
}
\renewcommand{\shortauthors}{Tambon et al.}

\begin{abstract}
\quad To foster the verifiability and testability of Deep Neural Networks (DNN), an increasing number of methods for test case generation techniques are being developed.

When confronted with testing DNN models, the user can apply any existing test generation technique.
However, it needs to do so for each technique and each DNN model under test, which can be expensive. Therefore,
a paradigm shift could benefit this testing process: rather than regenerating the test set
independently for each DNN model under test, we could transfer from existing DNN models.

This paper introduces GIST (Generated Inputs Sets Transferability), a novel approach for the efficient
transfer of test sets. Given a property selected by a user (e.g., neurons covered, faults), GIST enables the
selection of good test sets from the point of view of this property among available test sets. This allows the
user to recover similar properties on the transferred test sets as he would have obtained by generating the test
set from scratch with a test cases generation technique. Experimental results
show that GIST can select effective test sets for the given property to transfer. Moreover, GIST scales better
than reapplying test case generation techniques from scratch on DNN models under test.
%\quad As the demand for verifiability and testability of DNN continues to rise, an increasing number of methods for generating test sets are being developed. However, each of these techniques tends to emphasize specific testing aspects and can be quite time-consuming.

%A straightforward solution to mitigate this issue is to transfer test sets between some benchmarked DNN models and a new DNN model under test, based on a desirable property one wishes to transfer. 

%This paper introduces GIST (Generated Inputs Sets Transferability), a novel approach for the efficient transfer of test sets among Deep Learning DNN models. Given a property selected by a user (e.g., coverage criterion), GIST enables the selection of good test sets from the point of view of this property among available ones from a benchmark.  We empirically evaluate GIST on two properties (fault types and neuron coverage) with two modalities and different test set generation techniques to demonstrate the approach’s feasibility. Experimental results show that GIST can select effective test sets for the given property to transfer it to the DNN model under test.

%Our results suggest that GIST could be applied to transfer other properties and could generalize to different test sets’ generation techniques and modalities. 

\end{abstract}




\received{20 February 2007}
\received[revised]{12 March 2009}
\received[accepted]{5 June 2009}

\maketitle

\section{Introduction}

Deep Neural Networks (DNN) are now part of daily life and are used to tackle problems in different domains such as text \cite{Kowsari19}, image \cite{Voulodimos18}, or sound \cite{Purwins19} among others. The wide usage of DNN models such as Copilot\footnote{\url{https://adoption.microsoft.com/en-us/copilot/}}, ChatGPT\footnote{\url{https://chat.openai.com/}} or Midjourney\footnote{\url{https://www.midjourney.com/home/}} are only the most visible applications of deep learning. Yet, DNN are also prone to errors and misbehaviours that can have more or less dire impacts. As such, as with any software code, DNN programs should be adequately tested to ensure that they behave properly.
    
Improving the DNN's testability has been a growing concern recently along with the efforts to make DNN more transparent and robust, especially when they are to be embedded in safety-critical systems \cite{Tambon22}. To this end, multiple test inputs generation techniques have been introduced to generate test sets assessing different properties of the DNN relying on different metrics or heuristics each with their strengths and weaknesses: among others, we can cite DLFuzz \cite{Guo18} using neuron-coverage guide fuzzing, DeepRoad \cite{Zhang18} which used a GAN based approach, DeepJanus \cite{Riccio20} using the notion of frontier behaviour, DeepEvolution \cite{Braiek19} using metamorphic transformations to generate images or Themis \cite{Galhotra17} a fairness testing technique. Those are only a few examples and there exist several dozen more testing techniques covering a wide array of methods \cite{Braiek20}. This number will likely keep on growing.
This testing process is however limited. First, by design, none of these techniques can be the best in all situations, as they assess different testing aspects using very different criteria. Thus, using multiple techniques together is preferable. Secondly, those generation techniques all require a certain time budget. This budget is dependent on the number of test cases generated and the size of the DNN model under test. Finally, this cost needs to be paid every time a generation technique is applied for a given DNN model under test. This can quickly become time-consuming if multiple DNN models have to be tested. Therefore, we believe that this testing process would benefit from a paradigm change: instead of regenerating the test set independently for each DNN model under test, we could instead leverage transfer from existing DNN models.
\begin{figure}
    \centering    
    \includegraphics[width=\textwidth]{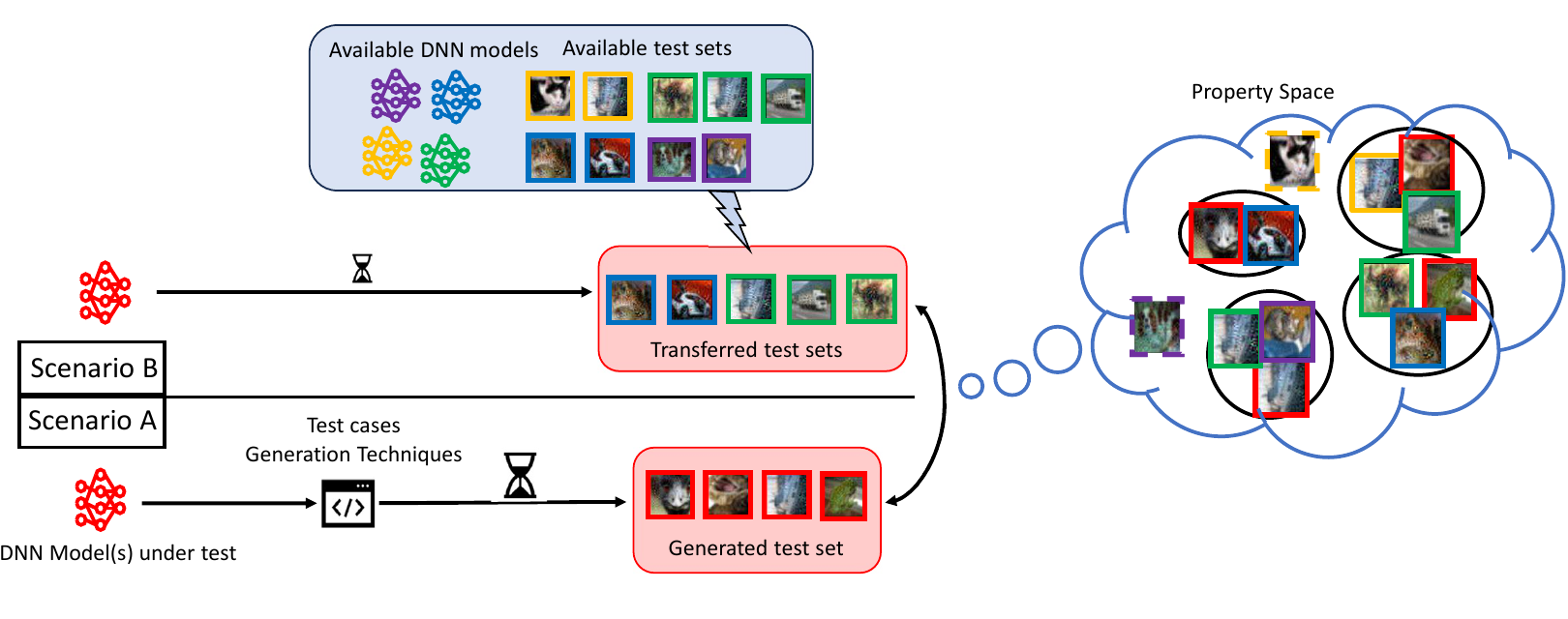}
    \caption{Transferring vs Generating test cases. (\textbf{Scenario A}) When generating the test cases, we apply the test cases generation technique on the DNN model under test which can be time-consuming and must be reapplied for any new DNN model under test and generation technique. (\textbf{Scenario B}) On the contrary, by transferring, we can reuse available test sets to obtain transferred test sets. Those transferred test sets should match what we would have obtained in terms of a desired property, for instance, the types of faults of the DNN model under test. Each black circle in the property space represents a cluster of fault types for the DNN model under test. Data points circled in dashed lines outside of those clusters are not faults for the DNN model under test.}
    \label{fig:intro_image}
\end{figure}

Consider the situation where we want to test one or several DNN models (\textcolor{red}{red} in Figure \ref{fig:intro_image}):
\begin{description}
    \item[\textit{Scenario A:}] The user selects a generation technique to generate relevant test cases for this DNN model (\textcolor{red}{red} images in Figure \ref{fig:intro_image}). After applying directly the test case generation technique, the user would obtain test cases with several properties of interest for assessing the DNN model under test. For instance, it could cover certain types of faults or neurons of the DNN model under test. This would however lead to the aforementioned issues. Notably, every new DNN model under test needs to have the test generation technique reapplied, and so for every test generation technique applied, which is time-consuming.
    \item[\textit{Scenario B:}] The user has access to other DNN models (\textcolor{blue}{blue}, \textcolor{codegreen}{green}, \textcolor{orange}{orange}, \textcolor{codepurple}{violet}) along with their generated test sets. Instead, he could transfer test sets from those DNN models directly. The transferred test sets should match the \textcolor{red}{red} test set in terms of the property of the test set the user is interested in, for example in that case the type of faults. This transfer approach relies on the hypothesis that DNN models and their generated test cases should not be treated independently. Indeed, it is likely that one can leverage existing similarities between DNN models in terms of features and architecture to reuse existing test sets. In Figure \ref{fig:intro_image}, by choosing the \textcolor{codegreen}{green} and \textcolor{blue}{blue} datasets, the user can effectively regenerate a subset of the fault types he would have had generating the \textcolor{red}{red} test set directly. This however required much less time as, instead of generating the test cases, the user would only have to select the most relevant ones. Moreover, this will scale with the number of DNN models the user needs to test as it avoids regenerating test sets every time.
\end{description}

This transfer paradigm begs, however, several questions: How to properly select the relevant test sets for transfer, as the \textcolor{red}{red} test set is not available in practice? How does the similarity between the DNN model under test and the available DNN models relate to test sets covering the relevant property in the property space? Those are the issues that our proposed approach, \textit{GIST} (Generated Inputs Sets Transferability in Deep Learning), is tackling.

\textit{GIST} requires several available DNN models, trained on the same task as the DNN model under test, their generated test sets from which to choose for transferring, as well as a property to transfer (\eg coverage of neurons, types of faults ...). As the property to transfer can not be measured on a new DNN model under test (in the above example, the faults of the DNN model under test exposed by the hypothetical \textcolor{red}{red} test set), \textit{GIST} works by leveraging a proxy to transfer the property of interest. Here, we refer to a proxy as any metric the user deems useful and that should correlate with the property to transfer while not relying on unavailable data during transfer, that is the hypothetical \textcolor{red}{red} test set in our case. \textit{GIST} works in two phases, an \textit{offline} phase for pre-computation and an \textit{online} phase when \textit{GIST} is applied to any number of DNN models under test. During an \textit{offline} phase, \textit{GIST} uses available DNN models (for instance, the \textcolor{blue}{blue}, \textcolor{orange}{orange}, or \textcolor{codegreen}{green} DNN models/test sets in the above example) to validate the proxy. To do so, GIST uses a \enquote{leave one out} procedure on each available DNN model, simulating having a DNN model under test, to verify the correlation between the proxy and the property. This is done only once per test case generation technique. In the \textit{online} phase, \textit{GIST} uses the proxy to transfer the relevant test sets, using only the information of the DNN models under test. \textit{GIST} being agnostic to the couple property/proxy chosen, it is left to the user to select them, though we recommend some initial proxies based on the results of our study.

To demonstrate \textit{GIST} feasibility, we empirically investigate the potential for transferability of test sets using as the properties to transfer a fault type coverage property inspired by recent works of Aghababaeyan et al. \cite{Aghababaeyan23} and a neuron coverage property based on k-Multi Section Neuron Coverage \cite{Ma18}. For the proxy, we leverage existing works on representational/functional similarities \cite{Klabunde23} of DNN. We used different DNN model architectures (for instance, VGG, ResNet ...) and variants (VGG16 or VGG19) for both image and text datasets that we refer to as DNN \textit{model types} in the following. We make available \textit{GIST} replication package \cite{ReplicationPackage}.

%To guide our investigation, we define the following research questions (RQs) divided into two blocks. The first block of RQs is exploratory and aims to investigate test transferability on its own to answer the previous questions we mention, \ie can faults transfer reasonably between DNN models and is there a link between the faults transferred and the DNN model under test/DNN model the transferred test set is from. This will additionally motivate the choice of the representational/functional similarities as a valid proxy, as well as shed some light on the results obtained in the second block of RQs:

%\Florian{To redo RQs}
%\begin{itemize}
%    \item[\textbf{RQ1}] \rqone
%    \item[\textbf{RQ2}] \rqtwo
%\end{itemize}

%The second block of RQs deals with the validation of \textit{GIST} requirements (\eg is the chosen proxy correctly working) as well as evaluating \textit{GIST} practical usage and efficiency, building on previous RQs:
%\begin{itemize}
%    \item[\textbf{RQ3}] \rqthree
%    \item[\textbf{RQ4}] \rqfour
%\end{itemize}

To guide our investigation, we define the following research questions (RQs):
\begin{itemize}
    \item[\textbf{RQ1}] \rqone
    \item[\textbf{RQ2}] \rqtwo
    \item[\textbf{RQ3}] \rqthree
    \item[\textbf{RQ4}] \rqfour
\end{itemize}

We evaluate \textit{GIST} on two datasets: CIFAR10 \cite{Krizhevsky09} for image and Movie Review \cite{Pang05} for text and select some available test input generation methods for each. For the image dataset, the particular techniques were picked both to cover the classification of generation techniques provided by Riccio et al. \cite{Riccio23}, as well as because they were relatively recent and a replication package was available to facilitate the implementation. Similarly, for the text dataset, we used the state-of-the-art method provided by the TextAttack library \cite{Morris20-1}. We first explore the relation between the different similarities and the DNN model types as well as the property and the DNN model type. This allows us to shed light on the link between similarities and the considered properties. Then, we show that some similarities can act as a proxy depending on the property/generation technique by verifying existing correlations. Finally, we show that those proxy similarities can be leveraged to select test sets to transfer for a given property.

This paper makes the following contributions:

\begin{itemize}
    \item We define a general framework \textit{GIST} for test input generated sets transferability.
    \item We analyze the transferability from the point of view of fault types and neuron coverage of three test input generation sets on two datasets of different modalities.  
    \item We show that \textit{GIST} can be used in practice by finding a metric that can select a reasonable test set to transfer on new DNN models under test without the need to reapply the test input generation method on the said DNN model under test.
    \item We provide a replication package with our code and a simple use case notebook to showcase \textit{GIST} idea \cite{ReplicationPackage}.
\end{itemize}

The remainder of the paper is structured as follows. In Section \ref{sec:approach}, we first define the general test set transferability problem, before focusing on the coverage properties and the similarity metrics we will use in the framework. In Section \ref{sec:exp} we detail the DNN models, dataset, and the test input generation technique used in this study. In Section \ref{sec:res} we present our experiments and results. We then discuss the implications and potential improvements of \textit{GIST} in Section \ref{sec:discussion}. Next, we present the related works (Section \ref{sec:related_works}) and threats to the validity of the study (Section \ref{sec:threats}). Finally, we conclude our paper in Section \ref{sec:conclusion}.

\section{Methodology}\label{sec:approach}

\subsection{Test set transferability problem}

%\textcolor{blue}{We first remind the readers of the context of our problem: we would like to apply a generation technique such as a fuzzing method on a new DNN model under test \cite{Guo18, Chen23}. This, however, takes a certain amount of time which is proportional to the DNN model size and number of inputs. This is amplified if we aim to train multiple different DNN models using different architectures, initialization... which would require regenerating the fuzzy set on the new DNN model. Instead, if we already have access to several DNN models (for instance, smaller in size compared to the DNN model under test) along with their test set, we could mitigate this by \textit{transferring} some test sets. That is to say, we could try to select the test sets among the available ones which best reflect the hypothetical test set for a given property, a hypothetical test set we would have otherwise obtained by directly applying the fuzzing process on the DNN model under test.}

Formally, we suppose we have access to a set of $n$ reference DNN models  $\mathcal{R} = \{R_1, ..., R_n\}$, all trained on the same task $\mathcal{D}$, each with generated test sets $\mathcal{T}_R = \{T_{{R_1}}, T_{{R_2}}, ..., T_{{R_n}} \}$. In our case, we consider that all test sets were obtained using the same generation technique $\textbf{G}$. We consider $O$ a DNN model under test, also trained on $\mathcal{D}$, with a test set $T_O$ generated using $\textbf{G}$. Note that this $T_O$ is hypothetical in practice, as it is what we aim to match using only selected test sets from other DNN models. We want to obtain the desirable property (exactly or as close as possible) $\mathcal{P}_O$ that we would have obtained with $T_O$ but only using available one or several $T_{R_i}$ (that we note as $T_R$ in the rest of the paper) from $\mathcal{T}_R$. In other words, the goal of the transfer is to find the test set(s) $T_{R}$ that maximizes $\mathcal{P}_O(T_{R}, T_O)$. That is, the test set(s) that maximize the desirable property $\mathcal{P}_O$ to transfer relative to a hypothetical test set $T_O$ which would have been obtained by applying $\textbf{G}$ on $O$.

This, however, raises an issue: we aim to maximize $\mathcal{P}_O$ which depends on $T_O$ which we do not have access to in practice. What we need is to find a proxy $\mathcal{P}$ which correlates with $\mathcal{P}_O$ without relying on $T_O$. Thus, the study and framework aim to show that it is possible to 1) recover the desirable property with the transferred test sets, 2) there exists a proxy $\mathcal{P}$ respecting the above requirements, 3) leverage the proxy to effectively recover $T_O$ from the perspective of a given desirable property.

\begin{figure}
    \centering
    
\includegraphics[width=\textwidth]{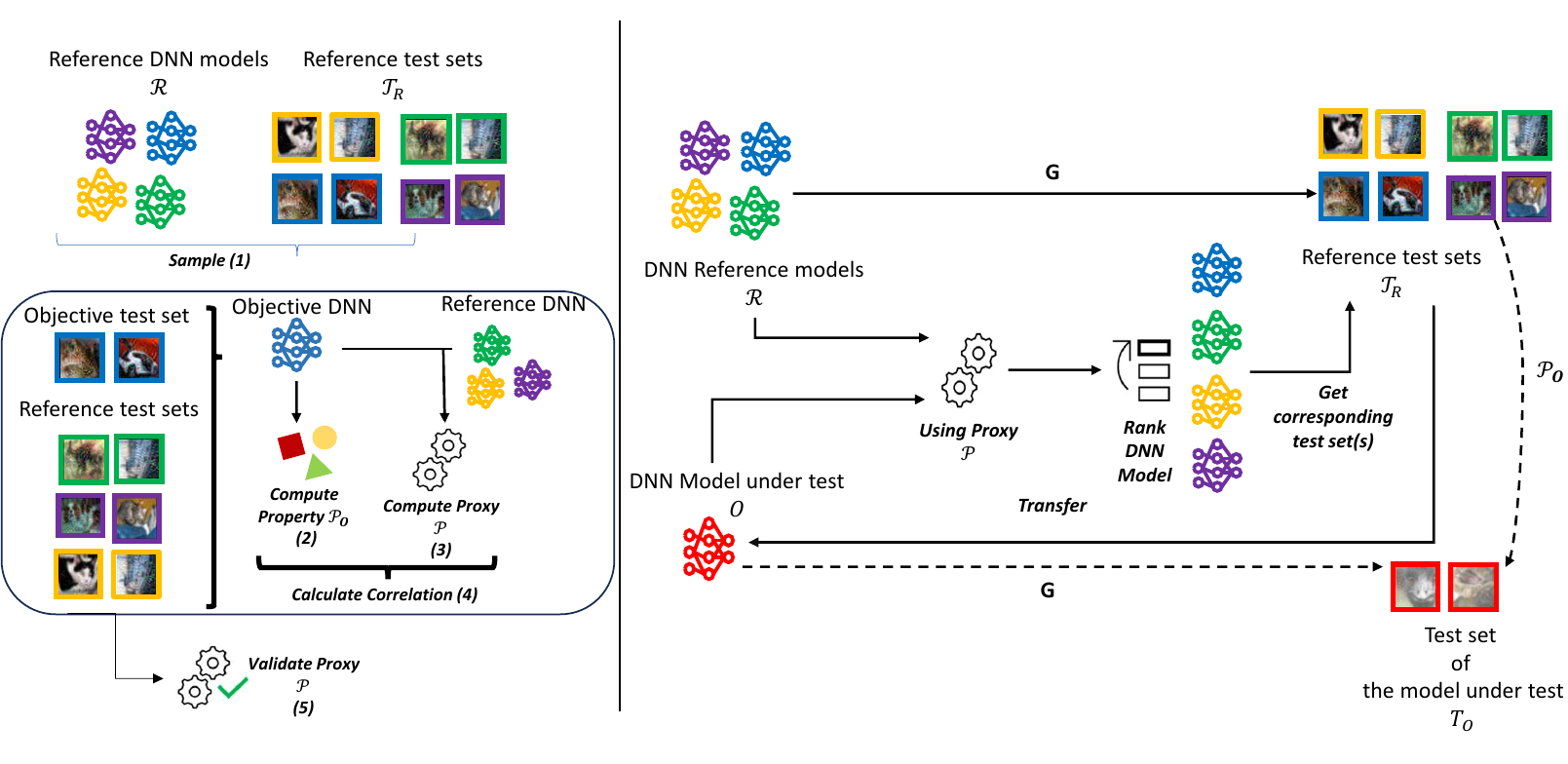}
\caption{General idea of \textit{GIST} for test sets transferability. (\textit{Left}) \textbf{Offline} computation of \textit{GIST} to determine the correct proxy $\mathcal{P}$. (\textit{Right}) \textbf{Online} mode to apply on a new DNN model under test using the proxy.}
\label{fig:gen_framework}
\end{figure}

\subsection{GIST: Generated Inputs Sets Transferability in Deep Learning}

The proposed test set transfer approach \textit{GIST} is composed of two phases: an offline phase during which the proxy relationship is established using available reference DNN models, and the online phase during which the test set transfer occurs on new DNN models under test. An illustration of \textit{GIST} is presented in Figure \ref{fig:gen_framework}. 

During the \textit{offline} phase (Figure \ref{fig:gen_framework} - Left), a property to transfer $\mathcal{P}_O$ is selected by the user based on requirements, and a potential proxy $\mathcal{P}$ is determined. \textit{GIST} is agnostic to any of those. Then, the proxy relation is verified. To do so, using the sets of reference DNN models $\mathcal{R}$ and reference test sets $\mathcal{T}_R$, \textit{GIST} samples, in turn, one of them (in Figure \ref{fig:gen_framework}, on the left \textit{(1)}) to be the pseudo DNN model under test (the \textit{Objective DNN} model in Figure \ref{fig:gen_framework}) and its associated test set (\textit{Objective test set}). \textit{GIST} then computes both the property $\mathcal{P}_O$ \textit{(2)} and the proxy $\mathcal{P}$ \textit{(3)} using the remaining DNN models as references. Note that, in that case, we can compute $\mathcal{P}_O$ since we \enquote{simulate} the fact that we have $T_O$ by substituting a reference DNN model and its test set $T_R$. Finally, \textit{GIST} evaluates the correlation for this objective DNN model between the property $\mathcal{P}_O$ and the proxy $\mathcal{P}$ \textit{(4)}. This is done for all reference DNN models available in $\mathcal{R}$. In the end, if the relation holds (\ie the correlation is deemed strong enough) for a sufficient number of DNN models, $\mathcal{P}$ is deemed to be usable \textit{(5)}. As determining what constitutes a good proxy in this way can be complex, multiple $\mathcal{P}$ can be used simultaneously and we can retain only the best one. This process is done only once for a given generation technique \textbf{G}. We give the pseudo-algorithm of this offline phase in Algorithm \ref{alg:one}.

Then, \textit{GIST} can be used \textit{online} (Figure \ref{fig:gen_framework} - Right) as many times as needed. In that setting, any new DNN model under test $O$ is presented, a DNN model from which we do not have a $T_O$ test set, and we wish to transfer test sets $T_{R}$ from the set $\mathcal{T}_R$ on using the property $\mathcal{P}_O$. Using the proxy $\mathcal{P}$, we can rank the reference DNN models from $\mathcal{R}$ based on their distance to $O$. In the end, the best DNN models, according to $\mathcal{P}$, are selected and their associated test sets are transferred on $O$. What constitutes the \enquote{best} can be something as simple as the closest one to the DNN model under test. Yet, we will see in Section \ref{sec:rq3} that we can leverage the proxy for particular selection heuristics. Since we observed a correlation in the offline phase, we have empirical evidence that the chosen test sets will transfer effectively for the considered property. Note that, in that setting, we indeed only need the trained DNN model under test during the online phase, as everything has been determined previously in the offline phase.  We give the pseudo-algorithm of this online phase in Algorithm \ref{alg:two}.

\begin{algorithm}
\caption{GIST - \textit{Offline} phase}\label{alg:one}
\KwData{Set of reference DNN models $\mathcal{R}$, Set of reference test sets $\mathcal{T}_R$ generated using \textbf{G}, a property to transfer $\mathcal{P}_O$ and proxy to assess $\mathcal{P}$}
\KwResult{Validity of the proxy $\mathcal{P}$}
corr $\gets$ []\;
\For{$R_i \in \mathcal{R}$}{
  obj\_DNN model $\gets R_i$\;
  ref\_DNN models $\gets \mathcal{R} \setminus \{R_i\}$\;
  property $\gets$ []\;
  proxy $\gets$ []\;
  \For{$R_p \in$ ref\_DNN models}{
       property.append($\mathcal{P}_O(T_{R_p}, T_{R_{obj\_DNN model}})$)\;
       proxy.append($\mathcal{P}(R_p, R_{obj\_DNN model}$))\;
  }
  corr.append(Correlation(property, proxy))\;
}
\Return CheckCorrelation(corr)\;
\end{algorithm}

\begin{algorithm}
\caption{GIST - \textit{Online} phase}\label{alg:two}
\KwData{Set of reference DNN models $\mathcal{R}$, DNN Model Under Test $O$, set of reference test sets $\mathcal{T}_R$ generated using \textbf{G}, a property to transfer $\mathcal{P}_O$ and a validated proxy $\mathcal{P}$}
\KwResult{Reference test set(s) to be transferred $\mathcal{T}_{R_{best\_NN models}}$}
proxy $\gets$ []\;
\For{$R_i \in \mathcal{R}$}{
  proxy.append($\mathcal{P}(R_i, O$))\;
}
best\_DNN models $\gets$ ChooseBest(proxy, $\mathcal{R}$)\;
\Return $\mathcal{T}_{R_{best\_NN models}}$\;
\end{algorithm}

\subsection{Property $\mathcal{P}_O$}\label{sec:property}

In our setting, the property $\mathcal{P}_O$ can be any property of the test set $T_O$ we aim to recover by selecting the test sets from $\mathcal{T}_R$. In our experiments, we defined two properties based on two different criteria: neuron coverage and fault-types coverage.

\subsubsection{Neuron Coverage based}

To test DNN, one idea that has been introduced is to quantify how many neurons are activated (\ie activation value is above a threshold) by a set of inputs. This echoes what is done in software testing with basic criteria such as code coverage. Pei et al. \cite{Pei19} introduced neuron Coverage to quantify this aspect. Later, Ma et al. \cite{Ma18} improved neuron Coverage by introducing $k$-Multi Section neuron Coverage (KMNC) as an extension of classical neuron Coverage which aims to assess how many sections of an activation range a test set is covering in a DNN model. Formally, for a neuron $n$, the range $[low_n, high_n]$ of its activation for the train set of the DNN model is divided into $k$ sections $S^n_i$. The $i$-th section is considered covered if for a given input $x$ in a test set $T$ we have $\phi(x, n) \in S^n_i$ where $\phi(x, n)$ is the activation of neuron $n$ with input $x$. We can then define the set of all sections covered by $T_i$ for $n$ as $S^n_{T_i}$.

In our case, we define a property based on this metric as follows: for a given DNN model under test $O$, we measure the activation bands $S^n_{T_O}$ as described above. We then do similarly for each reference test set $T_R$, by measuring the activation bands of the inputs of $T_R$ that lead to faults on the DNN model under test $O$. We obtain $S^n_{T_R}$. Finally, we can define a property $\mathcal{P}_O$ using this criteria and following our definition as:

\begin{equation}\label{eq:m_o_k}
    \mathcal{P}_O = \frac{\sum_{n \in N} \# (S^n_{T_O} \cap S^n_{T_R})}{\sum_{n \in N} \# S^n_{T_O}}
\end{equation}

where $N$ is a pool of neurons to be monitored and $\#$ represents the cardinality symbol. In our case, we will monitor the neurons post-extraction layer. In other words, this property translates to the overlapping fault-inducing activation regions between our hypothetical test set $T_O$ and any test set(s) $T_R$. Similarly to KMNC as a coverage criteria, the idea of the property is to quantify how many sections of the major activation regions of $T_O$ we recover with our chosen test sets.

\subsubsection{Fault-Types Coverage based}

Traditionally in the supervised learning setting, a fault in a DNN happens when, for a given test input of the test set, the DNN model's output is different from the corresponding label of the test input. For instance, the DNN predicts an image to be a cat when it's a dog. While interesting, we might have many images of dogs in our test set that are predicted as cats. In that situation, while our test set exhibits multiple faults, the faults are of the same nature (\ie they are not \textit{diverse}) which would hamper the relevance of the test set. If one were to compare it with traditional software testing, it is not necessarily helpful to cover the same branch in a program using multiple test inputs while a single test input would suffice to reveal the fault.

Aghababaeyan et al. \cite{Aghababaeyan23} introduced an approach to cluster DNN faulty test inputs. They notably showed that the obtained clusters generally correlate with faults and that different clusters correspond to different faults in the DNN model. As such, a test set would be more desirable if it, instead of revealing a higher number of faults, would reveal a higher number of \textit{faults types}\footnote{Note, just like neuron coverage, we are not interested in the particular faults covered (i.e. why is fault 1 covered and not fault 2) but the number that is covered. Nonetheless, in order to keep the same expression as neuron coverage, we will talk about \textit{fault-types} and not the \textit{number of fault-types.}}.

Their approach consists of 4 steps: 1) extracting the features over the data using an extractor and normalizing them, 2) adding two extra features (namely the predicted and original label of the test data), 3) reducing the dimensionality using UMAP \cite{Mcinnes18} and 4) clustering the obtained embedding using HDBSCAN \cite{Mcinnes17}. While they use this generation technique to quantify the fault types in a test set indiscriminately, we instead propose to use it for transferability. Thus, in our case, the property we wish to transfer is simply the number of fault types of $T_O$ that we would find in $T_R$ when using $O$. Note that, by definition, only the inputs from $T_R$ that lead to a fault in $O$ are used in the process. With that definition, $\mathcal{P}_O$ can be defined as the proportion of overlapping faults between $T_R$ and $T_O$ with regard to $T_O$ when using both test sets on $O$. Formally, we define a metric for faults type covered such as:

\begin{equation}\label{eq:m_o_f}
    \mathcal{P}_O = \frac{\# (\mathcal{F}_{T_R} \cap \mathcal{F}_{T_O})}{\# \mathcal{F}_{T_O}}
\end{equation}

Where $\mathcal{F}_T$ is the set of fault types of test set T and $\#$ represents the cardinality symbol.

To fit our approach, we made some changes to their method. First, instead of using a general extractor (in their case VGG19 trained on ImageNet since they only work on images), we will use the DNN model under test itself. This choice is motivated by the fact that we want to directly assess the resulting fault types on the DNN model under test itself, not some third-party DNN model. Moreover, \textit{GIST} should generalize to modalities other than image, where a general extractor such as VGG19 does not exist. Using the DNN model under test itself removes this constraint. Secondly, we make use of all the generated test sets $T_R$ as well as $T_O$ when creating the clusters, instead of the original test and train, as we wish to cluster the fault types on those test sets. The rest of the procedure is as we described above.

We make use of the DBCV \cite{Moulavi14} and Silhouette \cite{Rousseeuw87} scores to manually fine-tune the parameters of the HDBSCAN and UMAP algorithms. To validate our clustering analysis, we will verify that the obtained clusters correspond to individual fault types as shown in Section \ref{sec:valid}.

\subsection{Proxy $\mathcal{P}$}\label{sec:sim}

The last step is to find the proxy $\mathcal{P}$ that we can leverage for $\mathcal{P}_O$, while only using the information that is available to us. In our problem, that information is the training data of the task $D$, the DNN model under test $O$ and the reference DNN models $\mathcal{R}$ with their test sets $\mathcal{T}_R$. As we mentioned before, we do not have access to $T_O$ in practice. 

\begin{figure}[h]
    \centering
    \includegraphics[width=0.5\textwidth]{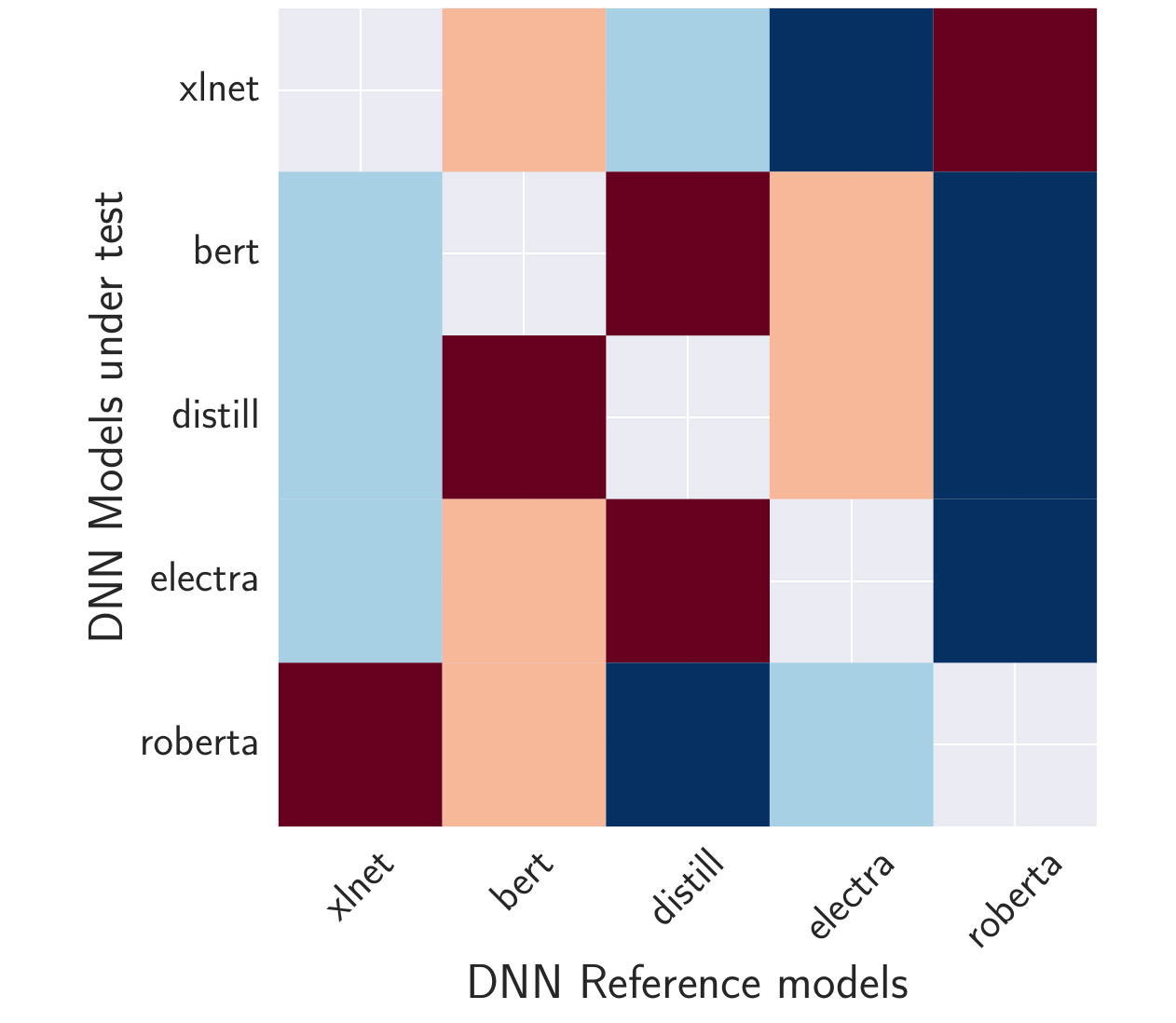}
    \caption{Relative $\mathcal{P}_O$ for the $T_O$ of each DNN model under test when applying different $T_R$ from reference DNN models. The red darker the colour, the higher the relative $\mathcal{P}_O$.}
    \label{fig:why_sim}
\end{figure}

Intuitively, the more two DNN models share similar knowledge, the more they should generalize similarly. Moreover, to some extent, they should generate similar inputs especially if they are trained on the same task $D$ and the same generation technique \textbf{G} is used on those DNN models. To get a sense of the intuition, we conducted a preliminary investigation by calculating for a diverse set of DNN models under test, in the offline setting of \textit{GIST}, the properties $\mathcal{P}_O$ previously mentioned for both criteria. An example of the result is presented in Figure \ref{fig:why_sim}.

In that example, we used NLP-based DNN models with the test sets $\mathcal{T}$ being generated by some NLP generation method. More details can be found later in Section \ref{sec:exp}. For each DNN model under test (and so each $T_O$), there is generally one reference DNN model (and so $T_R$) which maximizes the relative $\mathcal{P}_O$ property, symbolized by the darker red colour. The important observation is that those pairs appear to show some pattern: \textit{Bert} will generally be paired with \textit{Distill}, the architecture of one being derived from the other, or \textit{RoBERTa} is generally paired with \textit{XLNet}. As such, there seem to be some intrinsic patterns based on factors such as architecture or knowledge that seem to have an impact on how much of $T_O$ is recovered through $\mathcal{P}_O$. Previous studies showed that there is a link between such factors and the notion of representational/functional similarities \cite{Boix22, Li21, Klabunde23}.

Thus, we suppose that said similarities could be leveraged as a proxy in our case and so we make the following hypothesis:\\

\textit{The more similar two DNN models are in terms of their \enquote{knowledge}, the more the properties of their generated test set should overlap.} \\

The goal of RQ1 will be to validate this hypothesis. For the sake of describing the methodology, we assume the hypothesis to be true. In our case, as the similarities we describe in the following will be used as the proxy $\mathcal{P}$, we will refer to \textit{similarities} instead of \textit{proxy} when designating $\mathcal{P}$ in our experiments for clarity purposes.

The only part missing is thus to quantify this similarity. In that regard, we can leverage existing representational and/or functional similarities, that is similarities designed to quantify the \enquote{distance in knowledge} between two DNN models. Those similarities are generally split into two categories: \textit{representational} and \textit{functional}. Representational similarities make use of activations of pairs of layers for a fixed set of inputs. Layers in the pair can be of different sizes. On the other hand, functional similarity makes use of the output of both DNN models. Klabunde et al. \cite{Klabunde23} surveyed existing representational and functional similarities. 

While for functional similarities there is no choice concerning the layer taken (the output one), it has some importance for the representational similarities. Existing approaches \cite{Ding21, Kornblith19} mainly used representational similarities with intermediate layers of the extractor part of the DNN models. As we mainly measure similarity between DNN models, we should aim at choosing layers with a close enough functionality across compared DNN models. The only layer with this particular property within the extractor part of the network is the last layer of the extractor, just before the classifier head. Hence, we chose this layer for representational similarities. For the set of inputs, we make use of the \textit{training} data which were used to train both DNN models we want to compare. Using above-mentioned similarities, we can effectively obtain $\mathcal{P}(O, R)$, measuring the \enquote{similarity of knowledge} between the DNN model under test and a reference DNN model from $\mathcal{R}$. In our setting, the similarities can be used as originally described without any additional modification.

There exist different similarities, each quantifying the distance between two given DNN models differently. For instance, not all representational similarities are invariant to the same transformations over the representations (\eg translations, permutations...). Since we do not know which one(s) would relate better to each property $\mathcal{P}_O$, we remained conservative and investigated six of them, three representational and three functional similarities. Those similarities were chosen as they are the most widespread in the literature. We also ensure that each distance belongs to a different similarity type according to Klabunde et al. \cite{Klabunde23}. We briefly explain how each type of similarity is computed and detail each of the selected similarities in the following. We refer the reader to Klabunde et al. \cite{Klabunde23} for detailed explanations of the similarities.

In the case of functional similarity, logit vectors are used as they are or converted to a label if needed. Regarding representational similarity, as feature vectors might be of different dimensions, we follow existing works \cite{Ding21} to apply the representational similarity: given feature vectors of a layer of shape $(n, d)$ where $n$ is the number of training data and $d$ is the embedding dimension, said feature vectors are normalized. Then, for two DNN models $M_1$ and $M_2$, this effectively results in comparing two matrices $m_1 = (n, d_1)$ and $m_2 = (n, d_2)$. Note that, this way, we can indeed compare different architectures as the common dimension between the two matrices is the training data.

\subsubsection{Projection Weighted Canonical Correlation Analysis (PWCCA)}

Projection Weighted Canonical Correlation Analysis (PWCCA) \cite{Morcos18} is based on Canonical Correlation Analysis which aims at finding two sets of orthogonal vectors $w_1$ and $w_2$, each of dimension $d_1$ and $d_2$ respectively, such as the projection of $m_1$ and $m_2$ onto them are maximally correlated, i.e.,
\[
   \rho(m_1, m_2) = max_{w_1, w_2} \frac{\langle m_1w_1, m_2w_2 \rangle }{|| m_1w_1 ||.|| m_2w_2 ||} 
\]

Each of the $k = max(d_1, d_2)$ pair of vectors $(w_1^{(i)}, w_2^{(i)})$ will lead to a canonical correlation value $\rho^{(i)}$. PWCCA is calculated as:
\[
PWCCA(m_1, m_2) = \sum_i \frac{\alpha^{(i)}}{\sum_j \alpha^{(j)}} \rho^{(i)} 
\]

where $\alpha^{(i)} = \sum_j | \langle m_1w_1^{(i)}, m_1{_{-,j}} \rangle | $.

\subsubsection{Centered Kernel Alignment (CKA)}

Centered Kernel Alignment (CKA) \cite{Kornblith19} uses a kernel to calculate a $n \times n$ kernel matrix. The kernel can be any kernel function commonly used. In our case, we use a linear kernel which makes CKA be defined as:

\[
CKA(m_1, m_2) = 1 - \frac{|| m_1^T m_2 ||^2_F}{|| m_1^T m_1 ||^2_F || m_2^T m_2 ||^2_F}
\]

where $|| . ||^2_F$ is the Froebinus norm.

\subsubsection{Procrustes Orthogonal (Ortho)}

Procrustes Orthogonal (Ortho) \cite{Ding21} aims to find the best orthogonal transformation to align two matrices, i.e., $min_{P} || m_1P - m_2 ||^2_F $, which admits a closed form:

\[
Ortho(m_1, m_2) = || m_1 ||^2_F + || m_2 ||^2_F - 2 || m_1^T m_2 ||_*
\]

Where $|| . ||_*$ is the nuclear norm.

\subsubsection{Performance (Acc)}

A simple similarity consists of merely comparing the performance of two given DNN models on the same dataset. Given a performance function $p$ (i.e., evaluating how well a DNN model is doing), one can define a similarity function between two DNN models $M_1$ and $M_2$ as the absolute difference between their performance. In our case, we will use the accuracy:

\[
Acc(M_1, M_2) = | p(M_1) - p(M_2) |
\]

\subsubsection{Disagreement (Dis)}

Using hard predictions (i.e., the labels), one can define a similarity that is more fine-grained than simply computing the performance. Disagreement (Dis) \cite{Madani04} is defined as the average number of conflicts between DNN models over the predicted labels. More formally, for two DNN models $M_1$ and $M_2$:

\[
Dis(M_1, M_2) = \frac{1}{N} \sum_{i=1}^N 1\!\!1 \{ argmax_j M_{1_{i,j}} \neq argmax_j M_{2_{i,j}}\}
\]

\subsubsection{Divergence (Div)}

Finally, using soft predictions (i.e., the logits), one can compare the probability distribution between two DNN models at the instances level. The Kullback-Leibler divergence \cite{Kullback51} is one of the most popular statistical distances for this purpose yet it's not symmetrical. Instead, we chose the J divergence $J_{Div}$ \cite{Kullback51} which is a symmetrical version of the Kullback-Leibler divergence. We can then take the average of said distance overall predictions with the following:

\[
J_{Div}(M_1, M_2) = \frac{1}{2N} \sum_{i=1}^N KL(M_{1_i} || M_{2_i}) + KL(M_{2_i} ||M_{1_i})
\]

Where $KL$ is the Kullback-Leibler divergence. Note that other divergence could be used and so we refer the readers to Cha et al. \cite{Cha07} for an overview of possible divergence.

\section{Experimental Design}\label{sec:exp}

We remind the readers of our research questions:

\begin{itemize}
    \item[\textbf{RQ1: }] \textbf{\rqone}

    Before showing the correlation between the similarities $\mathcal{P}$ and the property $\mathcal{P}_O$, we first set to understand the relation between similarities and measured property as well as DNN model type. The goal is to understand how similarities rank DNN model types among each other and if there is any relationship between the different similarities chosen. The goal is the same for the properties and how they relate to similarities.
    
    \item[\textbf{RQ2: }] \textbf{\rqtwo}

    The goal of this research question is to validate the hypothesis we detailed in Section \ref{sec:sim}, \ie the more similar two DNN models are, the more they cover the property of the test set. To do so, we use the \textit{Offline} procedure described in Section \ref{sec:approach} and compute the correlation between the similarities $\mathcal{P}$ and the property $\mathcal{P}_O$, for each property and each technique.

    \item[\textbf{RQ3: }] \textbf{\rqthree}

    Building on the previous results, we aim to show \textit{GIST} can be operationalized. Thus, we study if the similarities can be leveraged to carefully select test sets for transfer. First, we assess simply by choosing the most similar reference test set. We are also interested in the effect of the DNN model seeds (\ie individual initialization of a DNN model) over this choice. Secondly, we further leverage those similarities by showing they can be used to combine reference test sets in a better way than randomly sampling them.

    \item[\textbf{RQ4: }] \textbf{\rqfour}

    Finally, we measure whether transferring the test sets is more effective than simply reapplying the test case generation techniques. That is, we aim to evaluate the trade-off between property covered / relative execution time. To do so, we will measure the property the transferred test sets covered compared to a time ratio. This time ratio is the ratio between the executive time of the \textit{Offline}/\textit{Online} parts of \textit{GIST} and the executive time of applying the test case generation techniques on the DNN models under test.

    %\item[\textbf{RQ4: }] \textbf{\rqfour}

    %Finally, building on the results of RQ3, we want to further verify that we can choose a criterion to select a test set to be transferred when we are within the defined setting of \textit{GIST}. In particular, we want to ensure that, not only do we have a correlation between similarity and fault types covered (as described in RQ3), but also that when choosing a particular test set for transfer with a precise criterion, said transfer is reasonably good compared to a random baseline, which would justify the practical use of \textit{GIST}. Moreover, drawing insights from previous research questions, we explore criteria to combine test sets in order to further increase the fault type coverage.
    
\end{itemize}

\subsection{Datasets and Models}

We used the following datasets and DNN models in our experiments:

\begin{itemize}
    \item[\textbf{Image:}] CIFAR10 \cite{Krizhevsky09} is a traditional image benchmark of $32\times32$ images distributed across 10 classes (dogs, cats, planes, boats...). The training set is composed of 50,000 images and the test set of 10,000 images. For the DNN models, we used the implementation of $VGG16$, $VGG19$, $PreResnet20$, $PreResnet110$, and $Densenet100bc$ described in this repository \cite{Bigballon19}.
    \item[\textbf{Text:}] Movie Reviews \cite{Pang05} is a dataset of short positive and negative reviews from the movie reviews website Rotten Tomatoes. The dataset is split into 8,530 training data and 2,132 validation/test data. For the DNN models, we used the implementation of $Electra-small$, $XLNet$, $DistilBERT$, $BERT$, and $RoBERTa$ from HuggingFace \cite{HuggingFace}.
\end{itemize}

We trained the DNN models using the parameters given in our replication package \cite{ReplicationPackage} and retained the weights that led to the best results on the original test set. For each DNN model type, we trained $10$ DNN model seeds, \ie $10$ different initializations of the same DNN model type, to account for the effect of the stochasticity over our approach. Accuracy for the Train and Test sets are shown in Table \ref{tab:acc}. The choice of DNN models was motivated by the need to have a diversity of DNN models' type while keeping the DNN models' size manageable to reduce training and test generation time. 

\begin{table}[h]
\caption{Train and Test accuracy on both datasets for each DNN model. Results are averaged over all the DNN model seeds with standard deviation in between parenthesis.}
    \centering
    \begin{tabular}{ccc|ccc}
    \toprule
         & \multicolumn{2}{c}{CIFAR10} & & \multicolumn{2}{c}{Movie Reviews} \\
         Model & Train Accuracy & Test Accuracy & Model & Train Accuracy & Test Accuracy  \\
         \midrule
         VGG16 & 98.54 \% (0.73) & 90.49 \% (0.54) & Electra & 87.39 \% (0.34) & 82.40 \% (0.47)\\ 
         \midrule
         VGG19 & 98.05 \% (1.34) & 89.96 \% (1.07) & DistilBERT & 89.92 \% (0.59) & 83.73 \% (0.25) \\
         \midrule
         PreResnet20 & 96.08 \% (0.16) & 89.82 \% (0.19) & BERT & 89.31 \% (0.80) & 83.94 \% (0.37)  \\
         \midrule
         PreResnet110 & 98.99 \% (0.09) & 91.47 \% (0.16) & RoBERTa & 90.91\% (0.73) & 87.52 \% (0.45)  \\
         \midrule
         Densenet100bc & 99.78 \% (0.06) & 93.13 \% (0.15) & XLNet & 91.15 \% (1.09) & 87.01 \% (0.31) \\
    \bottomrule
    \end{tabular}
    
    \label{tab:acc}
\end{table}

\subsection{Test Input Generation techniques}

To generate the test sets of the reference DNN models/DNN model under test, we used the following generation technique $\textbf{G}$ depending on the dataset we were considering:

\begin{itemize}
    \item[\textbf{Image:}] Using the classification of techniques presented by Riccio et al. \cite{Riccio23}, we selected two methods: one from their \textit{Raw Input Manipulation} category and one from their \textit{Generative Deep Learning Models} category. We could not use techniques from their \textit{Model-based Input Manipulatio}n category as they are not applicable to CIFAR10. For \textit{Raw Input Manipulation}, we could have picked methods such as DLFuzz \cite{Guo18} or DeepXplore \cite{Pei19}. However, we preferred to rely on a newer fuzzing technique that does not rely on neuron coverage for the fuzzing process which has shown to have several shortcomings \cite{Harel-Canada20, Li19}. We made use of FolFuzz \cite{Chen23} which uses First Order Loss for the fuzzing process. We leverage their technique, keeping the default parameters mentioned in the paper but just reducing the $\epsilon$ value, which controls how far the fuzzed example is from the original seed to $0.05$. In doing so, the generated test cases stay very close to the inputs used to generate them. For the \textit{Generative Deep Learning Models} category, we leveraged the Feature Perturbation-based method by Dunn et al. \cite{Dun21} that relies on Generative Adversarial Networks (GAN). As the technique makes use of a BigGAN architecture for the test input generation process, we use the MMGeneration library \cite{mmgeneration21} implementation and checkpoints of the CIFAR10 BigGAN. For both methods, we generated 1,000 test inputs for each DNN model listed in the previous subsection in order to have a representative set. To avoid bias, the same starting test inputs were used for the generation in both methods so the test set will mainly differ because of the generation process (and so the DNN models used) rather than the starting inputs used.
    \item[\textbf{Text:}] For the Text dataset, we make use of the TextAttack library \cite{Morris20-1}. Multiple methods are implemented which make use of similar properties (swapping words, adding letters...). We chose to use one of the latest generation methods TFAdjusted \cite{Morris20} which, contrary to other methods, has shown to be better at preserving grammatical and semantical similarity. We use the provided implementation with the default parameters. As TFAdjusted will not generate a lot of misclassified inputs (because of the added constraints to ensure similarity preservation), the 2,000 base inputs of the validation set won't be enough to generate a big enough test set. To mitigate this issue, we scrapped the Rotten Tomatoes website for movie reviews in the period 2014-2022 (different from the base dataset released in 2005 to avoid overlaps) in order to extract reviews similar to the base dataset. We then used the ratings (0-5 scale) from the users in order to label them as negative or positive. To be conservative on the labelling, we consider reviews to be negative if they receive a score strictly below 2.0 and to be positive if they receive a score strictly above 3.5. Reviews with ratings not following the constraints were discarded. This allowed us to gather around 9,000 inputs we could leverage for the TFAdjusted method. Accuracy on those inputs was consistent with the test accuracy obtained on the original Movie Review dataset. Applying said method on the extracted inputs returns around 400 test-generated inputs for each DNN model. Data are available in the replication package \cite{ReplicationPackage}.
\end{itemize}

\subsection{Features Extraction for Similarity}\label{sec:feat_ext}

As we mentioned in Section \ref{sec:approach}, to apply the similarity for the representational metrics, we need to extract features over some layer that has a close functionality across DNN models for the comparison to be relevant. Intuitively, among the inner layers, the layers before the classification head are good candidates for such tasks. In our case, and based on the DNN models we had to compare, we used: 

\begin{itemize}
    \item[\textbf{Image:}] The output of the final pooling layer. This pooling layer makes the interface between the feature extractor and the classifier head and is present across all classifiers. As such, it makes a good candidate to evaluate their similarity.
    \item[\textbf{Text:}] The pooled \textit{[CLS]} token. Similar to image-based DNN models, the text-based DNN models pool the embedding obtained by the feature extractor in order for it to be fed to a classification head. This pooling operation is generally done by taking the \textit{[CLS]} token's last hidden state. This token, generally added at the start of the sentence, acts as a summary of the semantics of each sentence thanks to the pretraining transformers undergo. As such, to have a relevant comparison, we took the last hidden state of each DNN model associated with the \textit{[CLS]} token. Note that for some transformers, such as $XLNet$, said \textit{[CLS]} token is situated at the end of the sentence and so we made the necessary adjustments when needed. Previous work made use of a similar approach to compare transformers \cite{Singh19}.
\end{itemize}

\section{Experimental Results}\label{sec:res}

\subsection{Clustering and validations}\label{sec:valid}

Before using the fault types criteria, we validate the clustering strategy still relates to fault types since we changed the original setting by Aghababaeyan et al. \cite{Aghababaeyan23}. To obtain the fault types of each test set on each object DNN model we applied the clustering method described in Section \ref{sec:approach}, by considering each individual DNN model as a DNN model under test while all other DNN models were considered reference DNN models. When clustering, we only used data from one generation technique at a time. We excluded from the reference DNN models all DNN models of the same type as the DNN model under test, \eg if the DNN model under test is a \textit{densenet100bc} trained with seed $0$, then we did not use any \textit{densenet100bc} trained with any other seed as part of the reference DNN models. The reasoning for this is that the test sets transfer procedure is supposed to be used when a new DNN model is being trained for which we have no reference. As such, using the same DNN model only differing by the seed is pointless in our case, as it defeats the purpose of having to transfer a test set without applying the generation technique $\textbf{G}$ on the DNN model under test.

We obtained DBCV and silhouette scores ranging from around 0.4 to 0.8 in all cases. The scores primarily differed between generation techniques and remained relatively similar for a given DNN model type across different DNN model seeds. The lowest scores were obtained when clustering data on GAN generated test sets, which can be expected as this technique relies on GAN-generated images whose distribution is not strictly similar to the trained data. Moreover, the generation technique modifies \textit{features} of the images rather than pixel values as is the case with fuzzing or swapping words as is the case with TFAdjusted. This would thus lead to more differences between data and make clustering harder. We included the hyperparameters used and the score obtained for each DNN model in our replication package \cite{ReplicationPackage} for further details.

As we mentioned in Section \ref{sec:approach}, we verify that clusters correspond to fault types, following a similar procedure as in the original method of Aghababaeyan et al. \cite{Aghababaeyan23}. We picked two DNN models per generation technique of test set generation and evaluated the five biggest clusters. To do that, we retrained the DNN model \cite{Hu22, Shen21} using the original training data and 85\% of the data of a cluster. Said data are generated using each of the generation techniques and do not overlap with the train data. We then evaluate the accuracy on the remaining 15\% of said clusters as well as other clusters. In a nutshell, to choose the 15 and 85 percentages we strictly followed the same method as Aghababaeyan et al. \cite{Aghababaeyan23}. We did so three times for each cluster selected to verify that the split choice did not affect the results much. We expect that different clusters should represent individual fault types. This would translate to the accuracy of the chosen cluster, from which we used part of the data to retrain the DNN model, to be higher than on other clusters. We did not expect a perfect separation of fault types because of the limitations of the clustering techniques as well as the noise introduced in the test sets by the test generation technique. Moreover, since the same starting test inputs were used to generate the different reference test sets, the same starting inputs could lead to generated test sets clustered in different fault types (because they have a different impact on the DNN model under test). However, when retraining the DNN model with those inputs, they could affect each other as they were generated using the same input even if they differ in fault types. Results are presented in Table \ref{tab:clust_val}.

\begin{table}[h]
    \centering
    \caption{Accuracy when retraining for the cluster from which data was used for retraining and for all other clusters. Results are averaged over three splits.}
    \resizebox{\textwidth}{!}{\begin{tabular}{c|c|c|c|c|c|c|c|c|}
    \toprule
    & DNN & Cluster $C_i$ & Accuracy on the & Average Accuracy on the & Model & Cluster $C_i$ & Accuracy on the & Average Accuracy on the\\
    & & & cluster $C_i$ &  other cluster $C_{j \neq i}$ & & & cluster $C_i$ &  other cluster $C_{j \neq i}$\\
    \midrule
    \midrule
    \multirow{5}{*}{\rotatebox[origin=c]{90}{Fuzz}}& \multicolumn{1}{c|}{\multirow{5}{*}{\rotatebox[origin=c]{90}{Densenet100bc}}} & Cluster 1 & 85.19\% & 10.32\% & \multicolumn{1}{c|}{\multirow{5}{*}{\rotatebox[origin=c]{90}{VGG19}}} & Cluster 1 & 70.37\% & 8.25\% \\ \cline{3-5} \cline{7-9}
    & & Cluster 2 & 74.07\% & 10.29\% & & Cluster 2 & 45.24\% & 9.88\% \\ \cline{3-5} \cline{7-9}
    & & Cluster 3 & 82.05\% & 9.69\% &  & Cluster 3 & 71.11\% & 12.15\% \\ \cline{3-5} \cline{7-9}
    & & Cluster 4 & 100.00\% & 6.59\% &  & Cluster 4 & 70.59\% & 10.49\% \\ \cline{3-5} \cline{7-9}
    & & Cluster 5 & 90.91\% & 9.65\% & & Cluster 5 & 71.43\% & 9.54\% \\ \cline{3-5} \cline{7-9}
    \midrule
    \midrule
    \multirow{5}{*}{\rotatebox[origin=c]{90}{GAN}} & \multicolumn{1}{c|}{\multirow{5}{*}{\rotatebox[origin=c]{90}{PreResnet20}}} & Cluster 1 & 33.33\% & 6.30\% & \multicolumn{1}{c|}{\multirow{5}{*}{\rotatebox[origin=c]{90}{VGG16}}} & Cluster 1 & 58.97\% & 7.70\% \\ \cline{3-5} \cline{7-9}
    & & Cluster 2 & 38.89\% & 8.87\% &  & Cluster 2 & 74.07\% & 9.17\% \\ \cline{3-5} \cline{7-9}
    & & Cluster 3 & 94.44\% & 9.01\% &  & Cluster 3 & 57.58\% & 11.23\% \\ \cline{3-5} \cline{7-9}
    & & Cluster 4 & 46.67\% & 5.63\% &  & Cluster 4 & 42.86\% & 7.85\% \\ \cline{3-5} \cline{7-9}
    & & Cluster 5 & 42.86\% & 6.33\% & & Cluster 5 & 52.38\% & 9.70\% \\ \cline{3-5} \cline{7-9}  
    \midrule
    \midrule
    \multirow{5}{*}{\rotatebox[origin=c]{90}{Text}} &\multicolumn{1}{c|}{\multirow{5}{*}{\rotatebox[origin=c]{90}{RoBERTa}}} & Cluster 1 & 100.0\% & 33.33\% & \multicolumn{1}{c|}{\multirow{5}{*}{\rotatebox[origin=c]{90}{DistilBERT}}} & Cluster 1 & 85.71\% & 36.52\% \\ \cline{3-5} \cline{7-9}  
    & & Cluster 2 & 100.0\% & 39.88\% &  & Cluster 2 & 95.83\% & 25.07\% \\ \cline{3-5} \cline{7-9}  
    & & Cluster 3 & 100.0\% & 32.71\% &  & Cluster 3 & 100.0\% & 54.86\% \\ \cline{3-5} \cline{7-9}  
    & & Cluster 4 & 93.33\% & 49.63\% &  & Cluster 4 & 83.33\% & 37.61\% \\ \cline{3-5} \cline{7-9}  
    & & Cluster 5 & 83.33\% & 8.03\% & & Cluster 5 & 92.31\% & 44.89\% \\ \cline{3-5} \cline{7-9}    
    \bottomrule
    \end{tabular}}    
    \label{tab:clust_val}
\end{table}

The results over the \textit{Fuzz}-based generation technique have the most striking difference between accuracy in/out clusters, which can be explained by the fact that this technique applies low-level perturbations and so the fuzzed image will match more closely with inputs from the original test set than for instance the GAN technique which generates images from noise. The accuracy for the other clusters is generally the highest for the \textit{Text}-based generation technique which we assume can be because the dataset only has two classes. Indeed, the noise introduced by the data from any clusters might make it easier to flip to the opposite (and so correct) class, even if the data do not belong to the same cluster. Moreover, we did not expect perfect separation, as it is likely that inputs do not exhibit just one fault type but share multiple ones in common. Finally, contrary to Aghababaeyan et al. \cite{Aghababaeyan23}, we did not make use of the original test set in the clustering, but we made use of the test set obtained using some input generation technique, which ultimately introduced additional noise into the data. Overall, we note that the accuracy of the clusters drastically improved compared to the accuracy of other clusters. This points towards clusters representing individual forms of fault types, thus validating our fault types clustering approach.

\subsection{RQ1: \rqone}

\begin{figure}\centering
\centering
\begin{subfigure}{0.4\textwidth}
    \includegraphics[width=\textwidth]{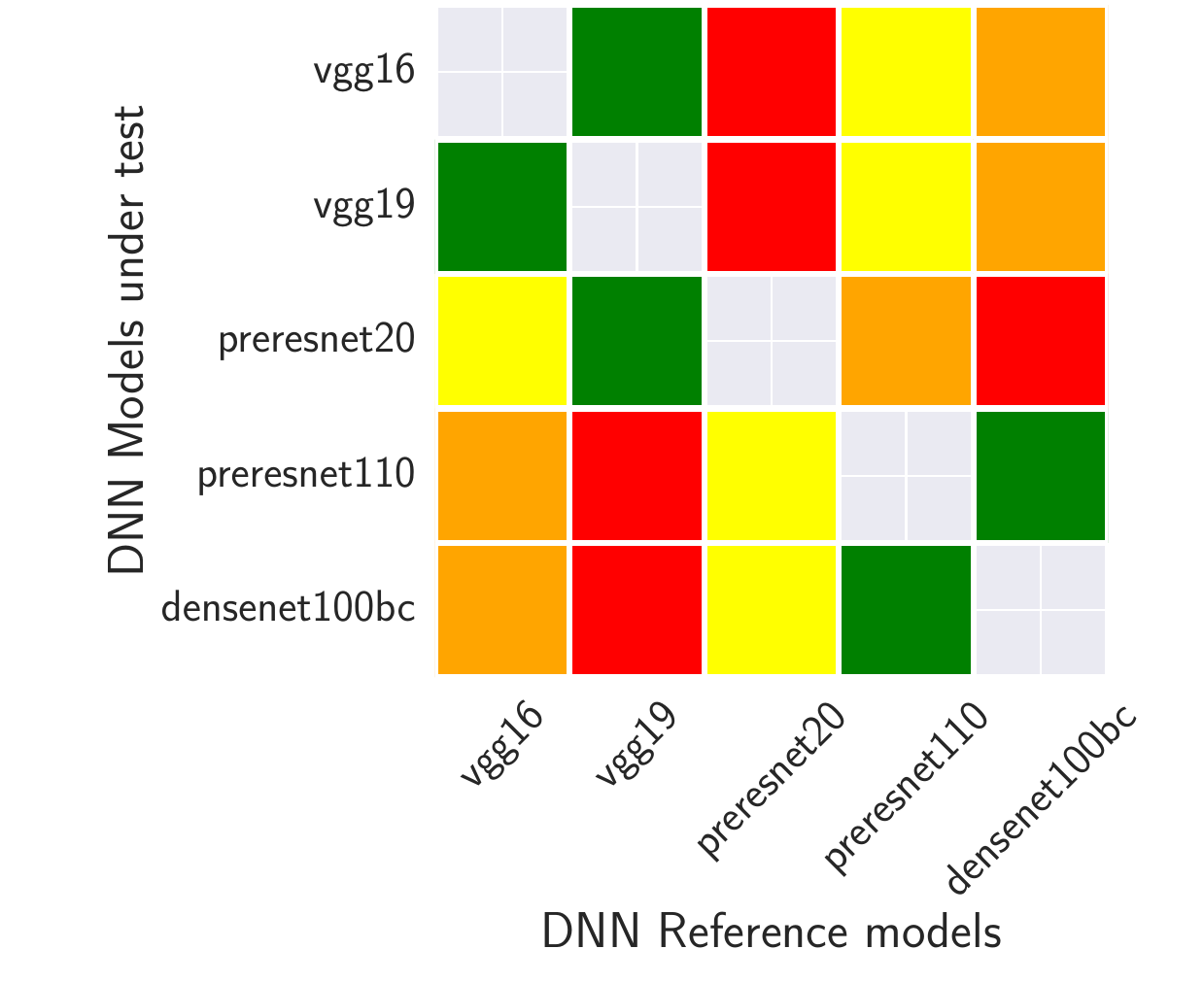}
    \caption{Fault types - \textit{Fuzz}-based generation}
    \label{subfig:fuzz_fault}
\end{subfigure}
\hfill
\begin{subfigure}{0.4\textwidth}
    \includegraphics[width=\textwidth]{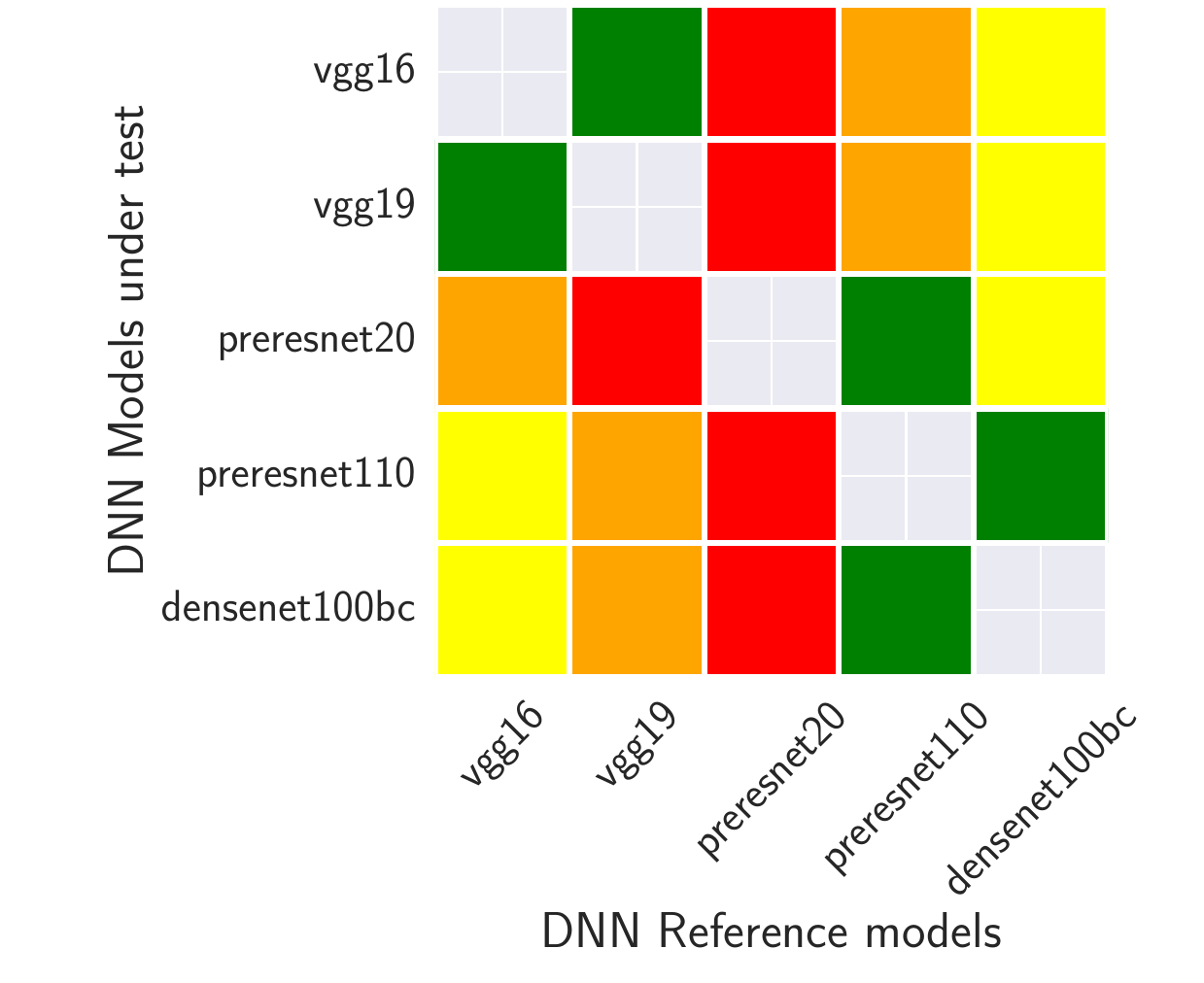}
    \caption{Neuron coverage - \textit{Fuzz}-based generation}
    \label{subfig:fuzz_nc}
\end{subfigure}
\hfill
\begin{subfigure}{0.4\textwidth}
    \includegraphics[width=\textwidth]{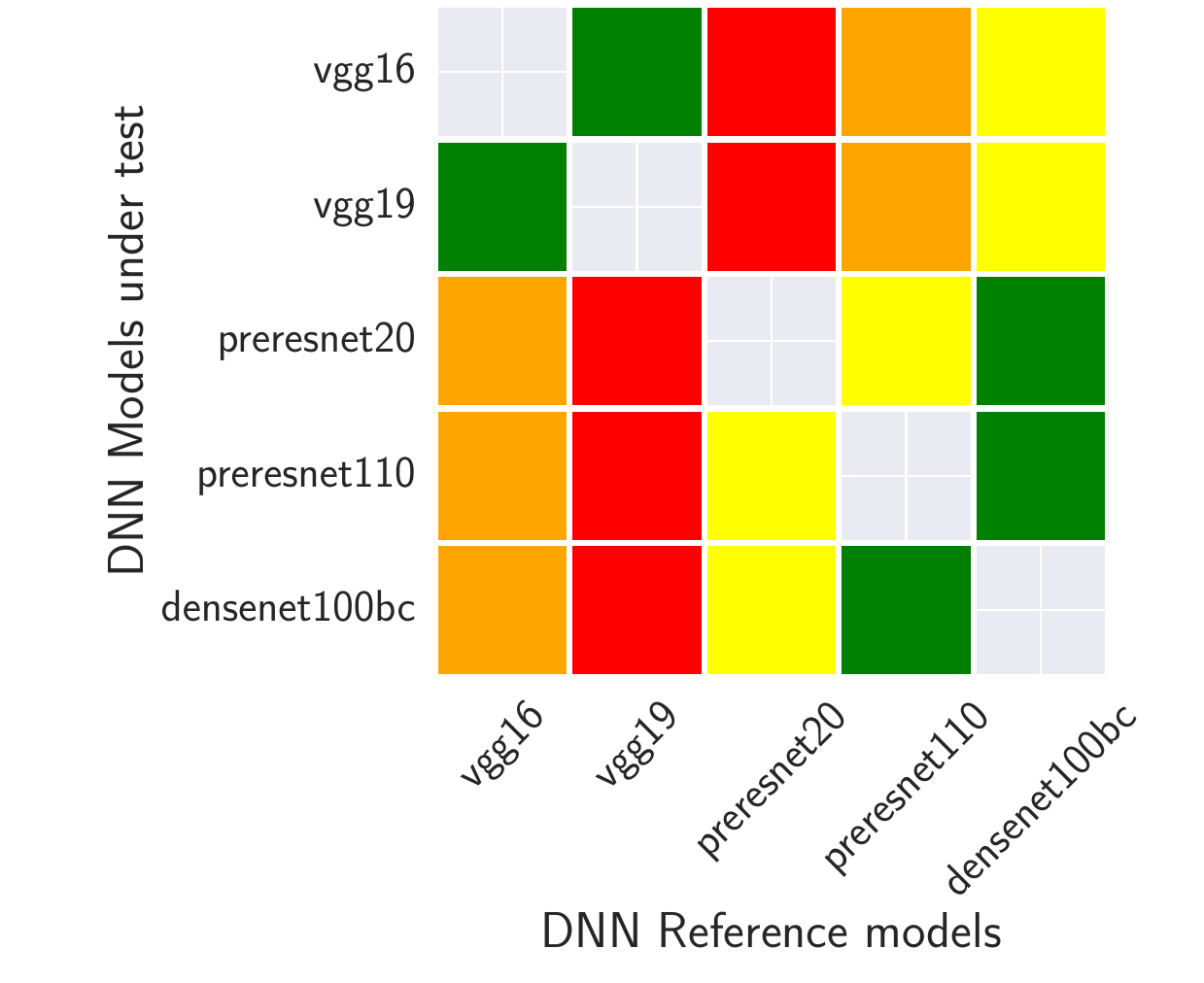}
    \caption{Fault types - \textit{GAN}-based generation}
    \label{subfig:gan_fault}
\end{subfigure}
\hfill
\begin{subfigure}{0.4\textwidth}
    \includegraphics[width=\textwidth]{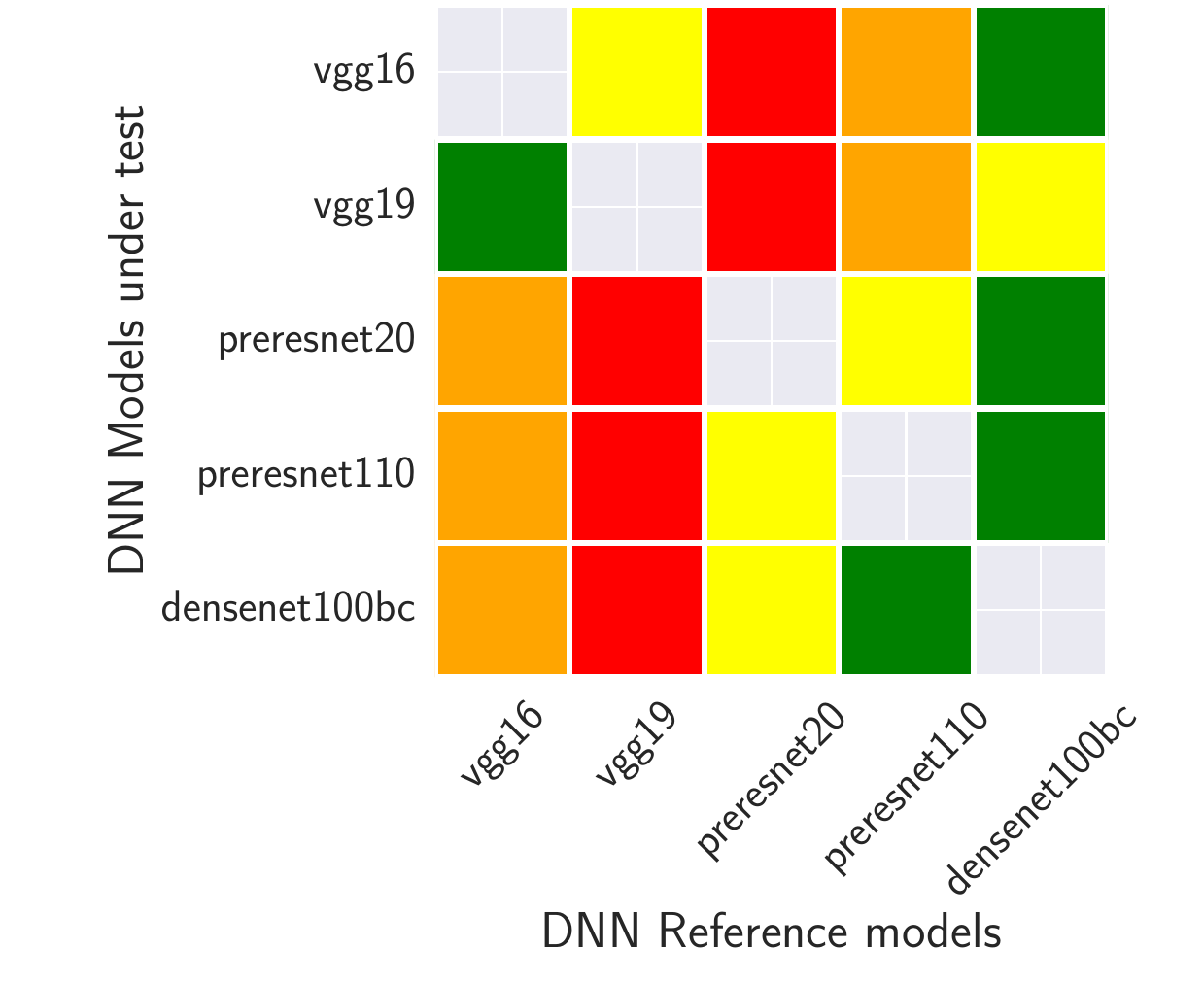}
    \caption{Neuron coverage - \textit{GAN}-based generation}
    \label{subfig:gan_nc}
\end{subfigure}
\hfill
\begin{subfigure}{\textwidth}
    \includegraphics[width=\textwidth]{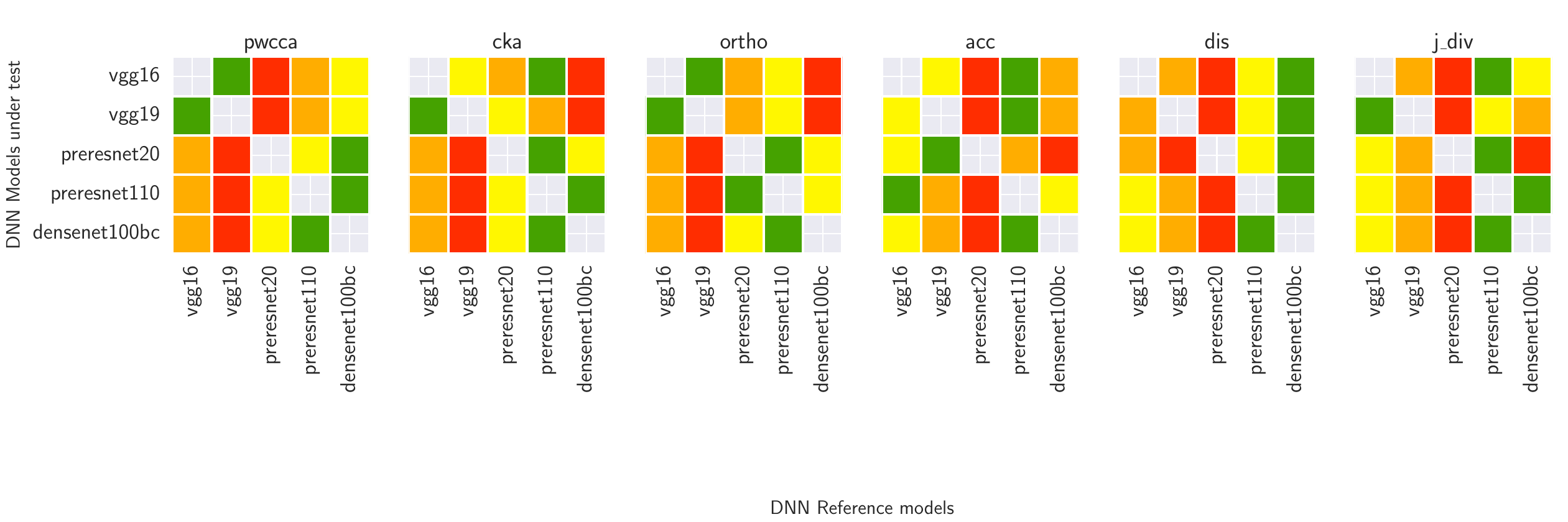}
    \caption{Similarities}
    \label{subfig:sim_cif}
\end{subfigure}
        
\caption{Average ranking of closest reference DNN model types in terms of $\mathcal{P}_O$ and similarity on image dataset according to a given DNN model under test. Colours indicate ranking (\textcolor{codegreen}{green} = 1$^{st}$, \textcolor{red}{red} = 4$^{th}$).}
\label{fig:comp_sim_cif}
\end{figure}

\begin{figure}\centering
\centering
\begin{subfigure}{0.4\textwidth}
    \includegraphics[width=\textwidth]{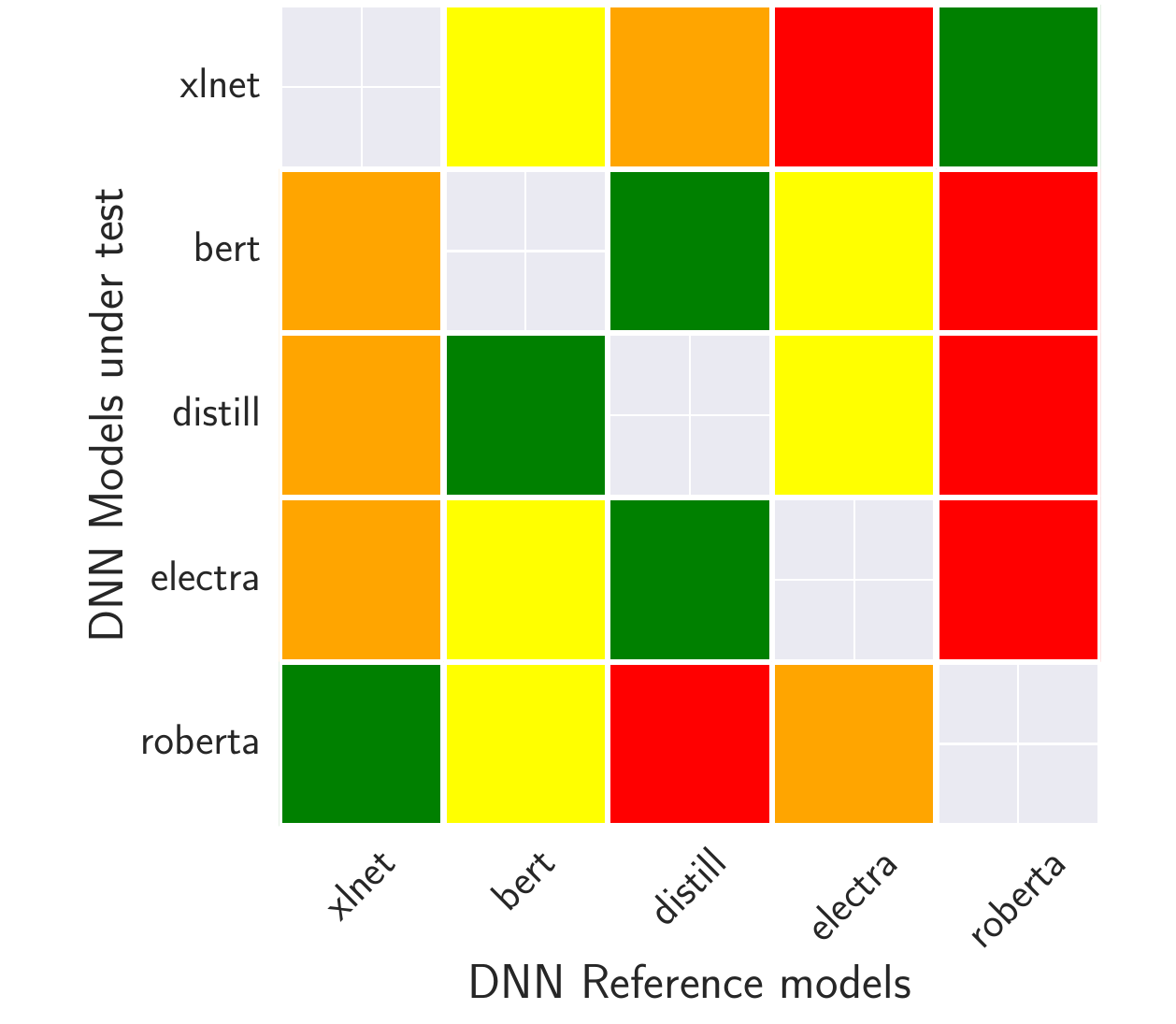}
    \caption{Fault types - \textit{Text}-based generation}
\end{subfigure}
\hfill
\begin{subfigure}{0.4\textwidth}
    \includegraphics[width=\textwidth]{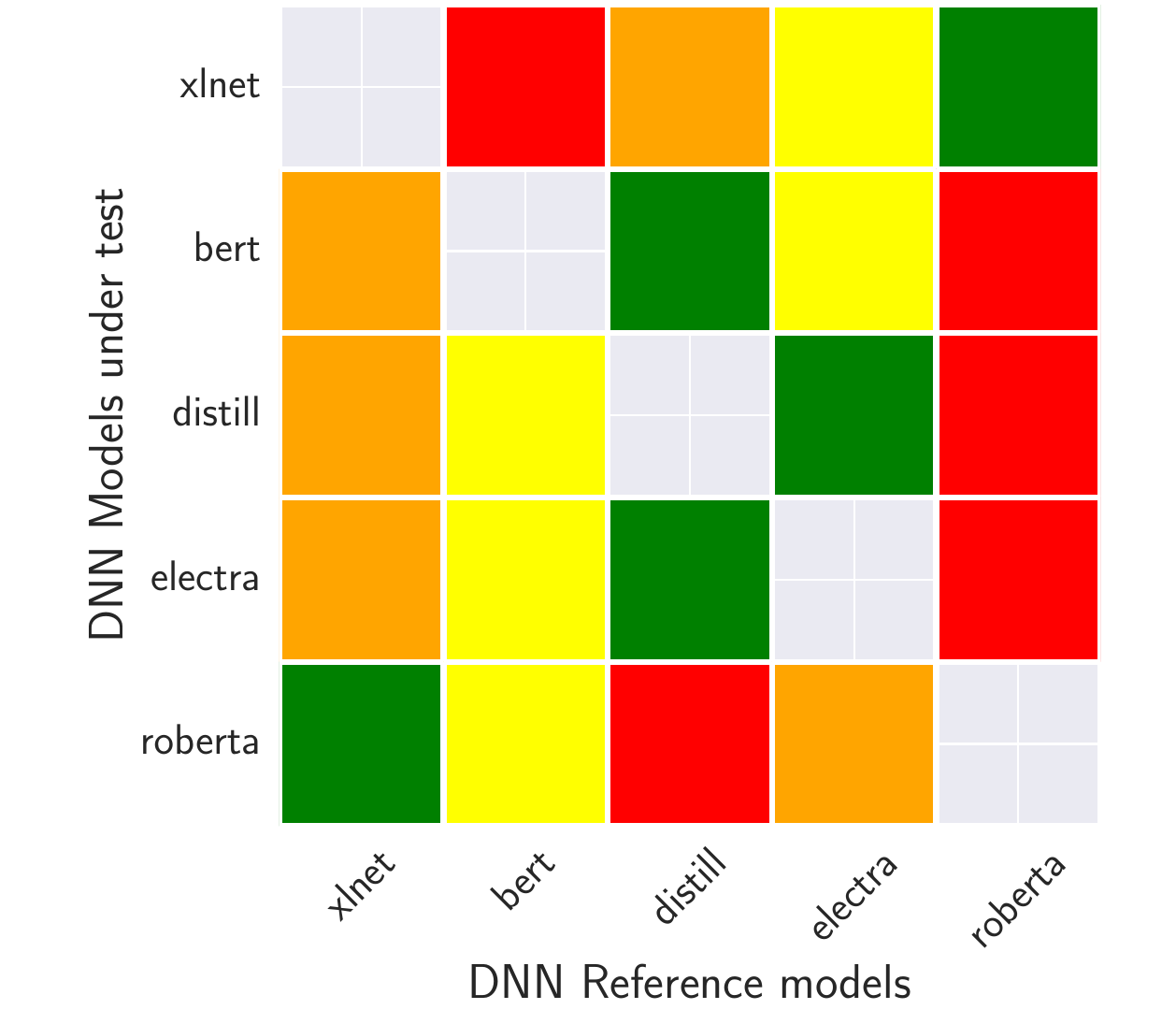}
    \caption{Neuron coverage - \textit{Text}-based generation}
\end{subfigure}
\hfill
\begin{subfigure}{\textwidth}
    \includegraphics[width=\textwidth]{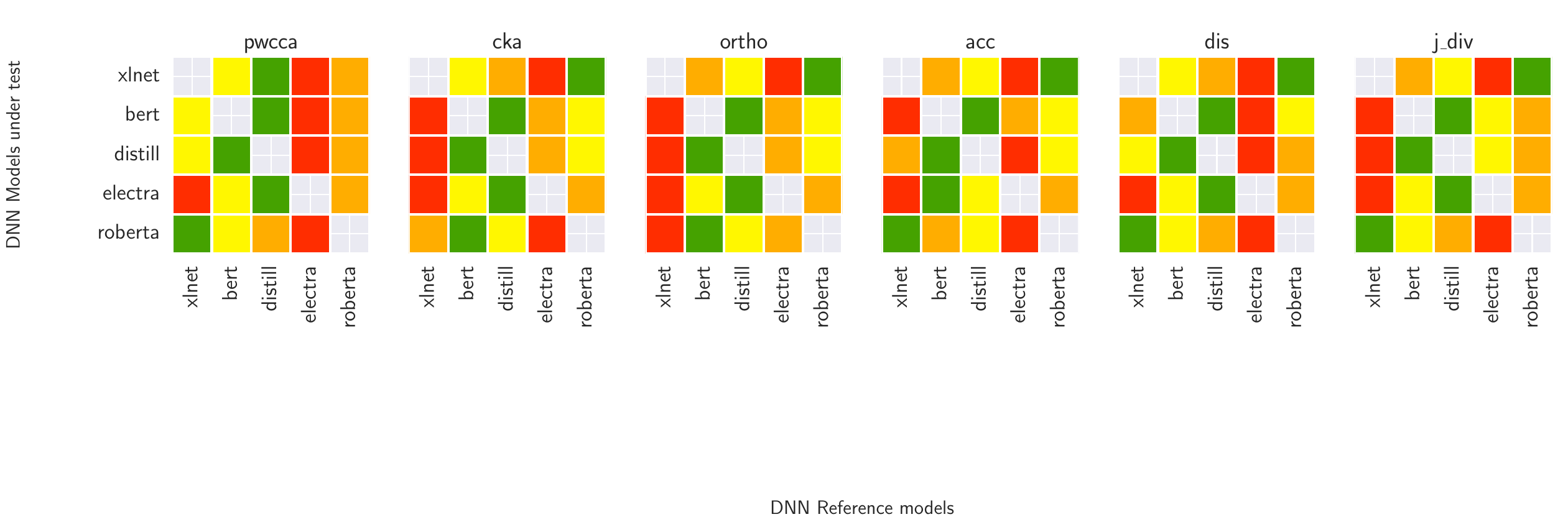}
    \caption{Similarities}
    \label{subfig:sim_mr}
\end{subfigure}
        
\caption{Average ranking of closest reference DNN model types in terms of $\mathcal{P}_O$ and similarity on text dataset according to a given DNN model under test. Colours indicate ranking (\textcolor{codegreen}{green} = 1$^{st}$, \textcolor{red}{red} = 4$^{th}$).}
\label{fig:comp_sim_mr}
\end{figure}

To study the relation between similarity and property, we first extract fault types and neuron activation on a given DNN model under test for each test set (all the reference test sets and the own test set of the DNN model under test). We then calculate $\mathcal{P}_O$ using Equations \ref{eq:m_o_k} and \ref{eq:m_o_f}. Then, we calculate the pairwise similarity, between the DNN model under test and each reference DNN model, using each of the similarity metrics described in Section \ref{sec:sim}. Then, we averaged the obtained $\mathcal{P}_O$ for each property for a given pair (type of DNN model under test / type of reference DNN model) across the DNN model seeds. Similarly, we averaged the obtained $\mathcal{P}$ for each similarity metric for a given pair (type of DNN model under test / type of reference DNN model) across the DNN model seeds. Finally, we rank the obtained value for both metrics from 1 (most similar / covering) to 4 (least similar / covering). In this part, we are interested in their relative value to understand the underlying mechanism. Results are presented in Figure \ref{fig:comp_sim_cif} for the \textit{Fuzz}-based and \textit{GAN}-based generation techniques and in Figure \ref{fig:comp_sim_mr} for the \textit{Text}-based generation technique.

The way the heatmap for the average $\mathcal{P}_O$ can be read is, row-wise, "for a given type of DNN model under test, the averaged rank in terms of the property covered by a given reference test set is \textit{colour}". For the similarity heatmap, it can also be read row-wise as "for a given type of DNN model under test, the given similarity rank reference DNN model as \textit{colour} on average" where the colour indicates the rank (\textcolor{codegreen}{green} = 1$^{st}$, \textcolor{red}{red} = 4$^{th}$). As one can see across the figures, there is some link between the rank (and so percentage) of the property covered and the DNN model type used. For instance, in Figure \ref{fig:comp_sim_cif}, both \textit{VGG} type DNN models generally lead to more of the property being covered when we compare their $T_O$/$T_R$ than the rest. Similarly, \textit{Preresnet110} as well as \textit{Densenet100bc} have more covering properties compared to the rest. Analogous observations can be made for the \textit{GAN}-based generation technique. For the \textit{Text}-based generation technique, \textit{Roberta} and \textit{XLNet} generally have more covering properties while \textit{Bert}, \textit{DistilBert} and \textit{Electra} have more covering properties. Note in that case that \textit{Bert} and \textit{DistilBert} have generally more covering properties than with \textit{Electra} which is logical as \textit{DistilBert} is obtained by distilling a \textit{Bert} DNN model. While heatmaps are similar between properties (fault-type and neuron coverage-based), there are still some differences: thus, for the \textit{Fuzz}-based generation technique and $PreResnet20$ DNN model under test, it is $VGG$ DNN models that are ranked better for the fault-type property (Figure \ref{subfig:fuzz_fault}), while for the neuron property, it is $PreResnet110$/$Denset100bc$ (Figure \ref{subfig:fuzz_nc}). Thus, it is likely that the best-fit similarity will not be the same in all situations.

If we compare the patterns obtained with the similarity heatmap, we can already infer that potential correlations can happen. As an example, for the fault-type property for the \textit{GAN}-based generation technique (Fig \ref{subfig:gan_fault} and \ref{subfig:gan_nc}), we see that the ranking of $\mathcal{P}_O$ closely follows the one made by $PWCCA$ on all DNN model under test (Figure \ref{subfig:sim_cif}). Thus, it is likely that, for this setting, there will be a correlation. Similarly for both properties for the \textit{Text}-based generation technique (Fig \ref{subfig:sim_mr}), $J_{Div}$. On the contrary, similarities such as $Acc$ exhibit patterns that are less aligned with the obtained $\mathcal{P}_O$ heatmaps, which means they are less likely to lead to positive correlations.

Looking at the similarity heatmap ranking (Figure \ref{subfig:sim_cif} and \ref{subfig:sim_mr}), we see that the focus on what are considered two similar DNN models is not the same. On representational similarities, $Ortho$ (Procrustes Orthogonal), and CKA to a lesser extent, tend to focus on the architecture of the DNN models when making the comparisons compared to $PWCCA$ or functional metrics. Thus, for image-based data, the $VGG$ DNN model types on one side and the $Preresnet$ DNN model types on the other side will tend to be linked together while for text-based data the \enquote{BERT} family will be grouped (RoBERTa, BERT and DistillBERT) together. On the contrary, DNN model types such as \textit{XLNet} are considered very different, having indeed a different architecture compared to \enquote{BERT} family DNN models. On the other hand, $PWCCA$ seems not to rely only on the architecture, which nonetheless seems to make it more aligned with the properties. Those observations on these representational similarities echo previous studies \cite{Boix22}. Functional similarities are more dissimilar among each other in terms of ranking, with $Acc$ being more different than $Dis$ and $J_{Div}$. Interestingly, $J_{Div}$ seems to better match the heatmap patterns of the representational similarities, showing potential alignments on those similarities. For instance, on the \textit{Text}-based generation technique, $J_{Div}$ closely follows $Ortho$ on all but $RoBERTa$ DNN model type. As there is no consensus across the metrics on the DNN model type, showing that similarities do rely on different aspects to quantify a similarity, studying multiple similarities for correlation seems to be a sound approach.

\begin{tcolorbox}[colback=blue!5,colframe=blue!40!black]
\textbf{Findings 1:} Similarities rank DNN model types differently based on their characteristics: for instance, representational and functional similarities will not necessarily consider the same DNN model types as most similar. While overlap exists, there is no agreement between similarities in any of our property/generation technique couples. This suggests that exploring multiple similarities as the proxy is important. Finally, $\mathcal{P}_O$ tends to be better covered by reference test sets associated with specific DNN model types. These patterns overlap with certain similarity ranking patterns suggesting those specific similarities could be leveraged as a proxy, in our case $PWCCA$ and $J_{Div}$.
\end{tcolorbox}

\subsection{RQ2: \rqtwo}

After analyzing similarities/property patterns, we aim to verify that (some of) the similarities we chose (\ie $\mathcal{P}$) are a correct proxy for the property we wish to transfer (\ie $\mathcal{P}_O$). To do so, similarly to RQ1, we compute both $\mathcal{P}$ and $\mathcal{P}_O$ following the offline procedure described in Figure \ref{fig:gen_framework} with each DNN model at our disposal is considered in turn as a DNN model under test. We then assess the correlation between those quantities using Kendall $\tau$ \cite{Kendall38} as recommended to evaluate such metrics \cite{Ding21}. Results are presented in Table \ref{tab:corr_faults} for the $\mathcal{P}_O$ based on Fault Types and in Table \ref{tab:corr_kmnc} based on neuron coverage. As each DNN model seed of each DNN model type is used as a DNN model under test, for each generation technique, we calculated the median, the 1st quartile as well as the 3rd quartile for Kendall $\tau$ across all DNN model seeds. We also count the number of times the correlation is significant at the $0.05$ and $0.1$ threshold across all DNN model seeds. Note that, in our case, we do not need to consider any correction over the $p$-value, such as the Bonferroni correction, as we do not perform multiple comparisons but consider the test individually \cite{Armstrong14}. Results are given in Table \ref{tab:corr_faults} for the fault-type property and in Table \ref{tab:corr_kmnc} for the neuron property.

\begin{table}[h]
%\fontsize{15}{17}\selectfont
\caption{Correlation for each generation technique for each given DNN model being considered as a DNN model under test for $\mathcal{P}_O$ based on Fault Types. For the same DNN model type (\eg \textit{Densenet100bc}), the results are given for all DNN model seeds. We highlight in bold the highest results. Generation techniques used are (top to bottom): \textit{Fuzz}-based, \textit{GAN}-based and \textit{Text}-based.}
\centering
\hspace*{-2em}
\resizebox{1.105\textwidth}{!}{
%\Huge
\begin{tabular}{c|*{6}{c}|*{6}{c}|*{6}{c}|*{6}{c}|*{6}{c}}
\toprule
DNN&\multicolumn{6}{c}{Densenet100bc}&\multicolumn{6}{c}{Preresnet110}&\multicolumn{6}{c}{Preresnet20}&\multicolumn{6}{c}{VGG19}&\multicolumn{6}{c}{VGG16} \\ \cline{2-31}
\multirow{2}{*}{Similarity} & &&&&&&&&&&&&&&&&&&&&&&&&&&&&& \\
($\mathcal{P}$)&\rotatebox[origin=c]{75}{PWCCA}&\rotatebox[origin=c]{75}{CKA}&\rotatebox[origin=c]{75}{Ortho}&\rotatebox[origin=c]{75}{Acc}&\rotatebox[origin=c]{75}{Dis}&\rotatebox[origin=c]{75}{J$_{Div}$}&\rotatebox[origin=c]{75}{PWCCA}&\rotatebox[origin=c]{75}{CKA}&\rotatebox[origin=c]{75}{Ortho}&\rotatebox[origin=c]{75}{Acc}&\rotatebox[origin=c]{75}{Dis}&\rotatebox[origin=c]{75}{J$_{Div}$}&\rotatebox[origin=c]{75}{PWCCA}&\rotatebox[origin=c]{75}{CKA}&\rotatebox[origin=c]{75}{Ortho}&\rotatebox[origin=c]{75}{Acc}&\rotatebox[origin=c]{75}{Dis}&\rotatebox[origin=c]{75}{J$_{Div}$}&\rotatebox[origin=c]{75}{PWCCA}&\rotatebox[origin=c]{75}{CKA}&\rotatebox[origin=c]{75}{Ortho}&\rotatebox[origin=c]{75}{Acc}&\rotatebox[origin=c]{75}{Dis}&\rotatebox[origin=c]{75}{J$_{Div}$}&\rotatebox[origin=c]{75}{PWCCA}&\rotatebox[origin=c]{75}{CKA}&\rotatebox[origin=c]{75}{Ortho}&\rotatebox[origin=c]{75}{Acc}&\rotatebox[origin=c]{75}{Dis}&\rotatebox[origin=c]{75}{J$_{Div}$} \\
\midrule
\midrule
tau-$\kappa$ (median) $\uparrow$& 0.44 &0.45&\textbf{0.46}&0.29&0.31&0.38&\textbf{0.43}&0.35&0.26&0.03&0.28&0.31&-0.22&-0.07&-0.07&\textbf{0.25}&-0.22&\textbf{0.25}&\textbf{0.55}&0.17&0.37&0.43&0.29&0.50&\textbf{0.54}&0.19&0.34&0.30&0.14&0.28\\[0.5em]
tau-$\kappa$ ($1^{st}$-Q) $\uparrow$&0.41&\textbf{0.43}&0.42&0.26&0.28&0.36&\textbf{0.35}&0.28&0.12&0.00&0.23&0.26&-0.34&-0.11&-0.11&0.20&-0.27&\textbf{0.21}&\textbf{0.51}&-0.03&0.32&0.32&0.26&0.37&\textbf{0.46}&0.09&0.31&0.24&0.10&0.16 \\[0.5em]
tau-$\kappa$ ($3^{rd}$-Q) $\uparrow$&0.47&0.48&\textbf{0.50}&0.36&0.37&0.45&\textbf{0.45}&0.42&0.27&0.16&0.34&0.36&-0.17&0.03&0.05&0.30&-0.15&\textbf{0.34}&0.57&0.20&0.40&0.49&0.32&\textbf{0.59}&\textbf{0.57}&0.27&0.35&0.35&0.21&0.41 \\[0.5em] \\[-1em] \cline{1-31}\\[-0.8em]
$\#$ p-value $< 0.05$&\textbf{100}\%&\textbf{100}\%&\textbf{100}\%&\textbf{100}\%&\textbf{100}\%&\textbf{100}\%&\textbf{90}\%&\textbf{90}\%&60\%&20\%&80\%&80\%&50\%&10\%&10\%&60\%&50\%&\textbf{70}\%&\textbf{100}\%&30\%&80\%&\textbf{100}\%&\textbf{100}\%&80\%&\textbf{100}\%&40\%&\textbf{100}\%&80\%&30\%&70\% \\[0.5em]
$\#$ p-value $< 0.1$&\textbf{100}\%&\textbf{100}\%&\textbf{100}\%&\textbf{100}\%&\textbf{100}\%&\textbf{100}\%&\textbf{90}\%&\textbf{90}\%&60\%&20\%&80\%&\textbf{90}\%&50\%&10\%&20\%&\textbf{90}\%&70\%&\textbf{90}\%&\textbf{100}\%&50\%&80\%&\textbf{100}\%&\textbf{100}\%&80\%&\textbf{100}\%&70\%&\textbf{100}\%&\textbf{100}\%&30\%&70\% \\[0.5em]
\bottomrule
DNN&\multicolumn{6}{c}{Densenet100bc}&\multicolumn{6}{c}{Preresnet110}&\multicolumn{6}{c}{Preresnet20}&\multicolumn{6}{c}{VGG19}&\multicolumn{6}{c}{VGG16} \\ \cline{2-31}
\multirow{2}{*}{Similarity} & &&&&&&&&&&&&&&&&&&&&&&&&&&&&& \\
($\mathcal{P}$)&\rotatebox[origin=c]{75}{PWCCA}&\rotatebox[origin=c]{75}{CKA}&\rotatebox[origin=c]{75}{Ortho}&\rotatebox[origin=c]{75}{Acc}&\rotatebox[origin=c]{75}{Dis}&\rotatebox[origin=c]{75}{J$_{Div}$}&\rotatebox[origin=c]{75}{PWCCA}&\rotatebox[origin=c]{75}{CKA}&\rotatebox[origin=c]{75}{Ortho}&\rotatebox[origin=c]{75}{Acc}&\rotatebox[origin=c]{75}{Dis}&\rotatebox[origin=c]{75}{J$_{Div}$}&\rotatebox[origin=c]{75}{PWCCA}&\rotatebox[origin=c]{75}{CKA}&\rotatebox[origin=c]{75}{Ortho}&\rotatebox[origin=c]{75}{Acc}&\rotatebox[origin=c]{75}{Dis}&\rotatebox[origin=c]{75}{J$_{Div}$}&\rotatebox[origin=c]{75}{PWCCA}&\rotatebox[origin=c]{75}{CKA}&\rotatebox[origin=c]{75}{Ortho}&\rotatebox[origin=c]{75}{Acc}&\rotatebox[origin=c]{75}{Dis}&\rotatebox[origin=c]{75}{J$_{Div}$}&\rotatebox[origin=c]{75}{PWCCA}&\rotatebox[origin=c]{75}{CKA}&\rotatebox[origin=c]{75}{Ortho}&\rotatebox[origin=c]{75}{Acc}&\rotatebox[origin=c]{75}{Dis}&\rotatebox[origin=c]{75}{J$_{Div}$} \\
\midrule
\midrule
tau-$\kappa$ (median) $\uparrow$&0.32 & 0.33 & \textbf{0.34} & 0.12 & 0.13 & 0.16 & \textbf{0.48} & 0.44 & 0.27 & -0.14 & 0.23 & 0.26 & \textbf{0.27} & 0.14 & 0.17 & -0.25 & 0.25 & -0.12 & \textbf{0.28} & 0.07 & 0.16 & 0.09 & 0.15 & 0.22 & \textbf{0.25} & 0.12 & 0.10 & 0.03 & 0.13 & 0.17\\[0.5em]
tau-$\kappa$ ($1^{st}$-Q) $\uparrow$&0.30 & \textbf{0.32} & \textbf{0.32} & 0.07 & 0.07 & 0.10 & \textbf{0.46} & 0.39 & 0.21 & -0.18 & 0.19 & 0.19 & \textbf{0.20} & 0.11 & 0.15 & -0.30 & 0.18 & -0.18 & \textbf{0.23} & -0.03 & 0.14 & 0.00 & 0.08 & 0.06 & \textbf{0.19} & 0.02 & 0.03 & -0.02 & 0.07 & 0.05\\[0.5em]
tau-$\kappa$ ($3^{rd}$-Q) $\uparrow$&\textbf{0.36} & 0.34 & \textbf{0.36} & 0.19 & 0.18 & 0.20 & 0.51 & \textbf{0.52} & 0.31 & -0.01 & 0.27 & 0.29 & \textbf{0.33} & 0.27 & 0.29 & -0.20 & 0.29 & -0.08 & \textbf{0.29} & 0.16 & 0.27 & 0.21 & 0.20 & 0.29 &\textbf{ 0.28} & 0.18 & 0.14 & 0.13 & 0.15 & 0.24\\[0.5em] \\[-1em] \cline{1-31}\\[-0.8em]
$\#$ p-value $< 0.05$&\textbf{100}\% & \textbf{100}\% & \textbf{100}\% & 10\% & 20\% & 20\% & \textbf{100}\% & 90\% & 70\% & 20\% & 60\% & 70\% & 60\% & 30\% & 40\% & \textbf{70}\% & 60\% & 20\% & \textbf{90}\% & 20\% & 40\% & 30\% & 10\% & 50\% & \textbf{70}\% & 10\% & 10\% & 0\% & 0\% & 40\%\\[0.5em]
$\#$ p-value $< 0.1$&\textbf{100}\% & \textbf{100}\% & \textbf{100}\% & 30\% & 30\% & 30\% & \textbf{100}\% & \textbf{100}\% & 90\% & 30\% & 70\% & 70\% & \textbf{80}\% & 30\% & 40\% & \textbf{80}\% & 70\% & 30\% & \textbf{90}\% & 20\% & 40\% & 50\% & 30\% & 60\% & \textbf{70}\% & 30\% & 20\% & 10\% & 10\% & 50\%\\[0.5em]
\bottomrule
DNN&\multicolumn{6}{c}{ROBERTA}&\multicolumn{6}{c}{XLNet}&\multicolumn{6}{c}{BERT}&\multicolumn{6}{c}{DistilBERT}&\multicolumn{6}{c}{Electra} \\ \cline{2-31}
\multirow{2}{*}{Similarity} & &&&&&&&&&&&&&&&&&&&&&&&&&&&&& \\
($\mathcal{P}$)&\rotatebox[origin=c]{75}{PWCCA}&\rotatebox[origin=c]{75}{CKA}&\rotatebox[origin=c]{75}{Ortho}&\rotatebox[origin=c]{75}{Acc}&\rotatebox[origin=c]{75}{Dis}&\rotatebox[origin=c]{75}{J$_{Div}$}&\rotatebox[origin=c]{75}{PWCCA}&\rotatebox[origin=c]{75}{CKA}&\rotatebox[origin=c]{75}{Ortho}&\rotatebox[origin=c]{75}{Acc}&\rotatebox[origin=c]{75}{Dis}&\rotatebox[origin=c]{75}{J$_{Div}$}&\rotatebox[origin=c]{75}{PWCCA}&\rotatebox[origin=c]{75}{CKA}&\rotatebox[origin=c]{75}{Ortho}&\rotatebox[origin=c]{75}{Acc}&\rotatebox[origin=c]{75}{Dis}&\rotatebox[origin=c]{75}{J$_{Div}$}&\rotatebox[origin=c]{75}{PWCCA}&\rotatebox[origin=c]{75}{CKA}&\rotatebox[origin=c]{75}{Ortho}&\rotatebox[origin=c]{75}{Acc}&\rotatebox[origin=c]{75}{Dis}&\rotatebox[origin=c]{75}{J$_{Div}$}&\rotatebox[origin=c]{75}{PWCCA}&\rotatebox[origin=c]{75}{CKA}&\rotatebox[origin=c]{75}{Ortho}&\rotatebox[origin=c]{75}{Acc}&\rotatebox[origin=c]{75}{Dis}&\rotatebox[origin=c]{75}{J$_{Div}$} \\
\midrule
\midrule
tau-$\kappa$ (median) $\uparrow$&0.33 & 0.03 & -0.23 & 0.31 & 0.49 & \textbf{0.50} & -0.03 & \textbf{0.47} & 0.46 & 0.42 & 0.44 & 0.45 & 0.16 & 0.28 & 0.26 & 0.17 & 0.15 & \textbf{0.59} & 0.09 & 0.41 & 0.44 & 0.00 & 0.28 & 0.63 & 0.54 & 0.56 & 0.43 & 0.41 & 0.56 & \textbf{0.62}\\[0.5em]
tau-$\kappa$ ($1^{st}$-Q) $\uparrow$&0.27 & -0.06 & -0.34 & 0.09 & 0.40 & \textbf{0.41} & -0.05 & \textbf{0.43} & 0.36 & 0.37 & 0.41 & 0.41 & 0.12 & 0.22 & 0.21 & 0.02 & 0.11 & \textbf{0.52} & 0.08 & 0.37 & 0.37 & -0.03 & 0.18 & \textbf{0.59} & 0.50 & 0.52 & 0.41 & 0.39 & 0.54 & \textbf{0.60}\\[0.5em]
tau-$\kappa$ ($3^{rd}$-Q) $\uparrow$&0.38 & 0.09 & -0.17 & 0.34 & \textbf{0.58} & \textbf{0.58} & 0.05 & 0.54 & 0.51 & 0.48 & \textbf{0.57} & 0.50 & 0.19 & 0.29 & 0.28 & 0.30 & 0.16 & \textbf{0.64} & 0.13 & 0.44 & 0.47 & 0.08 & 0.38 & 0.66 & 0.58 & 0.61 & 0.48 & 0.43 & 0.59 & \textbf{0.67}\\[0.5em] \\[-1em] \cline{1-31}\\[-0.8em]
$\#$ p-value $< 0.05$&90\% & 0\% & 50\% & 60\% & \textbf{100}\% & \textbf{100}\% & 0\% & \textbf{100}\% & 80\% & 90\% & \textbf{100}\% & 90\% & 10\% & 70\% & 70\% & 40\% & 0\% & \textbf{100}\% & 0\% & 90\% & \textbf{100}\% & 10\% & 70\% & \textbf{100}\% & \textbf{100}\% & \textbf{100}\% & \textbf{100}\% & \textbf{100}\% & \textbf{100}\% & \textbf{100}\%\\[0.5em]
$\#$ p-value $< 0.1$&90\% & 0\% & 60\% & 80\% & \textbf{100}\% & \textbf{100}\% & 10\% & \textbf{100}\% & 80\% & 90\% & \textbf{100}\% & \textbf{100}\% & 40\% & 80\% & 80\% & 40\% & 10\% & \textbf{100}\% & 10\% & \textbf{100}\% & \textbf{100}\% & 10\% & 70\% & \textbf{100}\% & \textbf{100}\% & \textbf{100}\% & \textbf{100}\% & \textbf{100}\% & \textbf{100}\% & \textbf{100}\%\\[0.5em]
\bottomrule
\end{tabular}}
\label{tab:corr_faults}
\end{table}

\begin{table}
%\fontsize{15}{17}\selectfont
\caption{Correlation for each generation technique for each given DNN model being considered as a DNN model under test for $\mathcal{P}_O$ based on neuron coverage. For the same DNN model type (\eg \textit{Densenet100bc}), the results are given for all DNN model seeds. We highlight in bold the highest results. Generation techniques used are (top to bottom): \textit{Fuzz}-based, \textit{GAN}-based and \textit{Text}-based.}
\centering
\hspace*{-2em}
\resizebox{1.1\textwidth}{!}{
%\Huge
\begin{tabular}{c|*{6}{c}|*{6}{c}|*{6}{c}|*{6}{c}|*{6}{c}}
\toprule
DNN&\multicolumn{6}{c}{Densenet100bc}&\multicolumn{6}{c}{Preresnet110}&\multicolumn{6}{c}{Preresnet20}&\multicolumn{6}{c}{VGG19}&\multicolumn{6}{c}{VGG16} \\ \cline{2-31}
\multirow{2}{*}{Similarity} & &&&&&&&&&&&&&&&&&&&&&&&&&&&&& \\
($\mathcal{P}$)&\rotatebox[origin=c]{75}{PWCCA}&\rotatebox[origin=c]{75}{CKA}&\rotatebox[origin=c]{75}{Ortho}&\rotatebox[origin=c]{75}{Acc}&\rotatebox[origin=c]{75}{Dis}&\rotatebox[origin=c]{75}{J$_{Div}$}&\rotatebox[origin=c]{75}{PWCCA}&\rotatebox[origin=c]{75}{CKA}&\rotatebox[origin=c]{75}{Ortho}&\rotatebox[origin=c]{75}{Acc}&\rotatebox[origin=c]{75}{Dis}&\rotatebox[origin=c]{75}{J$_{Div}$}&\rotatebox[origin=c]{75}{PWCCA}&\rotatebox[origin=c]{75}{CKA}&\rotatebox[origin=c]{75}{Ortho}&\rotatebox[origin=c]{75}{Acc}&\rotatebox[origin=c]{75}{Dis}&\rotatebox[origin=c]{75}{J$_{Div}$}&\rotatebox[origin=c]{75}{PWCCA}&\rotatebox[origin=c]{75}{CKA}&\rotatebox[origin=c]{75}{Ortho}&\rotatebox[origin=c]{75}{Acc}&\rotatebox[origin=c]{75}{Dis}&\rotatebox[origin=c]{75}{J$_{Div}$}&\rotatebox[origin=c]{75}{PWCCA}&\rotatebox[origin=c]{75}{CKA}&\rotatebox[origin=c]{75}{Ortho}&\rotatebox[origin=c]{75}{Acc}&\rotatebox[origin=c]{75}{Dis}&\rotatebox[origin=c]{75}{J$_{Div}$} \\
\midrule
\midrule
tau-$\kappa$ (median) $\uparrow$&0.37 & 0.39 & 0.41 & 0.45 & 0.47 & \textbf{0.54} & 0.37 & 0.29 & 0.07 & 0.25 & 0.61 & \textbf{0.62} & 0.36 & 0.41 & \textbf{0.42} & -0.36 & 0.43 & -0.09 & \textbf{0.57} & 0.11 & 0.33 & 0.43 & 0.36 & 0.54 & \textbf{0.60} & 0.22 & 0.36 & 0.36 & 0.25 & 0.40\\[0.5em]
tau-$\kappa$ ($1^{st}$-Q) $\uparrow$&0.34 & 0.36 & 0.38 & 0.43 & 0.46 & \textbf{0.53} & 0.35 & 0.25 & 0.02 & 0.19 & 0.57 & \textbf{0.58} & 0.32 & 0.36 & \textbf{0.39} & -0.39 & \textbf{0.39} & -0.15 & \textbf{0.55} & 0.03 & 0.20 & 0.38 & 0.30 & 0.45 & \textbf{0.57} & 0.10 & 0.31 & 0.33 & 0.19 & 0.33\\[0.5em]
tau-$\kappa$ ($3^{rd}$-Q) $\uparrow$&0.38 & 0.40 & 0.42 & 0.46 & 0.48 & \textbf{0.55} & 0.39 & 0.31 & 0.10 & 0.33 & 0.64 & \textbf{0.67} & 0.41 & 0.45 & \textbf{0.49} & -0.30 & 0.47 & -0.05 & \textbf{0.63} & 0.16 & 0.36 & 0.49 & 0.41 & 0.56 & \textbf{0.63} & 0.27 & 0.41 & 0.41 & 0.29 & 0.49\\[0.5em] \\[-1em] \cline{1-31}\\[-0.8em]
$\#$ p-value $< 0.05$&\textbf{100\%} & \textbf{100\%} & \textbf{100\%} & \textbf{100\%} & \textbf{100\%} & \textbf{100\%} & 90\% & 80\% & 0\% & 60\% & \textbf{100\%} & \textbf{100\%} & 90\% & 90\% & 90\% & \textbf{100\%} & 90\% & 0\% & \textbf{100\%} & 20\% & 70\% & \textbf{100\%} & \textbf{100\%} & 90\% & \textbf{100\%} & 50\% & 90\% & \textbf{100\%} & 60\% & 90\%\\[0.5em]
$\#$ p-value $< 0.1$&\textbf{100\%} & \textbf{100\%} & \textbf{100\%} & \textbf{100\%} & \textbf{100\%} & \textbf{100\%} & 90\% & 80\% & 0\% & 80\% & \textbf{100\%} & \textbf{100\%} & 90\% & 90\% & 90\% & \textbf{100\%} & \textbf{100\%} & 0\% & \textbf{100\%} & 30\% & 70\% & \textbf{100\%} & \textbf{100\%} & \textbf{100\%} & \textbf{100\%} & 60\% & \textbf{100\%} & \textbf{100\%} & 80\% & 90\%\\[0.5em]
\bottomrule
DNN&\multicolumn{6}{c}{Densenet100bc}&\multicolumn{6}{c}{Preresnet110}&\multicolumn{6}{c}{Preresnet20}&\multicolumn{6}{c}{VGG19}&\multicolumn{6}{c}{VGG16} \\ \cline{2-31}
\multirow{2}{*}{Similarity} & &&&&&&&&&&&&&&&&&&&&&&&&&&&&& \\
($\mathcal{P}$)&\rotatebox[origin=c]{75}{PWCCA}&\rotatebox[origin=c]{75}{CKA}&\rotatebox[origin=c]{75}{Ortho}&\rotatebox[origin=c]{75}{Acc}&\rotatebox[origin=c]{75}{Dis}&\rotatebox[origin=c]{75}{J$_{Div}$}&\rotatebox[origin=c]{75}{PWCCA}&\rotatebox[origin=c]{75}{CKA}&\rotatebox[origin=c]{75}{Ortho}&\rotatebox[origin=c]{75}{Acc}&\rotatebox[origin=c]{75}{Dis}&\rotatebox[origin=c]{75}{J$_{Div}$}&\rotatebox[origin=c]{75}{PWCCA}&\rotatebox[origin=c]{75}{CKA}&\rotatebox[origin=c]{75}{Ortho}&\rotatebox[origin=c]{75}{Acc}&\rotatebox[origin=c]{75}{Dis}&\rotatebox[origin=c]{75}{J$_{Div}$}&\rotatebox[origin=c]{75}{PWCCA}&\rotatebox[origin=c]{75}{CKA}&\rotatebox[origin=c]{75}{Ortho}&\rotatebox[origin=c]{75}{Acc}&\rotatebox[origin=c]{75}{Dis}&\rotatebox[origin=c]{75}{J$_{Div}$}&\rotatebox[origin=c]{75}{PWCCA}&\rotatebox[origin=c]{75}{CKA}&\rotatebox[origin=c]{75}{Ortho}&\rotatebox[origin=c]{75}{Acc}&\rotatebox[origin=c]{75}{Dis}&\rotatebox[origin=c]{75}{J$_{Div}$} \\
\midrule
\midrule
tau-$\kappa$ (median) $\uparrow$&0.47 & 0.48 & \textbf{0.51} & 0.13 & 0.15 & 0.19 & \textbf{0.57} & 0.51 & 0.29 & -0.12 & 0.27 & 0.29 & \textbf{0.42} & 0.27 & 0.28 & -0.40 & 0.41 & -0.26 & \textbf{0.34} & 0.08 & 0.11 & 0.15 & 0.19 & 0.24 & \textbf{0.30} & 0.10 & 0.06 & 0.06 & 0.17 & 0.21\\[0.5em]
tau-$\kappa$ ($1^{st}$-Q) $\uparrow$&\textbf{0.46} & 0.44 & 0.47 & 0.10 & 0.11 & 0.16 & \textbf{0.54} & 0.46 & 0.27 & -0.14 & 0.25 & 0.28 & \textbf{0.37} & 0.20 & 0.24 & -0.42 & \textbf{0.37} & -0.27 & \textbf{0.25} & -0.02 & 0.09 & 0.11 & 0.15 & 0.12 & \textbf{0.28} & 0.06 & 0.04 & 0.02 & 0.15 & 0.16\\[0.5em]
tau-$\kappa$ ($3^{rd}$-Q) $\uparrow$&0.49 & 0.50 & \textbf{0.52} & 0.16 & 0.18 & 0.23 & \textbf{0.59} & 0.54 & 0.32 & -0.07 & 0.31 & 0.32 & \textbf{0.45} & 0.31 & 0.34 & -0.37 & \textbf{0.45} & -0.22 & \textbf{0.39} & 0.16 & 0.21 & 0.18 & 0.24 & 0.31 & \textbf{0.36} & 0.13 & 0.08 & 0.11 & 0.25 & 0.27\\[0.5em] \\[-1em] \cline{1-31}\\[-0.8em]
$\#$ p-value $< 0.05$&\textbf{100\%} & \textbf{100\%} & \textbf{100\%} & 10\% & 10\% & 30\% & \textbf{100\%} & \textbf{100\%} & 90\% & 0\% & \textbf{100\%} & \textbf{100\%} & \textbf{100\%} & 60\% & \textbf{100\%} & \textbf{100\%} & \textbf{100\%} & 70\% & \textbf{80\%} & 20\% & 30\% & 20\% & 40\% & 70\% & \textbf{100\%} & 0\% & 0\% & 0\% & 40\% & 40\%\\[0.5em]
$\#$ p-value $< 0.1$&\textbf{100\%} & \textbf{100\%} & \textbf{100\%} & 20\% & 20\% & 50\% & \textbf{100\%} & \textbf{100\%} & \textbf{100\%} & 10\% & \textbf{100\%} & \textbf{100\%} & \textbf{100\%} & 90\% & \textbf{100\%} & \textbf{100\%} & \textbf{100\%} & 80\% & \textbf{90\%} & 30\% & 40\% & 40\% & 50\% & 80\% & \textbf{100\%} & 0\% & 0\% & 0\% & 40\% & 70\%\\[0.5em]
\bottomrule
DNN&\multicolumn{6}{c}{ROBERTA}&\multicolumn{6}{c}{XLNet}&\multicolumn{6}{c}{BERT}&\multicolumn{6}{c}{DistilBERT}&\multicolumn{6}{c}{Electra} \\ \cline{2-31}
\multirow{2}{*}{Similarity} & &&&&&&&&&&&&&&&&&&&&&&&&&&&&& \\
($\mathcal{P}$)&\rotatebox[origin=c]{75}{PWCCA}&\rotatebox[origin=c]{75}{CKA}&\rotatebox[origin=c]{75}{Ortho}&\rotatebox[origin=c]{75}{Acc}&\rotatebox[origin=c]{75}{Dis}&\rotatebox[origin=c]{75}{J$_{Div}$}&\rotatebox[origin=c]{75}{PWCCA}&\rotatebox[origin=c]{75}{CKA}&\rotatebox[origin=c]{75}{Ortho}&\rotatebox[origin=c]{75}{Acc}&\rotatebox[origin=c]{75}{Dis}&\rotatebox[origin=c]{75}{J$_{Div}$}&\rotatebox[origin=c]{75}{PWCCA}&\rotatebox[origin=c]{75}{CKA}&\rotatebox[origin=c]{75}{Ortho}&\rotatebox[origin=c]{75}{Acc}&\rotatebox[origin=c]{75}{Dis}&\rotatebox[origin=c]{75}{J$_{Div}$}&\rotatebox[origin=c]{75}{PWCCA}&\rotatebox[origin=c]{75}{CKA}&\rotatebox[origin=c]{75}{Ortho}&\rotatebox[origin=c]{75}{Acc}&\rotatebox[origin=c]{75}{Dis}&\rotatebox[origin=c]{75}{J$_{Div}$}&\rotatebox[origin=c]{75}{PWCCA}&\rotatebox[origin=c]{75}{CKA}&\rotatebox[origin=c]{75}{Ortho}&\rotatebox[origin=c]{75}{Acc}&\rotatebox[origin=c]{75}{Dis}&\rotatebox[origin=c]{75}{J$_{Div}$} \\
\midrule
\midrule
tau-$\kappa$ (median) $\uparrow$&0.26 & -0.03 & -0.35 & 0.25 & 0.46 & \textbf{0.47} & -0.10 & 0.38 & 0.33 & 0.36 & 0.37 & \textbf{0.39} & 0.13 & 0.13 & 0.10 & 0.11 & 0.14 & \textbf{0.45} & 0.06 & 0.28 & 0.30 & -0.07 & 0.23 & \textbf{0.55} & 0.52 & 0.49 & 0.41 & 0.33 & 0.53 & \textbf{0.55}\\[0.5em]
tau-$\kappa$ ($1^{st}$-Q) $\uparrow$&0.20 & -0.09 & -0.38 & 0.07 & \textbf{0.42} & 0.41 & -0.17 & \textbf{0.32} & 0.26 & 0.25 & 0.29 & 0.27 & 0.11 & 0.11 & 0.07 & 0.00 & 0.12 & \textbf{0.40} & -0.01 & 0.26 & 0.27 & -0.10 & 0.08 & \textbf{0.49} & 0.47 & 0.46 & 0.38 & 0.28 & 0.51 & \textbf{0.53}\\[0.5em]
tau-$\kappa$ ($3^{rd}$-Q) $\uparrow$&0.31 & 0.02 & -0.32 & 0.35 & 0.49 & \textbf{0.51} & -0.05 & \textbf{0.45} & 0.44 & 0.42 & 0.42 & 0.44 & 0.16 & 0.15 & 0.14 & 0.32 & 0.16 & \textbf{0.50} & 0.10 & 0.31 & 0.34 & 0.02 & 0.30 & \textbf{0.57} & \textbf{0.57} & 0.51 & 0.43 & 0.33 & 0.55 & 0.55\\[0.5em] \\[-1em] \cline{1-31}\\[-0.8em]
$\#$ p-value $< 0.05$&50\% & 0\% & 90\% & 60\% & \textbf{100}\% & \textbf{100}\% & 10\% & \textbf{90}\% & 80\% & \textbf{90}\% & \textbf{90}\% & \textbf{90}\% & 10\% & 10\% & 0\% & 40\% & 0\% & \textbf{100}\% & 0\% & \textbf{100}\% & \textbf{100}\% & 10\% & 50\% & \textbf{100}\% & \textbf{100}\% & \textbf{100}\% & \textbf{100}\% & 90\% & \textbf{100}\% & \textbf{100}\%\\[0.5em]
$\#$ p-value $< 0.1$&80\% & 0\% & 90\% & 60\% & \textbf{100}\% & \textbf{100}\% & 30\% & \textbf{100}\% & 80\% & 90\% & \textbf{100}\% & 90\% & 10\% & 20\% & 10\% & 40\% & 10\% & \textbf{100}\% & 10\% & \textbf{100}\% & \textbf{100}\% & 10\% & 60\% & \textbf{100}\% & \textbf{100}\% & \textbf{100}\% & \textbf{100}\% & 90\% & \textbf{100}\% & \textbf{100}\%\\[0.5em]
\bottomrule
\end{tabular}}
\label{tab:corr_kmnc}
\end{table}

First, we observe that no metric is overall the best for all generation techniques and all DNN models under test. As such, there does not seem to be a \enquote{one-metric fit-all} similarity that could be used in every situation. This was nonetheless to be expected as, analogically to the \textit{No Free Lunch} theorem, it is not likely that one metric can outperform all others in every situation. Especially when different generation techniques (and so different input types) are used and that the similarity focuses on particular invariance. Note that the DNN model seed can have its importance on the results, as illustrated by the variation for the same metric of the number of significant (strength of the) correlations. This illustrates the need to consider multiple DNN model seeds when doing this sort of study to quantify the effect said seed can have on the results. We also see that results drastically differ depending on the metric that is used and it's rare that all metrics lead to all correlations being significant.

However, it's possible to find, for each generation technique, one similarity that works reasonably well across all the selected DNN models under test. Regarding Fault Types based $\mathcal{P}_O$, $J_{Div}$ is the best overall for the \textit{Text}-based generation technique, with medium/strong correlations ($>0.4$) which are significant almost across all DNN model seeds at the $0.05$ threshold. For the \textit{GAN}-based generation technique, it's instead $PWCCA$ which seems to be preferable. One can see in that case that the correlation strength largely varies across DNN models under tests (median around $\sim 0.25$ - $0.40$). For the \textit{Fuzz}-based generation technique, the choice is between $PWCCA$ and $J_{Div}$: while the first one gives good results on 4 types of object DNN models (median correlation $\sim 0.4$ - $0.5$, almost all being significant), it performs relatively badly on \textit{PreResnet20} (negative correlation). On the contrary, $J_{Div}$ leads to a weaker correlation (median correlation $\sim 0.25$ - $0.5$, with a lesser number of one being significant), but is more consistent across DNN models. 

Regarding the neuron-based $\mathcal{P}_O$, similar results can be observed for the \textit{Text}-based and the \textit{GAN}-based generation techniques, with respectively $J_{Div}$ (medium/strong correlations $>0.35$) and $PWCCA$ (medium correlations $>0.3$). We note that the strength of the correlation has followed opposing trends: while for the \textit{Text}-based generation technique the correlations have become relatively weaker from the neuron-based coverage, it has gotten stronger on \textit{GAN}-based generation technique for the same type of DNN model under test. The main difference between the two properties is visible with the \textit{Fuzz}-based generation technique: for neuron-coverage based $\mathcal{P}_O$ it is $PWCCA$ which works consistently well across DNN model types (median around $\sim 0.35 - 0.6$), while $J_{Div}$ works well across most DNN model types except $Preresnet20$. $Dis$ would also be a reasonable similarity in that case as it works well on all DNN model types except $VGG16$. The obtained results for both properties match observed patterns in RQ1.

\begin{tcolorbox}[colback=blue!5,colframe=blue!40!black]
\textbf{Findings 2:} There is not one single effective similarity for all DNN models under tests and generation techniques. However, for each technique, we can find one similarity that leads to a positive significant correlation with each property across all DNN models under tests, for a majority of DNN model seeds used. Results can be impacted by the DNN model seed choice illustrating the need to consider this aspect in the process. $PWCCA$ and $J_{Div}$ seem to be the main similarities that achieve good significant correlations across our experiments. In particular, for both properties (fault type and neuron based), $PWCCA$ works best for \textit{GAN}-based generation technique and $J_{Div}$ works best for \textit{Text}-based generation technique. Both metrics work reasonably well for \textit{Fuzz}-based generation technique. Those results confirm the preliminary results of RQ1. 
\end{tcolorbox}

\subsection{RQ3: \rqthree}\label{sec:rq3}

We now want to verify if the most relevant reference test set in terms of property covered (\ie fault-type coverage and neuron coverage) can be decided from the similarity and how the DNN model seeds affect this outcome. To do so we will use two criteria: \textit{Top-1} and \textit{Top-5} which aim to compare the property coverage of the chosen test set against a randomly picked test set. Those criteria are inspired by the \textit{Top-1} and \textit{Top-5} accuracy metrics used in classification or information retrieval \cite{Li20} and aim to show the relevance of a chosen test set in comparison to other test sets in terms of the property to transfer. As we have shown in the previous RQ we have several significant correlations. Thus, the most similar DNN model according to the chosen representational/functional similarity should lead to the most covering test set for the given property. Thus, as a criterion, we can choose the test set of the most similar reference DNN model to be transferred in practice. To evaluate the effectiveness of the criterion, we compare the property coverage of the chosen test set to the coverage of the remaining test sets, as if we were choosing at random. Basically, we count how many other test sets have covered more of the property compared to the selected test set. This is the \textit{Top-1} metric. However, we have seen that the DNN model seed can affect the results so we also want to assess how the result would differ if the most similar DNN model was not selectable (for instance, because we did not train such an instance). This is done to assess the robustness of the method against seed selection. As such, we iteratively repeat the same procedure as the \textit{Top-1} metric while removing the first $k$ (with $k$ iteratively being $\{1, 2, 3, 4\}$) best DNN models from the choice and choosing the $k+1$ to become the best. This is the \textit{Top-5} metric. Similarly, as in the previous research question, we compute the result for each DNN model seed as the DNN model under test and we show the median, first, and third quartiles. Finally, we also provide the averaged value of $\mathcal{P}_O$ for each property across the DNN model seeds of the DNN model under test for the \textit{Top-1} and \textit{Top-5} metrics. Results are presented in Table \ref{tab:rand_above} for the fault-type property and in Table \ref{tab:rand_above_kmnc} for the neuron property.

\begin{table}[]
%\fontsize{15}{17}\selectfont
\caption{For fault-type based property, \textit{Top-1} and \textit{Top-5} metric for each generation technique, for each given DNN model seed being considered as a DNN model under test. For the same DNN model type (\eg \textit{Densenet100bc}), the results are given for all DNN model seeds. We also show the averaged value of fault-type coverage property for both \textit{Top-1} and \textit{Top-5}. We highlight in bold the best scores. Generation techniques used are (top to bottom): \textit{Fuzz}-based, \textit{GAN}-based and \textit{Text}-based.}
\centering
\hspace*{-2em}
\resizebox{1.1\textwidth}{!}{
\large
\begin{tabular}{c|*{6}{c}|*{6}{c}|*{6}{c}|*{6}{c}|*{6}{c}}
\toprule
DNN&\multicolumn{6}{c}{Densenet100bc}&\multicolumn{6}{c}{Preresnet110}&\multicolumn{6}{c}{Preresnet20}&\multicolumn{6}{c}{VGG19}&\multicolumn{6}{c}{VGG16} \\ \cline{2-31}
\multirow{2}{*}{Similarity} & &&&&&&&&&&&&&&&&&&&&&&&&&&&&& \\
($\mathcal{P}$)&\rotatebox[origin=c]{75}{PWCCA}&\rotatebox[origin=c]{75}{CKA}&\rotatebox[origin=c]{75}{Ortho}&\rotatebox[origin=c]{75}{Acc}&\rotatebox[origin=c]{75}{Dis}&\rotatebox[origin=c]{75}{J$_{Div}$}&\rotatebox[origin=c]{75}{PWCCA}&\rotatebox[origin=c]{75}{CKA}&\rotatebox[origin=c]{75}{Ortho}&\rotatebox[origin=c]{75}{Acc}&\rotatebox[origin=c]{75}{Dis}&\rotatebox[origin=c]{75}{J$_{Div}$}&\rotatebox[origin=c]{75}{PWCCA}&\rotatebox[origin=c]{75}{CKA}&\rotatebox[origin=c]{75}{Ortho}&\rotatebox[origin=c]{75}{Acc}&\rotatebox[origin=c]{75}{Dis}&\rotatebox[origin=c]{75}{J$_{Div}$}&\rotatebox[origin=c]{75}{PWCCA}&\rotatebox[origin=c]{75}{CKA}&\rotatebox[origin=c]{75}{Ortho}&\rotatebox[origin=c]{75}{Acc}&\rotatebox[origin=c]{75}{Dis}&\rotatebox[origin=c]{75}{J$_{Div}$}&\rotatebox[origin=c]{75}{PWCCA}&\rotatebox[origin=c]{75}{CKA}&\rotatebox[origin=c]{75}{Ortho}&\rotatebox[origin=c]{75}{Acc}&\rotatebox[origin=c]{75}{Dis}&\rotatebox[origin=c]{75}{J$_{Div}$} \\
\midrule
\midrule
Top-1 (median) $\downarrow$&11\%&5\%&\textbf{4\%}&54\%&54\%&26\%&\textbf{6\%}&16\%&30\%&51\%&29\%&15\%&62\%&17\%&\textbf{11\%}&88\%&70\%&53\%&\textbf{9\%}&12\%&11\%&11\%&40\%&11\%&18\%&10\%&15\%&7\%&11\%&\textbf{5\%}\\[0.1em]
Top-1 ($1^{st}$-Q) $\downarrow$&6\%&\textbf{3\%}&\textbf{3\%}&50\%&50\%&21\%&\textbf{1\%}&8\%&22\%&33\%&16\%&8\%&48\%&7\%&\textbf{6\%}&66\%&63\%&38\%&\textbf{3\%}&11\%&8\%&5\%&36\%&4\%&14\%&6\%&10\%&\textbf{3\%}&8\%&4\% \\[0.1em]
Top-1 ($3^{rd}$-Q) $\downarrow$&18\%&12\%&\textbf{10\%}&64\%&64\%&48\%&\textbf{9\%}&42\%&63\%&60\%&34\%&36\%&77\%&32\%&\textbf{28\%}&97\%&88\%&68\%&\textbf{14\%}&24\%&20\%&28\%&67\%&19\%&21\%&18\%&23\%&35\%&26\%& \textbf{16\%}\\[0.1em]
Top-1 Value $\uparrow$&33.86\% & 35.11\% & \textbf{35.33}\% & 27.20\% & 27.20\% & 30.17\% & \textbf{36.88}\% & 34.11\% & 32.36\% & 31.94\% & 33.33\% & 34.36\% & 31.08\% & 33.80\% & \textbf{34.38}\% & 28.07\% & 29.52\% & 31.29\% & 45.88\% & 45.52\% & 45.83\% & 43.39\% & 39.08\% & \textbf{46.33}\% & 42.45\% & 40.30\% & 44.14\% & 38.91\% & 35.33\% & \textbf{44.35}\%\\[0.1em] \\[-1em] \cline{1-31}\\[-1em]
Top-5 (median) $\downarrow$&11\%&\textbf{9\%}&11\%&47\%&40\%&24\%&\textbf{13\%}&26\%&40\%&46\%&21\%&\textbf{13\%}&70\%&\textbf{27\%}&31\%&56\%&75\%&40\%&10\%&11\%&11\%&29\%&43\%&\textbf{9\%}&30\%&\textbf{20\%}&35\%&30\%&28\%&\textbf{20\%}\\[0.1em]
Top-5 ($1^{st}$-Q) $\downarrow$&10\%&\textbf{8\%}&\textbf{8\%}&40\%&36\%&20\%&\textbf{10\%}&21\%&37\%&40\%&17\%&\textbf{10\%}&65\%&24\%&\textbf{23\%}&50\%&68\%&32\%&9\%&\textbf{8\%}&\textbf{8\%}&18\%&38\%&\textbf{8\%}&27\%&14\%&23\%&21\%&22\%&\textbf{10\%} \\[0.1em]
Top-5 ($3^{rd}$-Q) $\downarrow$&13\%&\textbf{10\%}&14\%&53\%&43\%&29\%&\textbf{19\%}&40\%&50\%&48\%&27\%&25\%&76\%&\textbf{35\%}&36\%&60\%&81\%&45\%&\textbf{13\%}&14\%&\textbf{13\%}&39\%&49\%&36\%&37\%&31\%&51\%&49\%&31\%&\textbf{24\%} \\[0.1em]
Top-5 Value $\uparrow$&33.54\% & \textbf{33.95}\% & 33.53\% & 29.35\% & 29.61\% & 32.80\% & \textbf{35.14}\% & 33.66\% & 32.02\% & 31.58\% & 34.41\% & 34.40\% & 29.82\% & \textbf{33.36}\% & 33.23\% & 31.42\% & 29.50\% & 32.68\% & 45.58\% & \textbf{45.81}\% & 46.79\% & 40.93\% & 40.10\% & 43.75\% & \textbf{43.90}\% & 39.62\% & 43.54\% & 37.74\% & 36.14\% & 40.17\%\\[0.1em]
\bottomrule
DNN&\multicolumn{6}{c}{Densenet100bc}&\multicolumn{6}{c}{Preresnet110}&\multicolumn{6}{c}{Preresnet20}&\multicolumn{6}{c}{VGG19}&\multicolumn{6}{c}{VGG16} \\ \cline{2-31}
\multirow{2}{*}{Similarity} & &&&&&&&&&&&&&&&&&&&&&&&&&&&&& \\
($\mathcal{P}$)&\rotatebox[origin=c]{75}{PWCCA}&\rotatebox[origin=c]{75}{CKA}&\rotatebox[origin=c]{75}{Ortho}&\rotatebox[origin=c]{75}{Acc}&\rotatebox[origin=c]{75}{Dis}&\rotatebox[origin=c]{75}{J$_{Div}$}&\rotatebox[origin=c]{75}{PWCCA}&\rotatebox[origin=c]{75}{CKA}&\rotatebox[origin=c]{75}{Ortho}&\rotatebox[origin=c]{75}{Acc}&\rotatebox[origin=c]{75}{Dis}&\rotatebox[origin=c]{75}{J$_{Div}$}&\rotatebox[origin=c]{75}{PWCCA}&\rotatebox[origin=c]{75}{CKA}&\rotatebox[origin=c]{75}{Ortho}&\rotatebox[origin=c]{75}{Acc}&\rotatebox[origin=c]{75}{Dis}&\rotatebox[origin=c]{75}{J$_{Div}$}&\rotatebox[origin=c]{75}{PWCCA}&\rotatebox[origin=c]{75}{CKA}&\rotatebox[origin=c]{75}{Ortho}&\rotatebox[origin=c]{75}{Acc}&\rotatebox[origin=c]{75}{Dis}&\rotatebox[origin=c]{75}{J$_{Div}$}&\rotatebox[origin=c]{75}{PWCCA}&\rotatebox[origin=c]{75}{CKA}&\rotatebox[origin=c]{75}{Ortho}&\rotatebox[origin=c]{75}{Acc}&\rotatebox[origin=c]{75}{Dis}&\rotatebox[origin=c]{75}{J$_{Div}$} \\
\midrule
\midrule
Top-1 (median) $\downarrow$&\textbf{10\%}&\textbf{10\%}&11\%&69\%&69\%&\textbf{10\%}&\textbf{0\%}&12\%&39\%&72\%&23\%&6\%&\textbf{17\%}&30\%&33\%&90\%&31\%&57\%&\textbf{10\%}&16\%&11\%&31\%&62\%&31\%&35\%&35\%&\textbf{3\%}&55\%&26\%&10\%\\[0.1em]
Top-1 ($1^{st}$-Q) $\downarrow$&\textbf{1\%}&3\%&3\%&66\%&66\%&3\%&\textbf{0\%}&11\%&19\%&61\%&17\%&1\%&\textbf{4\%}&18\%&18\%&86\%&13\%&36\%&\textbf{2\%}&10\%&6\%&23\%&39\%&12\%&26\%&16\%&\textbf{1\%}&31\%&9\%&8\% \\[0.1em]
Top-1 ($3^{rd}$-Q) $\downarrow$&\textbf{19\%}&\textbf{19\%}&24\%&75\%&75\%&62\%&\textbf{5\%}&28\%&51\%&79\%&27\%&18\%&\textbf{19\%}&46\%&68\%&97\%&42\%&64\%&\textbf{20\%}&26\%&\textbf{20\%}&79\%&78\%&57\%&51\%&64\%&29\%&69\%&57\%&\textbf{19\%} \\[0.1em] 
Top-1 Value $\uparrow$&\textbf{26.56}\% & 26.25\% & 26.09\% & 21.42\% & 21.42\% & 25.41\% & \textbf{25.12}\% & 23.35\% & 21.78\% & 19.33\% & 22.33\% & 24.36\% & \textbf{22.66}\% & 20.97\% & 20.69\% & 17.11\% & 21.85\% & 20.32\% & 35.40\% & 35.03\% & \textbf{35.99}\% & 32.29\% & 32.15\% & 34.15\% & 32.67\% & 32.58\% & \textbf{34.68}\% & 31.19\% & 33.50\% & 34.64\%\\[0.1em] \\[-1em] \cline{1-31}\\[-1em]
Top-5 (median) $\downarrow$&\textbf{13\%}&23\%&17\%&44\%&41\%&36\%&\textbf{13\%}&17\%&40\%&67\%&17\%&14\%&33\%&33\%&\textbf{31\%}&64\%&42\%&55\%&\textbf{20\%}&27\%&25\%&41\%&49\%&29\%&\textbf{22\%}&42\%&\textbf{22\%}&47\%&39\%&31\%\\[0.1em]
Top-5 ($1^{st}$-Q) $\downarrow$&\textbf{10\%}&19\%&14\%&37\%&37\%&32\%&12\%&11\%&34\%&65\%&15\%&\textbf{8\%}&\textbf{25\%}&30\%&29\%&60\%&31\%&52\%&\textbf{15\%}&18\%&16\%&40\%&44\%&23\%&16\%&23\%&\textbf{12\%}&44\%&29\%&28\% \\[0.25em]
Top-5 ($3^{rd}$-Q) $\downarrow$&\textbf{23\%}&28\%&24\%&46\%&46\%&42\%&\textbf{15\%}&32\%&43\%&78\%&21\%&18\%&\textbf{34\%}&44\%&41\%&71\%&52\%&59\%&\textbf{28\%}&29\%&\textbf{28\%}&56\%&54\%&41\%&39\%&56\%&\textbf{31\%}&56\%&45\%&36\% \\[0.1em]
Top-5 Value $\uparrow$&\textbf{25.99}\% & 25.12\% & 25.59\% & 22.85\% & 22.76\% & 24.05\% & \textbf{23.42}\% & 22.97\% & 21.33\% & 19.19\% & 23.21\% & 23.14\% & \textbf{21.55}\% & 21.16\% & 21.33\% & 18.68\% & 20.99\% & 20.00\% & \textbf{34.76}\% & 34.55\% & 34.66\% & 32.98\% & 32.50\% & 34.16\% & 33.74\% & 32.55\% & \textbf{34.28}\% & 31.29\% & 32.61\% & 32.81\%\\[0.1em]
\bottomrule
DNN&\multicolumn{6}{c}{ROBERTA}&\multicolumn{6}{c}{XLNet}&\multicolumn{6}{c}{BERT}&\multicolumn{6}{c}{DistilBERT}&\multicolumn{6}{c}{Electra} \\ \cline{2-31}
\multirow{2}{*}{Similarity} & &&&&&&&&&&&&&&&&&&&&&&&&&&&&& \\
($\mathcal{P}$)&\rotatebox[origin=c]{75}{PWCCA}&\rotatebox[origin=c]{75}{CKA}&\rotatebox[origin=c]{75}{Ortho}&\rotatebox[origin=c]{75}{Acc}&\rotatebox[origin=c]{75}{Dis}&\rotatebox[origin=c]{75}{J$_{Div}$}&\rotatebox[origin=c]{75}{PWCCA}&\rotatebox[origin=c]{75}{CKA}&\rotatebox[origin=c]{75}{Ortho}&\rotatebox[origin=c]{75}{Acc}&\rotatebox[origin=c]{75}{Dis}&\rotatebox[origin=c]{75}{J$_{Div}$}&\rotatebox[origin=c]{75}{PWCCA}&\rotatebox[origin=c]{75}{CKA}&\rotatebox[origin=c]{75}{Ortho}&\rotatebox[origin=c]{75}{Acc}&\rotatebox[origin=c]{75}{Dis}&\rotatebox[origin=c]{75}{J$_{Div}$}&\rotatebox[origin=c]{75}{PWCCA}&\rotatebox[origin=c]{75}{CKA}&\rotatebox[origin=c]{75}{Ortho}&\rotatebox[origin=c]{75}{Acc}&\rotatebox[origin=c]{75}{Dis}&\rotatebox[origin=c]{75}{J$_{Div}$}&\rotatebox[origin=c]{75}{PWCCA}&\rotatebox[origin=c]{75}{CKA}&\rotatebox[origin=c]{75}{Ortho}&\rotatebox[origin=c]{75}{Acc}&\rotatebox[origin=c]{75}{Dis}&\rotatebox[origin=c]{75}{J$_{Div}$} \\
\midrule
\midrule
Top-1 (median) $\downarrow$&\textbf{9\%}&34\%&33\%&28\%&11\%&\textbf{9\%}&40\%&\textbf{3\%}&7\%&15\%&\textbf{3\%}&14\%&15\%&\textbf{9\%}&14\%&44\%&11\%&10\%&10\%&19\%&\textbf{9\%}&53\%&11\%&12\%&\textbf{4\%}&17\%&24\%&34\%&17\%&7\%\\[0.1em]
Top-1 ($1^{st}$-Q) $\downarrow$&\textbf{3\%}&25\%&30\%&15\%&4\%&\textbf{3\%}&21\%&\textbf{3\%}&\textbf{3\%}&11\%&\textbf{3\%}&5\%&3\%&\textbf{1\%}&8\%&33\%&7\%&6\%&4\%&11\%&\textbf{3\%}&18\%&8\%&6\%&\textbf{3\%}&7\%&17\%&20\%&8\%&\textbf{3\%} \\[0.1em]
Top-1 ($3^{rd}$-Q) $\downarrow$&\textbf{15\%}&52\%&47\%&38\%&21\%&\textbf{15\%}&62\%&\textbf{5\%}&12\%&33\%&\textbf{5\%}&17\%&17\%&\textbf{10\%}&19\%&72\%&19\%&12\%&16\%&24\%&\textbf{9\%}&77\%&12\%&12\%&22\%&24\%&33\%&39\%&29\%&\textbf{10\%} \\[0.1em]
Top-1 Value $\uparrow$&\textbf{35.97}\% & 27.15\% & 25.89\% & 29.49\% & 35.73\% & \textbf{35.97}\% & 27.64\% & \textbf{35.56}\% & 34.42\% & 31.01\% & \textbf{35.56}\% & 33.23\% & 37.45\% & \textbf{40.60}\% & 37.51\% & 26.47\% & 38.22\% & \textbf{39.58}\% & 37.36\% & 36.04\% & 38.82\% & 27.26\% & 37.41\% & \textbf{37.54}\% & 35.35\% & 34.18\% & 31.94\% & 30.45\% & 34.26\% & \textbf{36.14}\%\\[0.1em] \\[-1em] \cline{1-31}\\[-1em]
Top-5 (median) $\downarrow$&22\%&40\%&45\%&23\%&10\%&\textbf{9\%}&49\%&\textbf{8\%}&12\%&13\%&\textbf{8\%}&\textbf{8\%}&10\%&\textbf{9\%}&12\%&38\%&11\%&\textbf{9\%}&\textbf{8\%}&10\%&\textbf{8\%}&40\%&10\%&11\%&\textbf{11\%}&14\%&20\%&37\%&14\%&\textbf{11\%}\\[0.1em]
Top-5 ($1^{st}$-Q) $\downarrow$&18\%&33\%&27\%&14\%&8\%&\textbf{7\%}&40\%&\textbf{6\%}&10\%&10\%&\textbf{6\%}&\textbf{6\%}&9\%&9\%&10\%&27\%&\textbf{8\%}&9\%&7\%&7\%&\textbf{5\%}&37\%&7\%&8\%&\textbf{7\%}&8\%&14\%&35\%&10\%&9\% \\[0.1em]
Top-5 ($3^{rd}$-Q) $\downarrow$&25\%&48\%&55\%&41\%&13\%&\textbf{10\%}&57\%&\textbf{11\%}&14\%&22\%&12\%&\textbf{11\%}&11\%&10\%&13\%&39\%&12\%&\textbf{9\%}&\textbf{10\%}&12\%&\textbf{10\%}&45\%&12\%&16\%&17\%&17\%&31\%&41\%&19\%&\textbf{16\%} \\[0.1em]
Top-5 Value $\uparrow$&31.53\% & 26.24\% & 24.82\% & 29.58\% & \textbf{35.22}\% & 35.06\% & 25.65\% & \textbf{33.53}\% & 31.96\% & 31.30\% & 33.28\% & 33.04\% & 38.53\% & \textbf{38.91}\% & 37.95\% & 31.02\% & 38.41\% & 38.45\% & \textbf{37.67}\% & 37.25\% & 37.91\% & 28.95\% & 37.31\% & 36.83\% & \textbf{35.34}\% & 35.06\% & 33.47\% & 29.26\% & 34.44\% & 35.21\%\\[0.1em]
\bottomrule
\end{tabular}}
\label{tab:rand_above}
\end{table}

\begin{table}[]
%\fontsize{15}{17}\selectfont
\caption{For neuron-based property, \textit{Top-1} and \textit{Top-5} metric for each generation technique, for each given DNN model seed being considered as a DNN model under test. For the same DNN model type (\eg \textit{Densenet100bc}), the results are given for all DNN model seeds. We also show the averaged value of fault-type coverage property for both \textit{Top-1} and \textit{Top-5}. We highlight in bold the best scores. Generation techniques used are (top to bottom): \textit{Fuzz}-based, \textit{GAN}-based and \textit{Text}-based.}
\centering
\hspace*{-2em}
\resizebox{1.1\textwidth}{!}{
\large
\begin{tabular}{c|*{6}{c}|*{6}{c}|*{6}{c}|*{6}{c}|*{6}{c}}
\toprule
DNN&\multicolumn{6}{c}{Densenet100bc}&\multicolumn{6}{c}{Preresnet110}&\multicolumn{6}{c}{Preresnet20}&\multicolumn{6}{c}{VGG19}&\multicolumn{6}{c}{VGG16} \\ \cline{2-31}
\multirow{2}{*}{Similarity} & &&&&&&&&&&&&&&&&&&&&&&&&&&&&& \\
($\mathcal{P}$)&\rotatebox[origin=c]{75}{PWCCA}&\rotatebox[origin=c]{75}{CKA}&\rotatebox[origin=c]{75}{Ortho}&\rotatebox[origin=c]{75}{Acc}&\rotatebox[origin=c]{75}{Dis}&\rotatebox[origin=c]{75}{J$_{Div}$}&\rotatebox[origin=c]{75}{PWCCA}&\rotatebox[origin=c]{75}{CKA}&\rotatebox[origin=c]{75}{Ortho}&\rotatebox[origin=c]{75}{Acc}&\rotatebox[origin=c]{75}{Dis}&\rotatebox[origin=c]{75}{J$_{Div}$}&\rotatebox[origin=c]{75}{PWCCA}&\rotatebox[origin=c]{75}{CKA}&\rotatebox[origin=c]{75}{Ortho}&\rotatebox[origin=c]{75}{Acc}&\rotatebox[origin=c]{75}{Dis}&\rotatebox[origin=c]{75}{J$_{Div}$}&\rotatebox[origin=c]{75}{PWCCA}&\rotatebox[origin=c]{75}{CKA}&\rotatebox[origin=c]{75}{Ortho}&\rotatebox[origin=c]{75}{Acc}&\rotatebox[origin=c]{75}{Dis}&\rotatebox[origin=c]{75}{J$_{Div}$}&\rotatebox[origin=c]{75}{PWCCA}&\rotatebox[origin=c]{75}{CKA}&\rotatebox[origin=c]{75}{Ortho}&\rotatebox[origin=c]{75}{Acc}&\rotatebox[origin=c]{75}{Dis}&\rotatebox[origin=c]{75}{J$_{Div}$} \\
\midrule
\midrule
Top-1 (median) $\downarrow$&11\% & \textbf{8}\% & 9\% & 41\% & 41\% & 22\% & \textbf{1}\% & 31\% & 60\% & 34\% & 15\% & 9\% & \textbf{10}\% & 26\% & 25\% & 98\% & 41\% & 91\% & 12\% & 20\% & 22\% & \textbf{8}\% & 51\% & 11\% & 15\% & 24\% & 4\% & 25\% & 49\% & \textbf{2}\%\\[0.5em]
Top-1 ($1^{st}$-Q) $\downarrow$&\textbf{8}\% & \textbf{8}\% & \textbf{8}\% & 40\% & 40\% & 20\% & \textbf{0}\% & 11\% & 56\% & 23\% & 10\% & 3\% & \textbf{8}\% & 25\% & 16\% & 96\% & 21\% & 86\% & 8\% & 11\% & 11\% & \textbf{2}\% & 30\% & 6\% & 12\% & 8\% & \textbf{1}\% & 20\% & 40\% & \textbf{1}\%\\[0.5em]
Top-1 ($3^{rd}$-Q) $\downarrow$&19\% & \textbf{9}\% & 14\% & 49\% & 49\% & 39\% & \textbf{10}\% & 49\% & 67\% & 43\% & 19\% & 15\% & 36\% & \textbf{34}\% & \textbf{34}\% & 98\% & 61\% & 94\% & \textbf{19}\% & 25\% & 27\% & 26\% & 54\% & 21\% & 28\% & 41\% & \textbf{5}\% & 52\% & 56\% & 9\%\\[0.5em] 
Top-1 Value $\uparrow$&43.81\% & \textbf{44.48}\% & 44.38\% & 39.07\% & 39.07\% & 41.13\% & \textbf{46.08}\% & 42.62\% & 39.59\% & 41.93\% & 44.07\% & 45.05\% & \textbf{44.85}\% & 44.65\% & 44.81\% & 36.61\% & 43.65\% & 38.79\% & 15.99\% & 15.81\% & 15.75\% & 15.99\% & 14.70\% & \textbf{16.18}\% & 16.07\% & 15.66\% & \textbf{17.01}\% & 15.28\% & 14.74\% & 16.73\%\\[0.1em] \\[-1em] \cline{1-31}\\[-1em]
Top-5 (median) $\downarrow$&10\% & \textbf{9}\% & \textbf{9}\% & 35\% & 30\% & 19\% & 8\% & 36\% & 65\% & 37\% & 12\% & \textbf{7}\% & 38\% & \textbf{17}\% & \textbf{17}\% & 95\% & 38\% & 73\% & \textbf{10}\% & 12\% & 11\% & 29\% & 41\% & 12\% & 10\% & 27\% & \textbf{8}\% & 35\% & 43\% & 24\%\\[0.5em]
Top-5 ($1^{st}$-Q) $\downarrow$&9\% & \textbf{7}\% & 9\% & 32\% & 25\% & 18\% & 6\% & 22\% & 63\% & 32\% & 11\% & \textbf{5}\% & 33\% & \textbf{14}\% & \textbf{14}\% & 94\% & 34\% & 58\% & 9\% & 9\% & \textbf{7}\% & 16\% & 39\% & \textbf{7}\% & 7\% & 13\% & \textbf{6}\% & 24\% & 38\% & 10\%\\[0.5em]
Top-5 ($3^{rd}$-Q) $\downarrow$&11\% & \textbf{9}\% & 10\% & 38\% & 31\% & 20\% & \textbf{10}\% & 47\% & 72\% & 40\% & 15\% & \textbf{8}\% & 40\% & \textbf{18}\% & 23\% & 96\% & 47\% & 79\% & \textbf{13}\% & 17\% & 15\% & 41\% & 46\% & 27\% & 11\% & 43\% & \textbf{10}\% & 45\% & 49\% & 27\%\\[0.5em] 
Top-5 Value $\uparrow$&44.02\% & \textbf{44.45}\% & 44.17\% & 40.49\% & 40.88\% & 43.22\% & \textbf{44.87}\% & 42.39\% & 39.06\% & 41.48\% & 44.35\% & 44.78\% & 43.93\% & \textbf{45.33}\% & 45.20\% & 38.78\% & 43.79\% & 42.08\% & 16.22\% & 16.11\% & \textbf{16.31}\% & 15.13\% & 14.78\% & 15.82\% & 16.49\% & 15.39\% & \textbf{16.56}\% & 14.95\% & 14.84\% & 15.80\%\\[0.1em]
\bottomrule
DNN&\multicolumn{6}{c}{Densenet100bc}&\multicolumn{6}{c}{Preresnet110}&\multicolumn{6}{c}{Preresnet20}&\multicolumn{6}{c}{VGG19}&\multicolumn{6}{c}{VGG16} \\ \cline{2-31}
\multirow{2}{*}{Similarity} & &&&&&&&&&&&&&&&&&&&&&&&&&&&&& \\
($\mathcal{P}$)&\rotatebox[origin=c]{75}{PWCCA}&\rotatebox[origin=c]{75}{CKA}&\rotatebox[origin=c]{75}{Ortho}&\rotatebox[origin=c]{75}{Acc}&\rotatebox[origin=c]{75}{Dis}&\rotatebox[origin=c]{75}{J$_{Div}$}&\rotatebox[origin=c]{75}{PWCCA}&\rotatebox[origin=c]{75}{CKA}&\rotatebox[origin=c]{75}{Ortho}&\rotatebox[origin=c]{75}{Acc}&\rotatebox[origin=c]{75}{Dis}&\rotatebox[origin=c]{75}{J$_{Div}$}&\rotatebox[origin=c]{75}{PWCCA}&\rotatebox[origin=c]{75}{CKA}&\rotatebox[origin=c]{75}{Ortho}&\rotatebox[origin=c]{75}{Acc}&\rotatebox[origin=c]{75}{Dis}&\rotatebox[origin=c]{75}{J$_{Div}$}&\rotatebox[origin=c]{75}{PWCCA}&\rotatebox[origin=c]{75}{CKA}&\rotatebox[origin=c]{75}{Ortho}&\rotatebox[origin=c]{75}{Acc}&\rotatebox[origin=c]{75}{Dis}&\rotatebox[origin=c]{75}{J$_{Div}$}&\rotatebox[origin=c]{75}{PWCCA}&\rotatebox[origin=c]{75}{CKA}&\rotatebox[origin=c]{75}{Ortho}&\rotatebox[origin=c]{75}{Acc}&\rotatebox[origin=c]{75}{Dis}&\rotatebox[origin=c]{75}{J$_{Div}$} \\
\midrule
\midrule
Top-1 (median) $\downarrow$&8\% & 6\% & 9\% & 74\% & 74\% & \textbf{4}\% & \textbf{0}\% & 18\% & 22\% & 74\% & 24\% & 5\% & \textbf{0}\% & 21\% & 30\% & 95\% & 15\% & 75\% & 14\% & 25\% & \textbf{11}\% & 42\% & 57\% & 19\% & 21\% & 19\% & \textbf{9}\% & 60\% & 35\% & 26\%\\[0.5em]
Top-1 ($1^{st}$-Q) $\downarrow$&3\% & 3\% & 3\% & 66\% & 66\% & \textbf{2}\% & \textbf{0}\% & 10\% & 18\% & 66\% & 18\% & 1\% & \textbf{0}\% & 18\% & 18\% & 95\% & 8\% & 64\% & 10\% & 19\% & \textbf{3}\% & 19\% & 43\% & 13\% & 5\% & \textbf{4}\% & 5\% & 49\% & 20\% & 18\%\\[0.5em]
Top-1 ($3^{rd}$-Q) $\downarrow$&\textbf{10}\% & \textbf{10}\% & 12\% & 79\% & 79\% & 64\% & \textbf{4}\% & 24\% & 28\% & 81\% & 28\% & 8\% & \textbf{2}\% & 48\% & 52\% & 97\% & 31\% & 79\% & \textbf{23}\% & 36\% & 36\% & 68\% & 69\% & 36\% & 34\% & 69\% & \textbf{12}\% & 81\% & 68\% & 34\%\\[0.5em] 
Top-1 Value $\uparrow$&\textbf{39.07}\% & 38.73\% & 38.69\% & 32.78\% & 32.78\% & 37.54\% & \textbf{44.89}\% & 42.00\% & 40.68\% & 37.20\% & 41.39\% & 43.72\% & \textbf{44.89}\% & 41.33\% & 40.91\% & 34.70\% & 42.55\% & 39.12\% & 18.74\% & 18.58\% & \textbf{18.93}\% & 18.04\% & 17.77\% & 18.62\% & 17.55\% & 17.32\% & 17.85\% & 16.53\% & 17.12\% & \textbf{17.57}\%\\[0.1em] \\[-1em] \cline{1-31}\\[-1em]
Top-5 (median) $\downarrow$&\textbf{6}\% & 14\% & 11\% & 48\% & 39\% & 31\% & \textbf{10}\% & 18\% & 40\% & 69\% & 14\% & \textbf{10}\% & \textbf{15}\% & 42\% & 41\% & 73\% & 27\% & 59\% & \textbf{20}\% & 29\% & 26\% & 46\% & 52\% & 24\% & 22\% & 47\% & \textbf{15}\% & 52\% & 38\% & 34\%\\[0.5em]
Top-5 ($1^{st}$-Q) $\downarrow$&\textbf{5}\% & 13\% & 8\% & 43\% & 35\% & 28\% & \textbf{6}\% & 11\% & 36\% & 65\% & 12\% & 9\% & \textbf{14}\% & 35\% & 34\% & 67\% & 21\% & 54\% & \textbf{12}\% & 20\% & 20\% & 36\% & 41\% & 19\% & 18\% & 25\% & \textbf{11}\% & 42\% & 33\% & 28\%\\[0.5em]
Top-5 ($3^{rd}$-Q) $\downarrow$&\textbf{10}\% & 16\% & 15\% & 52\% & 46\% & 34\% & \textbf{12}\% & 24\% & 44\% & 74\% & 17\% & \textbf{12}\% & \textbf{23}\% & 45\% & 44\% & 75\% & 30\% & 75\% & \textbf{30}\% & 37\% & 36\% & 50\% & 61\% & 40\% & 26\% & 51\% & \textbf{24}\% & 60\% & 39\% & 37\%\\[0.5em] 
Top-5 Value $\uparrow$&\textbf{38.85}\% & 37.97\% & 38.45\% & 34.73\% & 34.75\% & 36.72\% & \textbf{42.91}\% & 41.85\% & 39.40\% & 37.37\% & 42.62\% & 42.52\% & \textbf{42.69}\% & 41.03\% & 41.17\% & 37.43\% & 42.03\% & 39.24\% & \textbf{18.73}\% & 18.56\% & 18.55\% & 18.12\% & 18.01\% & 18.63\% & 17.62\% & 17.05\% & \textbf{17.86}\% & 16.75\% & 17.34\% & 17.41\%\\[0.1em]
\bottomrule
DNN&\multicolumn{6}{c}{ROBERTA}&\multicolumn{6}{c}{XLNet}&\multicolumn{6}{c}{BERT}&\multicolumn{6}{c}{DistilBERT}&\multicolumn{6}{c}{Electra} \\ \cline{2-31}
\multirow{2}{*}{Similarity} & &&&&&&&&&&&&&&&&&&&&&&&&&&&&& \\
($\mathcal{P}$)&\rotatebox[origin=c]{75}{PWCCA}&\rotatebox[origin=c]{75}{CKA}&\rotatebox[origin=c]{75}{Ortho}&\rotatebox[origin=c]{75}{Acc}&\rotatebox[origin=c]{75}{Dis}&\rotatebox[origin=c]{75}{J$_{Div}$}&\rotatebox[origin=c]{75}{PWCCA}&\rotatebox[origin=c]{75}{CKA}&\rotatebox[origin=c]{75}{Ortho}&\rotatebox[origin=c]{75}{Acc}&\rotatebox[origin=c]{75}{Dis}&\rotatebox[origin=c]{75}{J$_{Div}$}&\rotatebox[origin=c]{75}{PWCCA}&\rotatebox[origin=c]{75}{CKA}&\rotatebox[origin=c]{75}{Ortho}&\rotatebox[origin=c]{75}{Acc}&\rotatebox[origin=c]{75}{Dis}&\rotatebox[origin=c]{75}{J$_{Div}$}&\rotatebox[origin=c]{75}{PWCCA}&\rotatebox[origin=c]{75}{CKA}&\rotatebox[origin=c]{75}{Ortho}&\rotatebox[origin=c]{75}{Acc}&\rotatebox[origin=c]{75}{Dis}&\rotatebox[origin=c]{75}{J$_{Div}$}&\rotatebox[origin=c]{75}{PWCCA}&\rotatebox[origin=c]{75}{CKA}&\rotatebox[origin=c]{75}{Ortho}&\rotatebox[origin=c]{75}{Acc}&\rotatebox[origin=c]{75}{Dis}&\rotatebox[origin=c]{75}{J$_{Div}$} \\
\midrule
\midrule
Top-1 (median) $\downarrow$&\textbf{6}\% & 34\% & 64\% & 25\% & 11\% & \textbf{6}\% & 30\% & \textbf{5}\% & 10\% & 16\% & \textbf{5}\% & 12\% & 20\% & \textbf{6}\% & 19\% & 52\% & 14\% & 8\% & 10\% & 35\% & \textbf{5}\% & 71\% & 22\% & 24\% & \textbf{11}\% & 22\% & 30\% & 49\% & 19\% & \textbf{11}\%\\[0.5em]
Top-1 ($1^{st}$-Q) $\downarrow$&\textbf{3}\% & 28\% & 61\% & 13\% & 6\% & \textbf{3}\% & 19\% & \textbf{3}\% & 6\% & 11\% & \textbf{3}\% & 8\% & 11\% & \textbf{2}\% & 13\% & 24\% & 6\% & 3\% & \textbf{2}\% & 24\% & 5\% & 15\% & 18\% & 8\% & 5\% & 15\% & 22\% & 46\% & 9\% & \textbf{3}\%\\[0.5em]
Top-1 ($3^{rd}$-Q) $\downarrow$&\textbf{9}\% & 50\% & 67\% & 53\% & 17\% & \textbf{9}\% & 48\% & \textbf{8}\% & 17\% & 22\% & \textbf{8}\% & 15\% & 42\% & \textbf{13}\% & 42\% & 83\% & 50\% & 26\% & 28\% & 40\% & \textbf{11}\% & 77\% & 29\% & 30\% & 17\% & 31\% & 35\% & 56\% & 26\% & \textbf{15}\%\\[0.5em] 
Top-1 Value $\uparrow$&22.01\% & 16.33\% & 14.55\% & 18.37\% & 21.42\% & \textbf{22.08}\% & 19.22\% & \textbf{23.61}\% & 22.51\% & 21.23\% & \textbf{23.61}\% & 22.32\% & 29.80\% & \textbf{31.59}\% & 29.49\% & 26.99\% & 29.67\% & 31.24\% & \textbf{32.20}\% & 30.79\% & 32.59\% & 28.37\% & 31.39\% & 31.60\% & 24.51\% & 23.03\% & 22.40\% & 20.83\% & 23.42\% & \textbf{24.93}\%\\[0.1em] \\[-1em] \cline{1-31}\\[-1em]
Top-5 (median) $\downarrow$&20\% & 46\% & 54\% & 20\% & \textbf{8}\% & \textbf{8}\% & 52\% & \textbf{8}\% & 12\% & 17\% & 10\% & 9\% & 21\% & 18\% & 24\% & 37\% & 22\% & \textbf{15}\% & 19\% & 19\% & \textbf{15}\% & 46\% & 16\% & 21\% & \textbf{9}\% & 14\% & 16\% & 41\% & 14\% & 11\%\\[0.5em]
Top-5 ($1^{st}$-Q) $\downarrow$&16\% & 43\% & 52\% & 13\% & 7\% & \textbf{5}\% & 43\% & 8\% & 11\% & 13\% & 8\% & \textbf{7}\% & 15\% & 12\% & 20\% & 30\% & 19\% & \textbf{12}\% & 13\% & 15\% & \textbf{10}\% & 44\% & 13\% & 16\% & \textbf{6}\% & 12\% & 13\% & 40\% & 11\% & 8\%\\[0.5em]
Top-5 ($3^{rd}$-Q) $\downarrow$&26\% & 48\% & 57\% & 44\% & 11\% & \textbf{8}\% & 56\% & \textbf{9}\% & 13\% & 23\% & 11\% & 10\% & 27\% & \textbf{20}\% & 33\% & 42\% & 29\% & \textbf{20}\% & \textbf{21}\% & 23\% & \textbf{21}\% & 48\% & \textbf{21}\% & 26\% & \textbf{15}\% & 17\% & 26\% & 50\% & \textbf{15}\% & 16\%\\[0.5em] 
Top-5 Value $\uparrow$&19.12\% & 15.95\% & 15.08\% & 18.53\% & 21.37\% & \textbf{21.46}\% & 18.20\% & \textbf{22.63}\% & 21.65\% & 21.25\% & 22.44\% & 22.26\% & 30.03\% & \textbf{30.29}\% & 29.51\% & 28.42\% & 29.82\% & 30.06\% & \textbf{31.59}\% & \textbf{31.59}\% & 31.77\% & 28.82\% & 31.52\% & 31.17\% & \textbf{24.61}\% & 24.17\% & 23.86\% & 21.37\% & 23.82\% & 24.36\%\\[0.1em]
\bottomrule
\end{tabular}}
\label{tab:rand_above_kmnc}
\end{table}

For all generation techniques and both properties, the metrics which led to the highest and most consistent correlations across DNN models under tests in previous RQs yield good results in terms of the \textit{Top-1} and \textit{Top-5}. However, other metrics can be better at choosing the best reference test set for some particular DNN model under test. For instance, \textit{CKA} for the \textit{Fuzz}-based generation technique also works well for DNN models such as \textit{PreResnet20} or \textit{VGG19} while \textit{CKA} did not have a significant correlation across all DNN model seeds. As such, this shows that the absence of a correlation across the reference sets does not necessarily mean that the most similar ones are not among the most property covering sets. Nonetheless, while the chosen metric is not necessarily the best for a given DNN model, it will generally be good across all DNN models under test. Indeed, if we rank the different similarity metrics for both \textit{Top-1} and \textit{Top-5} metrics (1 the best, 6 the lowest), the chosen metrics will have the highest average rank compared to all others. For instance for \textit{GAN}-based generation technique, \textit{PWCCA} has an average rank of $1.6$ for \textit{Top-1} criterion, where other metrics are above $2$ (i.e. on average \textit{PWCCA} is ranked $1.6$ out of $6$ on the DNN models under test). We obtain similar results for $J_{Div}$ on \textit{Fuzz}-based and \textit{Text}-based generation techniques both on \textit{Top-1} and \textit{Top-5} criteria. This is the case no matter the property chosen. Moreover, as the \textit{Top-1} and \textit{Top-5} values are lower than $50\%$, choosing the similarity that led to overall good correlations in the previous RQs is better than picking at random a reference test set. 

Regarding the value of the coverage for the reference test sets of the \textit{Top-1} and \textit{Top-5} DNN model seeds, the delta of \textit{Top-1} and \textit{Top-5} values do not change much on average. This shows that top DNN model seeds' reference test sets are consistent for both covering properties and a given DNN model under test. In terms of absolute value, there are discrepancies based on the property, the generation technique and the DNN model type. On fault-types based property, the coverage range is $\sim$ 20\%-45\% for the relevant similarities (\ie $PWCCA$ and $J_{Div}$). On the neuron-based property, the range is wider with $\sim$ 15\%-45\%. In particular, $VGG$ and $RoBERTa$/$XLNet$ DNN model types have their $T_O$ less covered in terms of this property. Nonetheless, by selecting one test set through our similarity, we can already cover in many cases more than a third of the covering property of $T_O$.

%This low property coverage was nonetheless to be expected. Indeed, since all test sets are generated specifically to fail on a particular DNN model, transferring them to a new DNN model under test would likely lead to fewer unique faulty inputs. Moreover, said faulty inputs would likely not all trigger the same neurons and fault-types as the generated $T_O$ would on the DNN model under test $O$. Thus, the overlap between a $T_O$ and $T_R$ would likely be small. Nonetheless, by selecting one test set through our similarity, we can already cover in many cases more than a third of the covering property of $T_O$.

\begin{tcolorbox}[colback=blue!5,colframe=blue!40!black]
\textbf{Findings 3:} Similarity metrics that showed good correlations generally enable the selection of good test sets for transferability. Even if said metrics are not necessarily the best for all DNN models under test, they will work better on average. Moreover, a selection of test sets based on the metric is always better than a random selection, further showing the utility of the method. Property coverage varies between DNN models under test/generation technique/property, but it's possible to cover up to $\sim 45\%$ of the $T_O$ property with a single test set.
\end{tcolorbox}

\begin{figure}\centering
\subfloat{\includegraphics[width=\linewidth]{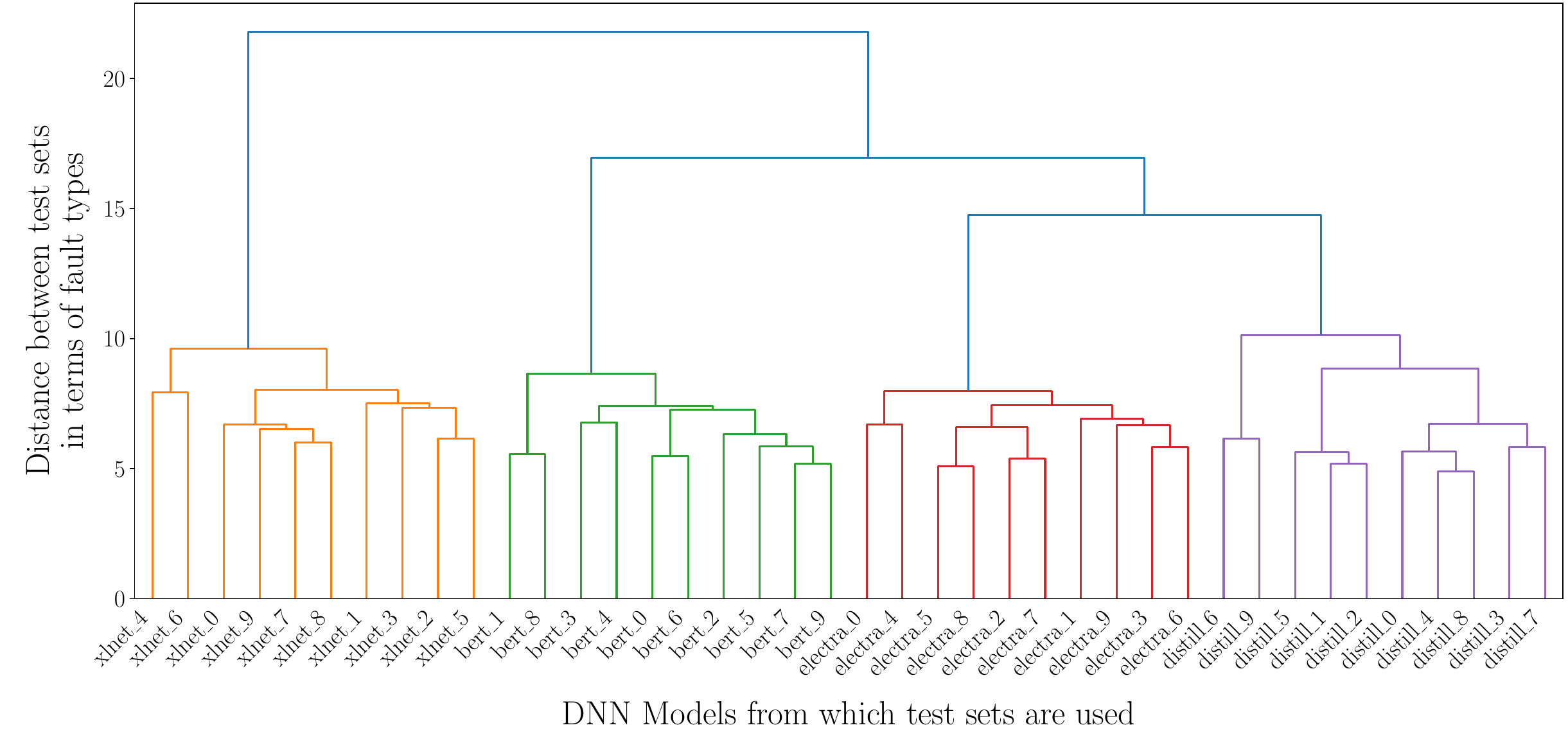}}\hfill
\subfloat{\includegraphics[width=\linewidth]{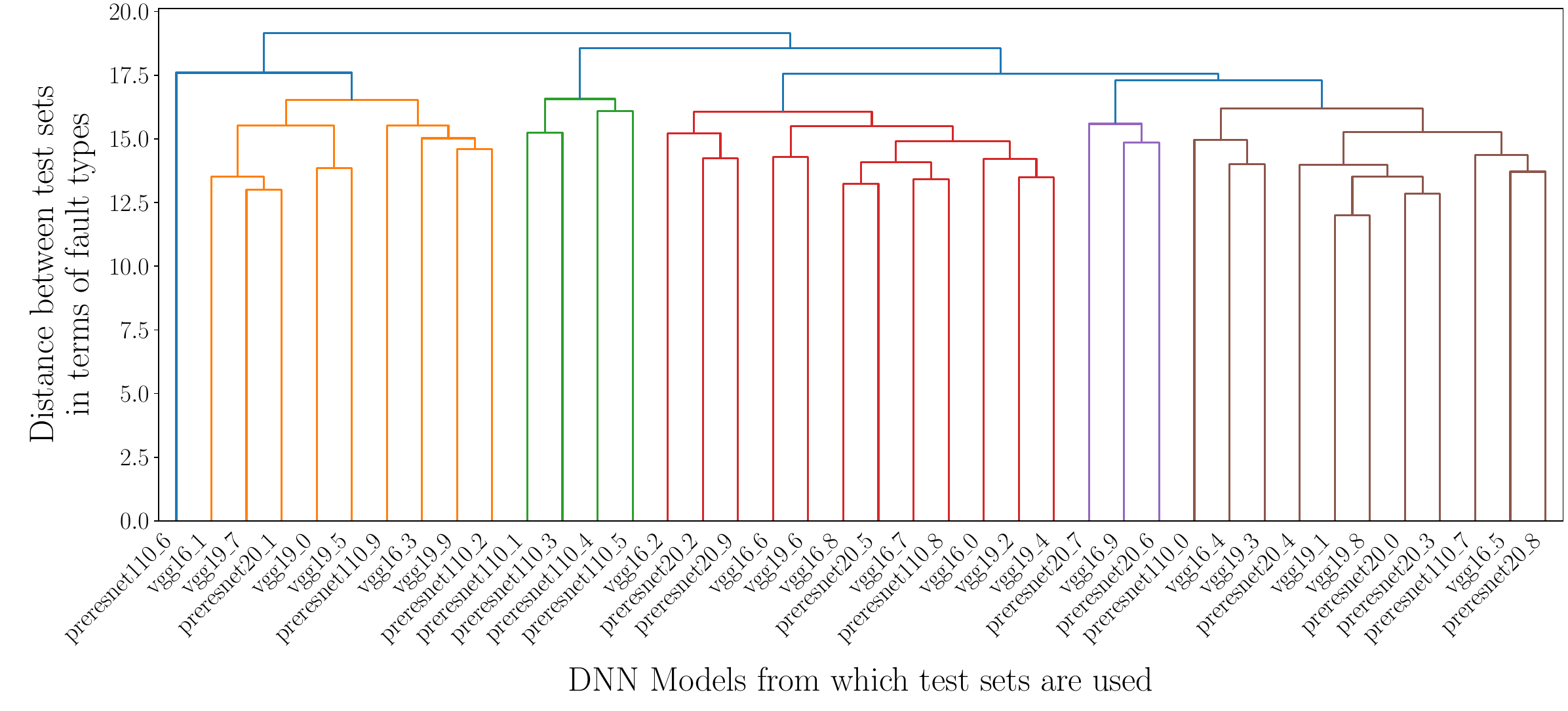}}\par
\caption{Distances between test sets for the fault types property on the DNN model under test. (\textbf{Top}) \textit{Text}-based generation technique when DNN model under test is of type \textit{RoBERTa}, (\textbf{Bottom}) \textit{GAN}-based generation technique when DNN model under test is of type \textit{Densenet100bc}.}
\label{fig:dendrograms}
\end{figure}

As we have seen in previous RQs, the average $\mathcal{M_O}$ coverage for each $T_O$ of a DNN model under test for a given reference test set does not exceed $50\%$. Thus, even by transferring one good test set, the coverage is relatively low and can be improved. One straightforward way to improve the coverage would be to combine different test sets. Ideally, we would like the combined test sets to individually cover as much of a property (fault-type or neuron activation bands) as possible while individually covering different regions of the property to increase coverage. 

Leveraging previous similarity metrics, we propose two heuristic criteria for combining test sets. First, since we have shown that some similarity metrics correlate with $\mathcal{M_O}$ in the previous RQs, the \textit{n}-th test sets of the most similar DNN models should generally cover more the property. If there is little overlap in the covered property (for instance, test sets not covering the same subset of fault types), the combination of said test sets should naturally improve the property coverage.
We call this criterion \textit{Overall Best First} (OBF) in the following. On the other hand, if the overlap in terms of covered property for the test sets is more pronounced or if the test sets individually do not cover much relatively of the property, the combination will not improve much in terms of the property. In that case, it might be better to instead combine test sets from different DNN model types as they can mitigate the overlap. In that case, we could combine test sets of the $n$-th most similar DNN model types. To do this, for each DNN model type (\eg \textit{roberta}, \textit{xlnet} ...) we can determine the most similar (according to the chosen similarity) to the DNN model under test and use their combined test sets together. We call this criterion \textit{Each Best First} (EBF) in the following.

As motivating examples to illustrate the defined criteria, we selected two DNN model seeds (one RoBERTa and one Densenet100bc) on two different generation techniques (\textit{Text}-based and \textit{GAN}-based generation technique). For each of those DNN model seeds, we took available reference test sets and extracted the fault types on a given DNN model under test for each reference test set after using the clustering procedure defined in Section \ref{sec:approach}. We created vectors of $k$ dimensions, where each dimension corresponds to one fault type identified for the $T_O$ of the DNN model under test. The value of each dimension represents how many inputs of the reference test set were clustered in the given fault type cluster. We then leverage hierarchical clustering \cite{King14} to cluster the reference tests set for each DNN model under test according to their distribution across fault types. We can represent the results using a dendrogram for each DNN model under test. Results are presented in Figure \ref{fig:dendrograms}.

\begin{table}
\fontsize{15}{17}\selectfont
\caption{Overall Best First, Each Best First and Random criterion score for the property based on fault-type coverage. Results are given for each generation technique and each DNN model under test using the chosen similarity metrics. For the same DNN model type (\eg \textit{Densenet100bc}), the results are given for all DNN model seeds. We highlight in bold the highest results.}
\centering
\resizebox{\textwidth}{!}{\begin{tabular}{cc||cc|cc|cc||cc|cc|cc||cc|cc|cc||cc|cc|cc||cc|cc|cc}
\toprule
&DNN&\multicolumn{6}{c}{Densenet100bc}&\multicolumn{6}{c}{Preresnet110}&\multicolumn{6}{c}{Preresnet20}&\multicolumn{6}{c}{VGG19}&\multicolumn{6}{c}{VGG16} \\ \cline{3-32}
\huge{Fuzz}& \multirow{2}{*}{Fault types cov.} & &&&&&&&&&&&&&&&&&&&&&&&&&&&&& \\
&&\multicolumn{2}{c|}{Median}&\multicolumn{2}{c|}{$1^{st}$-Q}&\multicolumn{2}{c||}{$3^{rd}$-Q}&\multicolumn{2}{c|}{Median}&\multicolumn{2}{c|}{$1^{st}$-Q}&\multicolumn{2}{c||}{$3^{rd}$-Q}&\multicolumn{2}{c|}{Median}&\multicolumn{2}{c|}{$1^{st}$-Q}&\multicolumn{2}{c||}{$3^{rd}$-Q}&\multicolumn{2}{c|}{Median}&\multicolumn{2}{c|}{$1^{st}$-Q}&\multicolumn{2}{c||}{$3^{rd}$-Q}&\multicolumn{2}{c|}{Median}&\multicolumn{2}{c|}{$1^{st}$-Q}&\multicolumn{2}{c}{$3^{rd}$-Q}\\
\midrule
\midrule
\multirow{9}{*}{\rotatebox[origin=c]{90}{Selection Heuristic}}&EBF 2 $\uparrow$&\multicolumn{2}{c|}{\textbf{47.81}}&\multicolumn{2}{c|}{\textbf{46.61}}&\multicolumn{2}{c||}{\textbf{49.02}}&\multicolumn{2}{c|}{50.73}&\multicolumn{2}{c|}{49.38}&\multicolumn{2}{c||}{51.38}&\multicolumn{2}{c|}{\textbf{52.46}}&\multicolumn{2}{c|}{\textbf{51.71}}&\multicolumn{2}{c||}{\textbf{52.96}}&\multicolumn{2}{c|}{65.28}&\multicolumn{2}{c|}{60.74}&\multicolumn{2}{c||}{67.14}&\multicolumn{2}{c|}{62.63}&\multicolumn{2}{c|}{60.26}&\multicolumn{2}{c}{64.07}\\

&OBF 2 $\uparrow$&\multicolumn{2}{c|}{\textbf{47.81}}&\multicolumn{2}{c|}{\textbf{46.61}}&\multicolumn{2}{c||}{\textbf{49.02}}&\multicolumn{2}{c|}{\textbf{54.11}}&\multicolumn{2}{c|}{\textbf{52.85}}&\multicolumn{2}{c||}{\textbf{55.28}}&\multicolumn{2}{c|}{49.91}&\multicolumn{2}{c|}{47.69}&\multicolumn{2}{c||}{51.52}&\multicolumn{2}{c|}{\textbf{67.03}}&\multicolumn{2}{c|}{\textbf{61.54}}&\multicolumn{2}{c||}{\textbf{70.23}}&\multicolumn{2}{c|}{\textbf{63.65}}&\multicolumn{2}{c|}{\textbf{62.25}}&\multicolumn{2}{c}{\textbf{66.67}}\\

&rand 2 $\uparrow$&\multicolumn{2}{c|}{46.19}&\multicolumn{2}{c|}{43.02}&\multicolumn{2}{c||}{49.48}&\multicolumn{2}{c|}{50.26}&\multicolumn{2}{c|}{47.74}&\multicolumn{2}{c||}{52.68}&\multicolumn{2}{c|}{50.21}&\multicolumn{2}{c|}{47.80}&\multicolumn{2}{c||}{52.24}&\multicolumn{2}{c|}{58.63}&\multicolumn{2}{c|}{54.94}&\multicolumn{2}{c||}{62.03}&\multicolumn{2}{c|}{55.60}&\multicolumn{2}{c|}{51.98}&\multicolumn{2}{c}{59.13}\\ \cline{2-32}

&EBF 3 $\uparrow$&\multicolumn{2}{c|}{59.47}&\multicolumn{2}{c|}{58.56}&\multicolumn{2}{c||}{60.62}&\multicolumn{2}{c|}{63.40}&\multicolumn{2}{c|}{61.16}&\multicolumn{2}{c||}{64.47}&\multicolumn{2}{c|}{\textbf{66.15}}&\multicolumn{2}{c|}{\textbf{64.40}}&\multicolumn{2}{c||}{\textbf{66.79}}&\multicolumn{2}{c|}{74.07}&\multicolumn{2}{c|}{71.34}&\multicolumn{2}{c||}{76.22}&\multicolumn{2}{c|}{71.93}&\multicolumn{2}{c|}{70.93}&\multicolumn{2}{c}{73.07}\\

&OBF 3 $\uparrow$&\multicolumn{2}{c|}{\textbf{62.31}}&\multicolumn{2}{c|}{\textbf{61.73}}&\multicolumn{2}{c||}{\textbf{64.49}}&\multicolumn{2}{c|}{\textbf{65.17}}&\multicolumn{2}{c|}{\textbf{63.84}}&\multicolumn{2}{c||}{\textbf{66.04}}&\multicolumn{2}{c|}{63.12}&\multicolumn{2}{c|}{60.41}&\multicolumn{2}{c||}{65.47}&\multicolumn{2}{c|}{\textbf{77.75}}&\multicolumn{2}{c|}{\textbf{71.93}}&\multicolumn{2}{c||}{\textbf{78.87}}&\multicolumn{2}{c|}{\textbf{75.55}}&\multicolumn{2}{c|}{\textbf{73.63}}&\multicolumn{2}{c}{\textbf{77.30}}\\

&rand 3 $\uparrow$&\multicolumn{2}{c|}{58.06}&\multicolumn{2}{c|}{54.16}&\multicolumn{2}{c||}{61.91}&\multicolumn{2}{c|}{62.05}&\multicolumn{2}{c|}{59.59}&\multicolumn{2}{c||}{64.50}&\multicolumn{2}{c|}{62.01}&\multicolumn{2}{c|}{60.06}&\multicolumn{2}{c||}{64.08}&\multicolumn{2}{c|}{70.38}&\multicolumn{2}{c|}{67.57}&\multicolumn{2}{c||}{73.06}&\multicolumn{2}{c|}{68.60}&\multicolumn{2}{c|}{65.61}&\multicolumn{2}{c}{71.31}\\ \cline{2-32}

&EBF 4 $\uparrow$&\multicolumn{2}{c|}{67.57}&\multicolumn{2}{c|}{66.63}&\multicolumn{2}{c||}{68.10}&\multicolumn{2}{c|}{71.15}&\multicolumn{2}{c|}{\textbf{70.28}}&\multicolumn{2}{c||}{72.45}&\multicolumn{2}{c|}{\textbf{73.12}}&\multicolumn{2}{c|}{\textbf{71.62}}&\multicolumn{2}{c||}{74.27}&\multicolumn{2}{c|}{79.68}&\multicolumn{2}{c|}{78.83}&\multicolumn{2}{c||}{80.98}&\multicolumn{2}{c|}{77.11}&\multicolumn{2}{c|}{76.53}&\multicolumn{2}{c}{79.02}\\

&OBF 4 $\uparrow$&\multicolumn{2}{c|}{\textbf{71.80}}&\multicolumn{2}{c|}{\textbf{71.21}}&\multicolumn{2}{c||}{\textbf{73.37}}&\multicolumn{2}{c|}{71.69}&\multicolumn{2}{c|}{69.40}&\multicolumn{2}{c||}{72.30}&\multicolumn{2}{c|}{71.35}&\multicolumn{2}{c|}{68.29}&\multicolumn{2}{c||}{\textbf{74.62}}&\multicolumn{2}{c|}{\textbf{82.75}}&\multicolumn{2}{c|}{\textbf{79.38}}&\multicolumn{2}{c||}{\textbf{83.76}}&\multicolumn{2}{c|}{\textbf{81.07}}&\multicolumn{2}{c|}{\textbf{78.17}}&\multicolumn{2}{c}{\textbf{82.69}}\\

&rand 4 $\uparrow$&\multicolumn{2}{c|}{68.97}&\multicolumn{2}{c|}{66.49}&\multicolumn{2}{c||}{71.71}&\multicolumn{2}{c|}{\textbf{71.87}}&\multicolumn{2}{c|}{69.48}&\multicolumn{2}{c||}{\textbf{74.07}}&\multicolumn{2}{c|}{71.20}&\multicolumn{2}{c|}{69.37}&\multicolumn{2}{c||}{73.50}&\multicolumn{2}{c|}{78.86}&\multicolumn{2}{c|}{76.00}&\multicolumn{2}{c||}{80.89}&\multicolumn{2}{c|}{76.92}&\multicolumn{2}{c|}{74.37}&\multicolumn{2}{c}{79.10}\\

\bottomrule
&DNN&\multicolumn{6}{c}{Densenet100bc}&\multicolumn{6}{c}{Preresnet110}&\multicolumn{6}{c}{Preresnet20}&\multicolumn{6}{c}{VGG19}&\multicolumn{6}{c}{VGG16} \\ \cline{3-32}
\huge{GAN}& \multirow{2}{*}{Similarity} & &&&&&&&&&&&&&&&&&&&&&&&&&&&&& \\
&&\multicolumn{2}{c|}{Median}&\multicolumn{2}{c|}{$1^{st}$-Q}&\multicolumn{2}{c||}{$3^{rd}$-Q}&\multicolumn{2}{c|}{Median}&\multicolumn{2}{c|}{$1^{st}$-Q}&\multicolumn{2}{c||}{$3^{rd}$-Q}&\multicolumn{2}{c|}{Median}&\multicolumn{2}{c|}{$1^{st}$-Q}&\multicolumn{2}{c||}{$3^{rd}$-Q}&\multicolumn{2}{c|}{Median}&\multicolumn{2}{c|}{$1^{st}$-Q}&\multicolumn{2}{c||}{$3^{rd}$-Q}&\multicolumn{2}{c|}{Median}&\multicolumn{2}{c|}{$1^{st}$-Q}&\multicolumn{2}{c}{$3^{rd}$-Q}\\
\midrule
\midrule
\multirow{9}{*}{\rotatebox[origin=c]{90}{Selection Heuristic}}&EBF 2 $\uparrow$&\multicolumn{2}{c|}{40.67}&\multicolumn{2}{c|}{38.87}&\multicolumn{2}{c||}{42.58}&\multicolumn{2}{c|}{39.85}&\multicolumn{2}{c|}{37.42}&\multicolumn{2}{c||}{40.90}&\multicolumn{2}{c|}{\textbf{38.13}}&\multicolumn{2}{c|}{\textbf{36.45}}&\multicolumn{2}{c||}{\textbf{40.04}}&\multicolumn{2}{c|}{54.94}&\multicolumn{2}{c|}{53.12}&\multicolumn{2}{c||}{55.95}&\multicolumn{2}{c|}{53.31}&\multicolumn{2}{c|}{50.80}&\multicolumn{2}{c}{54.46}\\

&OBF 2 $\uparrow$&\multicolumn{2}{c|}{\textbf{44.64}}&\multicolumn{2}{c|}{\textbf{41.55}}&\multicolumn{2}{c||}{\textbf{45.75}}&\multicolumn{2}{c|}{\textbf{40.98}}&\multicolumn{2}{c|}{\textbf{40.26}}&\multicolumn{2}{c||}{\textbf{42.67}}&\multicolumn{2}{c|}{37.69}&\multicolumn{2}{c|}{36.30}&\multicolumn{2}{c||}{39.95}&\multicolumn{2}{c|}{\textbf{55.56}}&\multicolumn{2}{c|}{\textbf{54.55}}&\multicolumn{2}{c||}{\textbf{56.85}}&\multicolumn{2}{c|}{\textbf{53.80}}&\multicolumn{2}{c|}{\textbf{51.83}}&\multicolumn{2}{c}{\textbf{54.92}}\\

&rand 2 $\uparrow$&\multicolumn{2}{c|}{38.85}&\multicolumn{2}{c|}{37.11}&\multicolumn{2}{c||}{40.70}&\multicolumn{2}{c|}{35.55}&\multicolumn{2}{c|}{33.93}&\multicolumn{2}{c||}{37.37}&\multicolumn{2}{c|}{35.69}&\multicolumn{2}{c|}{34.23}&\multicolumn{2}{c||}{37.70}&\multicolumn{2}{c|}{51.98}&\multicolumn{2}{c|}{50.45}&\multicolumn{2}{c||}{54.28}&\multicolumn{2}{c|}{51.44}&\multicolumn{2}{c|}{49.11}&\multicolumn{2}{c}{53.27}\\ \cline{2-32}

&EBF 3 $\uparrow$&\multicolumn{2}{c|}{52.72}&\multicolumn{2}{c|}{49.74}&\multicolumn{2}{c||}{54.67}&\multicolumn{2}{c|}{49.42}&\multicolumn{2}{c|}{47.33}&\multicolumn{2}{c||}{50.64}&\multicolumn{2}{c|}{\textbf{50.55}}&\multicolumn{2}{c|}{47.78}&\multicolumn{2}{c||}{\textbf{52.69}}&\multicolumn{2}{c|}{67.07}&\multicolumn{2}{c|}{66.26}&\multicolumn{2}{c||}{67.74}&\multicolumn{2}{c|}{65.45}&\multicolumn{2}{c|}{63.54}&\multicolumn{2}{c}{66.88}\\

&OBF 3 $\uparrow$&\multicolumn{2}{c|}{\textbf{57.41}}&\multicolumn{2}{c|}{\textbf{56.61}}&\multicolumn{2}{c||}{\textbf{58.91}}&\multicolumn{2}{c|}{\textbf{52.79}}&\multicolumn{2}{c|}{\textbf{51.23}}&\multicolumn{2}{c||}{\textbf{53.18}}&\multicolumn{2}{c|}{49.60}&\multicolumn{2}{c|}{\textbf{47.92}}&\multicolumn{2}{c||}{51.45}&\multicolumn{2}{c|}{\textbf{67.23}}&\multicolumn{2}{c|}{\textbf{66.42}}&\multicolumn{2}{c||}{\textbf{68.40}}&\multicolumn{2}{c|}{\textbf{65.92}}&\multicolumn{2}{c|}{\textbf{64.76}}&\multicolumn{2}{c}{\textbf{67.60}}\\

&rand 3 $\uparrow$&\multicolumn{2}{c|}{50.63}&\multicolumn{2}{c|}{48.80}&\multicolumn{2}{c||}{52.47}&\multicolumn{2}{c|}{46.68}&\multicolumn{2}{c|}{44.71}&\multicolumn{2}{c||}{48.99}&\multicolumn{2}{c|}{47.80}&\multicolumn{2}{c|}{46.20}&\multicolumn{2}{c||}{49.51}&\multicolumn{2}{c|}{64.76}&\multicolumn{2}{c|}{63.13}&\multicolumn{2}{c||}{66.87}&\multicolumn{2}{c|}{63.80}&\multicolumn{2}{c|}{62.17}&\multicolumn{2}{c}{65.50}\\ \cline{2-32}

&EBF 4 $\uparrow$&\multicolumn{2}{c|}{59.95}&\multicolumn{2}{c|}{57.83}&\multicolumn{2}{c||}{61.77}&\multicolumn{2}{c|}{57.89}&\multicolumn{2}{c|}{55.55}&\multicolumn{2}{c||}{59.36}&\multicolumn{2}{c|}{\textbf{58.86}}&\multicolumn{2}{c|}{56.65}&\multicolumn{2}{c||}{\textbf{61.54}}&\multicolumn{2}{c|}{74.97}&\multicolumn{2}{c|}{73.11}&\multicolumn{2}{c||}{75.84}&\multicolumn{2}{c|}{73.11}&\multicolumn{2}{c|}{71.96}&\multicolumn{2}{c}{73.76}\\

&OBF 4 $\uparrow$&\multicolumn{2}{c|}{\textbf{65.72}}&\multicolumn{2}{c|}{\textbf{64.39}}&\multicolumn{2}{c||}{\textbf{66.73}}&\multicolumn{2}{c|}{\textbf{61.94}}&\multicolumn{2}{c|}{\textbf{61.49}}&\multicolumn{2}{c||}{\textbf{63.82}}&\multicolumn{2}{c|}{58.12}&\multicolumn{2}{c|}{\textbf{57.84}}&\multicolumn{2}{c||}{59.69}&\multicolumn{2}{c|}{\textbf{76.03}}&\multicolumn{2}{c|}{\textbf{74.92}}&\multicolumn{2}{c||}{\textbf{77.40}}&\multicolumn{2}{c|}{\textbf{75.69}}&\multicolumn{2}{c|}{\textbf{75.04}}&\multicolumn{2}{c}{\textbf{76.35}}\\

&rand 4 $\uparrow$&\multicolumn{2}{c|}{60.24}&\multicolumn{2}{c|}{58.96}&\multicolumn{2}{c||}{62.19}&\multicolumn{2}{c|}{57.43}&\multicolumn{2}{c|}{55.18}&\multicolumn{2}{c||}{59.55}&\multicolumn{2}{c|}{57.11}&\multicolumn{2}{c|}{55.24}&\multicolumn{2}{c||}{58.94}&\multicolumn{2}{c|}{73.77}&\multicolumn{2}{c|}{71.79}&\multicolumn{2}{c||}{75.43}&\multicolumn{2}{c|}{72.69}&\multicolumn{2}{c|}{70.83}&\multicolumn{2}{c}{74.70}\\

\bottomrule
&DNN&\multicolumn{6}{c}{ROBERTA}&\multicolumn{6}{c}{XLNet}&\multicolumn{6}{c}{BERT}&\multicolumn{6}{c}{DistilBERT}&\multicolumn{6}{c}{Electra} \\ \cline{3-32}
\huge{Text}& \multirow{2}{*}{Similarity} & &&&&&&&&&&&&&&&&&&&&&&&&&&&&& \\
&&\multicolumn{2}{c|}{Median}&\multicolumn{2}{c|}{$1^{st}$-Q}&\multicolumn{2}{c||}{$3^{rd}$-Q}&\multicolumn{2}{c|}{Median}&\multicolumn{2}{c|}{$1^{st}$-Q}&\multicolumn{2}{c||}{$3^{rd}$-Q}&\multicolumn{2}{c|}{Median}&\multicolumn{2}{c|}{$1^{st}$-Q}&\multicolumn{2}{c||}{$3^{rd}$-Q}&\multicolumn{2}{c|}{Median}&\multicolumn{2}{c|}{$1^{st}$-Q}&\multicolumn{2}{c||}{$3^{rd}$-Q}&\multicolumn{2}{c|}{Median}&\multicolumn{2}{c|}{$1^{st}$-Q}&\multicolumn{2}{c}{$3^{rd}$-Q}\\
\midrule
\midrule
\multirow{9}{*}{\rotatebox[origin=c]{90}{Selection Heuristic}}&EBF 2 $\uparrow$&\multicolumn{2}{c|}{\textbf{53.15}}&\multicolumn{2}{c|}{\textbf{50.82}}&\multicolumn{2}{c||}{\textbf{56.21}}&\multicolumn{2}{c|}{\textbf{52.22}}&\multicolumn{2}{c|}{\textbf{49.77}}&\multicolumn{2}{c||}{\textbf{53.08}}&\multicolumn{2}{c|}{\textbf{53.58}}&\multicolumn{2}{c|}{\textbf{51.68}}&\multicolumn{2}{c||}{\textbf{56.61}}&\multicolumn{2}{c|}{\textbf{59.63}}&\multicolumn{2}{c|}{\textbf{57.26}}&\multicolumn{2}{c||}{\textbf{62.50}}&\multicolumn{2}{c|}{\textbf{54.34}}&\multicolumn{2}{c|}{\textbf{53.52}}&\multicolumn{2}{c}{\textbf{56.91}}\\

&OBF 2 $\uparrow$&\multicolumn{2}{c|}{49.30}&\multicolumn{2}{c|}{48.82}&\multicolumn{2}{c||}{52.54}&\multicolumn{2}{c|}{48.59}&\multicolumn{2}{c|}{47.15}&\multicolumn{2}{c||}{49.66}&\multicolumn{2}{c|}{52.55}&\multicolumn{2}{c|}{49.30}&\multicolumn{2}{c||}{53.67}&\multicolumn{2}{c|}{50.59}&\multicolumn{2}{c|}{50.05}&\multicolumn{2}{c||}{51.76}&\multicolumn{2}{c|}{51.22}&\multicolumn{2}{c|}{49.92}&\multicolumn{2}{c}{53.08}\\

&rand 2 $\uparrow$&\multicolumn{2}{c|}{42.28}&\multicolumn{2}{c|}{37.01}&\multicolumn{2}{c||}{49.30}&\multicolumn{2}{c|}{43.59}&\multicolumn{2}{c|}{39.22}&\multicolumn{2}{c||}{47.78}&\multicolumn{2}{c|}{43.66}&\multicolumn{2}{c|}{38.45}&\multicolumn{2}{c||}{50.98}&\multicolumn{2}{c|}{45.75}&\multicolumn{2}{c|}{36.11}&\multicolumn{2}{c||}{51.39}&\multicolumn{2}{c|}{44.63}&\multicolumn{2}{c|}{39.77}&\multicolumn{2}{c}{49.77}\\ \cline{2-32}

&EBF 3 $\uparrow$&\multicolumn{2}{c|}{\textbf{64.60}}&\multicolumn{2}{c|}{\textbf{61.72}}&\multicolumn{2}{c||}{\textbf{66.14}}&\multicolumn{2}{c|}{\textbf{65.85}}&\multicolumn{2}{c|}{\textbf{63.86}}&\multicolumn{2}{c||}{\textbf{67.07}}&\multicolumn{2}{c|}{\textbf{65.65}}&\multicolumn{2}{c|}{\textbf{62.30}}&\multicolumn{2}{c||}{\textbf{66.67}}&\multicolumn{2}{c|}{\textbf{69.43}}&\multicolumn{2}{c|}{\textbf{64.54}}&\multicolumn{2}{c||}{\textbf{71.01}}&\multicolumn{2}{c|}{\textbf{67.01}}&\multicolumn{2}{c|}{\textbf{66.38}}&\multicolumn{2}{c}{\textbf{68.09}}\\

&OBF 3 $\uparrow$&\multicolumn{2}{c|}{59.77}&\multicolumn{2}{c|}{56.75}&\multicolumn{2}{c||}{61.32}&\multicolumn{2}{c|}{56.51}&\multicolumn{2}{c|}{54.93}&\multicolumn{2}{c||}{57.29}&\multicolumn{2}{c|}{59.88}&\multicolumn{2}{c|}{57.84}&\multicolumn{2}{c||}{61.87}&\multicolumn{2}{c|}{56.00}&\multicolumn{2}{c|}{54.06}&\multicolumn{2}{c||}{60.09}&\multicolumn{2}{c|}{60.94}&\multicolumn{2}{c|}{58.40}&\multicolumn{2}{c}{63.30}\\

&rand 3 $\uparrow$&\multicolumn{2}{c|}{54.30}&\multicolumn{2}{c|}{48.83}&\multicolumn{2}{c||}{60.69}&\multicolumn{2}{c|}{53.85}&\multicolumn{2}{c|}{49.92}&\multicolumn{2}{c||}{59.50}&\multicolumn{2}{c|}{55.92}&\multicolumn{2}{c|}{50.18}&\multicolumn{2}{c||}{62.96}&\multicolumn{2}{c|}{57.54}&\multicolumn{2}{c|}{53.77}&\multicolumn{2}{c||}{63.41}&\multicolumn{2}{c|}{59.77}&\multicolumn{2}{c|}{56.24}&\multicolumn{2}{c}{63.85}\\ \cline{2-32}

&EBF 4 $\uparrow$&\multicolumn{2}{c|}{\textbf{73.91}}&\multicolumn{2}{c|}{\textbf{72.40}}&\multicolumn{2}{c||}{\textbf{76.06}}&\multicolumn{2}{c|}{\textbf{76.80}}&\multicolumn{2}{c|}{\textbf{73.69}}&\multicolumn{2}{c||}{\textbf{78.37}}&\multicolumn{2}{c|}{\textbf{75.11}}&\multicolumn{2}{c|}{\textbf{72.60}}&\multicolumn{2}{c||}{\textbf{76.11}}&\multicolumn{2}{c|}{\textbf{76.31}}&\multicolumn{2}{c|}{\textbf{73.82}}&\multicolumn{2}{c||}{\textbf{78.26}}&\multicolumn{2}{c|}{\textbf{75.74}}&\multicolumn{2}{c|}{\textbf{73.20}}&\multicolumn{2}{c}{\textbf{77.13}}\\

&OBF 4 $\uparrow$&\multicolumn{2}{c|}{64.85}&\multicolumn{2}{c|}{63.15}&\multicolumn{2}{c||}{66.83}&\multicolumn{2}{c|}{61.67}&\multicolumn{2}{c|}{60.52}&\multicolumn{2}{c||}{62.79}&\multicolumn{2}{c|}{66.67}&\multicolumn{2}{c|}{63.30}&\multicolumn{2}{c||}{67.34}&\multicolumn{2}{c|}{61.37}&\multicolumn{2}{c|}{59.13}&\multicolumn{2}{c||}{64.01}&\multicolumn{2}{c|}{67.57}&\multicolumn{2}{c|}{66.32}&\multicolumn{2}{c}{70.26}\\

&rand 4 $\uparrow$&\multicolumn{2}{c|}{69.12}&\multicolumn{2}{c|}{64.02}&\multicolumn{2}{c||}{72.09}&\multicolumn{2}{c|}{68.26}&\multicolumn{2}{c|}{63.97}&\multicolumn{2}{c||}{71.79}&\multicolumn{2}{c|}{69.94}&\multicolumn{2}{c|}{64.57}&\multicolumn{2}{c||}{73.24}&\multicolumn{2}{c|}{70.11}&\multicolumn{2}{c|}{66.10}&\multicolumn{2}{c||}{74.29}&\multicolumn{2}{c|}{70.18}&\multicolumn{2}{c|}{66.05}&\multicolumn{2}{c}{73.06}\\
\bottomrule
\end{tabular}}
\label{tab:crit_comp}
\end{table}

\begin{table}
\fontsize{15}{17}\selectfont
\caption{Overall Best First, Each Best First and Random criterion score for the property based on neuron coverage. Results are given for each generation technique and each DNN model under test using the chosen similarity metrics. For the same DNN model type (\eg \textit{Densenet100bc}), the results are given for all DNN model seeds. We highlight in bold the highest results.}
\centering
\resizebox{\textwidth}{!}{\begin{tabular}{cc||cc|cc|cc||cc|cc|cc||cc|cc|cc||cc|cc|cc||cc|cc|cc}
\toprule
&DNN&\multicolumn{6}{c}{Densenet100bc}&\multicolumn{6}{c}{Preresnet110}&\multicolumn{6}{c}{Preresnet20}&\multicolumn{6}{c}{VGG19}&\multicolumn{6}{c}{VGG16} \\ \cline{3-32}
\huge{Fuzz}& \multirow{2}{*}{Fault types cov.} & &&&&&&&&&&&&&&&&&&&&&&&&&&&&& \\
&&\multicolumn{2}{c|}{Median}&\multicolumn{2}{c|}{$1^{st}$-Q}&\multicolumn{2}{c||}{$3^{rd}$-Q}&\multicolumn{2}{c|}{Median}&\multicolumn{2}{c|}{$1^{st}$-Q}&\multicolumn{2}{c||}{$3^{rd}$-Q}&\multicolumn{2}{c|}{Median}&\multicolumn{2}{c|}{$1^{st}$-Q}&\multicolumn{2}{c||}{$3^{rd}$-Q}&\multicolumn{2}{c|}{Median}&\multicolumn{2}{c|}{$1^{st}$-Q}&\multicolumn{2}{c||}{$3^{rd}$-Q}&\multicolumn{2}{c|}{Median}&\multicolumn{2}{c|}{$1^{st}$-Q}&\multicolumn{2}{c}{$3^{rd}$-Q}\\
\midrule
\midrule
\multirow{9}{*}{\rotatebox[origin=c]{90}{Selection Heuristic}}&EBF 2 $\uparrow$&\multicolumn{2}{c|}{60.87}&\multicolumn{2}{c|}{58.35}&\multicolumn{2}{c||}{61.77}&\multicolumn{2}{c|}{60.88}&\multicolumn{2}{c|}{60.31}&\multicolumn{2}{c||}{62.19}&\multicolumn{2}{c|}{\textbf{64.85}}&\multicolumn{2}{c|}{61.72}&\multicolumn{2}{c||}{\textbf{65.54}}&\multicolumn{2}{c|}{21.81}&\multicolumn{2}{c|}{20.42}&\multicolumn{2}{c||}{23.63}&\multicolumn{2}{c|}{23.90}&\multicolumn{2}{c|}{20.24}&\multicolumn{2}{c}{28.06}\\

&OBF 2 $\uparrow$&\multicolumn{2}{c|}{\textbf{63.23}}&\multicolumn{2}{c|}{\textbf{63.11}}&\multicolumn{2}{c||}{\textbf{63.59}}&\multicolumn{2}{c|}{\textbf{63.72}}&\multicolumn{2}{c|}{\textbf{62.93}}&\multicolumn{2}{c||}{\textbf{64.15}}&\multicolumn{2}{c|}{63.57}&\multicolumn{2}{c|}{\textbf{62.07}}&\multicolumn{2}{c||}{64.02}&\multicolumn{2}{c|}{\textbf{22.84}}&\multicolumn{2}{c|}{\textbf{21.28}}&\multicolumn{2}{c||}{\textbf{25.54}}&\multicolumn{2}{c|}{\textbf{25.49}}&\multicolumn{2}{c|}{\textbf{21.45}}&\multicolumn{2}{c}{\textbf{30.04}}\\

&rand 2 $\uparrow$&\multicolumn{2}{c|}{57.50}&\multicolumn{2}{c|}{55.45}&\multicolumn{2}{c||}{59.70}&\multicolumn{2}{c|}{58.75}&\multicolumn{2}{c|}{57.16}&\multicolumn{2}{c||}{60.34}&\multicolumn{2}{c|}{62.36}&\multicolumn{2}{c|}{60.33}&\multicolumn{2}{c||}{63.87}&\multicolumn{2}{c|}{20.52}&\multicolumn{2}{c|}{18.26}&\multicolumn{2}{c||}{23.14}&\multicolumn{2}{c|}{22.09}&\multicolumn{2}{c|}{18.37}&\multicolumn{2}{c}{26.03}\\ \cline{2-32}

&EBF 3 $\uparrow$&\multicolumn{2}{c|}{70.78}&\multicolumn{2}{c|}{68.85}&\multicolumn{2}{c||}{71.28}&\multicolumn{2}{c|}{70.31}&\multicolumn{2}{c|}{69.77}&\multicolumn{2}{c||}{71.49}&\multicolumn{2}{c|}{\textbf{73.64}}&\multicolumn{2}{c|}{\textbf{71.64}}&\multicolumn{2}{c||}{\textbf{74.11}}&\multicolumn{2}{c|}{26.94}&\multicolumn{2}{c|}{24.96}&\multicolumn{2}{c||}{30.18}&\multicolumn{2}{c|}{30.38}&\multicolumn{2}{c|}{24.94}&\multicolumn{2}{c}{35.32}\\

&OBF 3 $\uparrow$&\multicolumn{2}{c|}{\textbf{73.46}}&\multicolumn{2}{c|}{\textbf{73.04}}&\multicolumn{2}{c||}{\textbf{73.72}}&\multicolumn{2}{c|}{\textbf{72.59}}&\multicolumn{2}{c|}{\textbf{72.28}}&\multicolumn{2}{c||}{\textbf{73.08}}&\multicolumn{2}{c|}{72.69}&\multicolumn{2}{c|}{71.62}&\multicolumn{2}{c||}{73.11}&\multicolumn{2}{c|}{\textbf{27.64}}&\multicolumn{2}{c|}{\textbf{26.41}}&\multicolumn{2}{c||}{\textbf{33.20}}&\multicolumn{2}{c|}{\textbf{32.73}}&\multicolumn{2}{c|}{\textbf{26.96}}&\multicolumn{2}{c}{\textbf{38.35}}\\

&rand 3 $\uparrow$&\multicolumn{2}{c|}{68.77}&\multicolumn{2}{c|}{66.95}&\multicolumn{2}{c||}{70.23}&\multicolumn{2}{c|}{68.94}&\multicolumn{2}{c|}{67.84}&\multicolumn{2}{c||}{69.85}&\multicolumn{2}{c|}{72.15}&\multicolumn{2}{c|}{70.66}&\multicolumn{2}{c||}{73.31}&\multicolumn{2}{c|}{25.46}&\multicolumn{2}{c|}{23.37}&\multicolumn{2}{c||}{29.10}&\multicolumn{2}{c|}{28.92}&\multicolumn{2}{c|}{23.52}&\multicolumn{2}{c}{34.39}\\ \cline{2-32}

&EBF 4 $\uparrow$&\multicolumn{2}{c|}{76.86}&\multicolumn{2}{c|}{75.32}&\multicolumn{2}{c||}{76.95}&\multicolumn{2}{c|}{76.41}&\multicolumn{2}{c|}{75.33}&\multicolumn{2}{c||}{77.13}&\multicolumn{2}{c|}{\textbf{78.92}}&\multicolumn{2}{c|}{\textbf{77.30}}&\multicolumn{2}{c||}{\textbf{79.18}}&\multicolumn{2}{c|}{29.76}&\multicolumn{2}{c|}{28.38}&\multicolumn{2}{c||}{33.87}&\multicolumn{2}{c|}{34.53}&\multicolumn{2}{c|}{27.75}&\multicolumn{2}{c}{39.85}\\

&OBF 4 $\uparrow$&\multicolumn{2}{c|}{\textbf{79.12}}&\multicolumn{2}{c|}{\textbf{78.83}}&\multicolumn{2}{c||}{\textbf{79.54}}&\multicolumn{2}{c|}{\textbf{78.21}}&\multicolumn{2}{c|}{\textbf{77.64}}&\multicolumn{2}{c||}{\textbf{78.66}}&\multicolumn{2}{c|}{78.29}&\multicolumn{2}{c|}{77.26}&\multicolumn{2}{c||}{78.75}&\multicolumn{2}{c|}{\textbf{32.55}}&\multicolumn{2}{c|}{\textbf{29.72}}&\multicolumn{2}{c||}{\textbf{37.20}}&\multicolumn{2}{c|}{\textbf{38.74}}&\multicolumn{2}{c|}{\textbf{30.89}}&\multicolumn{2}{c}{\textbf{45.04}}\\

&rand 4 $\uparrow$&\multicolumn{2}{c|}{76.00}&\multicolumn{2}{c|}{75.00}&\multicolumn{2}{c||}{77.16}&\multicolumn{2}{c|}{75.63}&\multicolumn{2}{c|}{74.67}&\multicolumn{2}{c||}{76.48}&\multicolumn{2}{c|}{77.90}&\multicolumn{2}{c|}{76.67}&\multicolumn{2}{c||}{78.87}&\multicolumn{2}{c|}{29.47}&\multicolumn{2}{c|}{27.27}&\multicolumn{2}{c||}{33.95}&\multicolumn{2}{c|}{34.33}&\multicolumn{2}{c|}{27.42}&\multicolumn{2}{c}{40.52}\\

\bottomrule
&DNN&\multicolumn{6}{c}{Densenet100bc}&\multicolumn{6}{c}{Preresnet110}&\multicolumn{6}{c}{Preresnet20}&\multicolumn{6}{c}{VGG19}&\multicolumn{6}{c}{VGG16} \\ \cline{3-32}
\huge{GAN}& \multirow{2}{*}{Similarity} & &&&&&&&&&&&&&&&&&&&&&&&&&&&&& \\
&&\multicolumn{2}{c|}{Median}&\multicolumn{2}{c|}{$1^{st}$-Q}&\multicolumn{2}{c||}{$3^{rd}$-Q}&\multicolumn{2}{c|}{Median}&\multicolumn{2}{c|}{$1^{st}$-Q}&\multicolumn{2}{c||}{$3^{rd}$-Q}&\multicolumn{2}{c|}{Median}&\multicolumn{2}{c|}{$1^{st}$-Q}&\multicolumn{2}{c||}{$3^{rd}$-Q}&\multicolumn{2}{c|}{Median}&\multicolumn{2}{c|}{$1^{st}$-Q}&\multicolumn{2}{c||}{$3^{rd}$-Q}&\multicolumn{2}{c|}{Median}&\multicolumn{2}{c|}{$1^{st}$-Q}&\multicolumn{2}{c}{$3^{rd}$-Q}\\
\midrule
\midrule
\multirow{9}{*}{\rotatebox[origin=c]{90}{Selection Heuristic}}&EBF 2 $\uparrow$&\multicolumn{2}{c|}{55.35}&\multicolumn{2}{c|}{54.62}&\multicolumn{2}{c||}{56.02}&\multicolumn{2}{c|}{60.99}&\multicolumn{2}{c|}{60.69}&\multicolumn{2}{c||}{61.48}&\multicolumn{2}{c|}{\textbf{62.82}}&\multicolumn{2}{c|}{61.57}&\multicolumn{2}{c||}{\textbf{63.70}}&\multicolumn{2}{c|}{\textbf{26.33}}&\multicolumn{2}{c|}{23.98}&\multicolumn{2}{c||}{29.21}&\multicolumn{2}{c|}{28.09}&\multicolumn{2}{c|}{24.08}&\multicolumn{2}{c}{32.14}\\

&OBF 2 $\uparrow$&\multicolumn{2}{c|}{\textbf{58.63}}&\multicolumn{2}{c|}{\textbf{57.27}}&\multicolumn{2}{c||}{\textbf{59.70}}&\multicolumn{2}{c|}{\textbf{62.51}}&\multicolumn{2}{c|}{\textbf{62.24}}&\multicolumn{2}{c||}{\textbf{63.35}}&\multicolumn{2}{c|}{62.70}&\multicolumn{2}{c|}{\textbf{62.32}}&\multicolumn{2}{c||}{63.44}&\multicolumn{2}{c|}{26.31}&\multicolumn{2}{c|}{\textbf{24.78}}&\multicolumn{2}{c||}{\textbf{29.34}}&\multicolumn{2}{c|}{\textbf{28.97}}&\multicolumn{2}{c|}{\textbf{24.14}}&\multicolumn{2}{c}{\textbf{32.16}}\\

&rand 2 $\uparrow$&\multicolumn{2}{c|}{53.52}&\multicolumn{2}{c|}{51.67}&\multicolumn{2}{c||}{54.84}&\multicolumn{2}{c|}{57.12}&\multicolumn{2}{c|}{55.73}&\multicolumn{2}{c||}{58.51}&\multicolumn{2}{c|}{59.62}&\multicolumn{2}{c|}{58.38}&\multicolumn{2}{c||}{60.83}&\multicolumn{2}{c|}{25.41}&\multicolumn{2}{c|}{23.33}&\multicolumn{2}{c||}{28.29}&\multicolumn{2}{c|}{27.74}&\multicolumn{2}{c|}{22.12}&\multicolumn{2}{c}{30.29}\\ \cline{2-32}

&EBF 3 $\uparrow$&\multicolumn{2}{c|}{65.37}&\multicolumn{2}{c|}{64.99}&\multicolumn{2}{c||}{65.95}&\multicolumn{2}{c|}{69.69}&\multicolumn{2}{c|}{69.01}&\multicolumn{2}{c||}{70.31}&\multicolumn{2}{c|}{71.76}&\multicolumn{2}{c|}{71.19}&\multicolumn{2}{c||}{72.66}&\multicolumn{2}{c|}{31.91}&\multicolumn{2}{c|}{29.52}&\multicolumn{2}{c||}{35.91}&\multicolumn{2}{c|}{34.95}&\multicolumn{2}{c|}{29.18}&\multicolumn{2}{c}{38.77}\\

&OBF 3 $\uparrow$&\multicolumn{2}{c|}{\textbf{69.60}}&\multicolumn{2}{c|}{\textbf{68.49}}&\multicolumn{2}{c||}{\textbf{70.06}}&\multicolumn{2}{c|}{\textbf{71.52}}&\multicolumn{2}{c|}{\textbf{71.14}}&\multicolumn{2}{c||}{\textbf{72.17}}&\multicolumn{2}{c|}{\textbf{72.41}}&\multicolumn{2}{c|}{\textbf{71.63}}&\multicolumn{2}{c||}{\textbf{73.12}}&\multicolumn{2}{c|}{\textbf{32.79}}&\multicolumn{2}{c|}{\textbf{30.25}}&\multicolumn{2}{c||}{\textbf{37.22}}&\multicolumn{2}{c|}{\textbf{35.85}}&\multicolumn{2}{c|}{\textbf{30.15}}&\multicolumn{2}{c}{\textbf{40.74}}\\

&rand 3 $\uparrow$&\multicolumn{2}{c|}{64.40}&\multicolumn{2}{c|}{63.18}&\multicolumn{2}{c||}{65.53}&\multicolumn{2}{c|}{67.11}&\multicolumn{2}{c|}{65.98}&\multicolumn{2}{c||}{68.28}&\multicolumn{2}{c|}{69.90}&\multicolumn{2}{c|}{68.89}&\multicolumn{2}{c||}{71.05}&\multicolumn{2}{c|}{31.38}&\multicolumn{2}{c|}{29.13}&\multicolumn{2}{c||}{35.95}&\multicolumn{2}{c|}{34.60}&\multicolumn{2}{c|}{27.90}&\multicolumn{2}{c}{38.14}\\ \cline{2-32}

&EBF 4 $\uparrow$&\multicolumn{2}{c|}{71.41}&\multicolumn{2}{c|}{70.99}&\multicolumn{2}{c||}{72.24}&\multicolumn{2}{c|}{74.72}&\multicolumn{2}{c|}{74.06}&\multicolumn{2}{c||}{75.08}&\multicolumn{2}{c|}{77.19}&\multicolumn{2}{c|}{76.84}&\multicolumn{2}{c||}{77.57}&\multicolumn{2}{c|}{37.14}&\multicolumn{2}{c|}{32.91}&\multicolumn{2}{c||}{41.70}&\multicolumn{2}{c|}{40.03}&\multicolumn{2}{c|}{33.42}&\multicolumn{2}{c}{44.94}\\

&OBF 4 $\uparrow$&\multicolumn{2}{c|}{\textbf{76.01}}&\multicolumn{2}{c|}{\textbf{75.23}}&\multicolumn{2}{c||}{\textbf{76.19}}&\multicolumn{2}{c|}{\textbf{77.31}}&\multicolumn{2}{c|}{\textbf{76.78}}&\multicolumn{2}{c||}{\textbf{77.46}}&\multicolumn{2}{c|}{\textbf{78.22}}&\multicolumn{2}{c|}{\textbf{77.24}}&\multicolumn{2}{c||}{\textbf{78.39}}&\multicolumn{2}{c|}{\textbf{38.95}}&\multicolumn{2}{c|}{\textbf{34.41}}&\multicolumn{2}{c||}{\textbf{43.39}}&\multicolumn{2}{c|}{\textbf{41.93}}&\multicolumn{2}{c|}{\textbf{35.17}}&\multicolumn{2}{c}{\textbf{46.25}}\\

&rand 4 $\uparrow$&\multicolumn{2}{c|}{72.16}&\multicolumn{2}{c|}{71.32}&\multicolumn{2}{c||}{72.90}&\multicolumn{2}{c|}{74.38}&\multicolumn{2}{c|}{73.56}&\multicolumn{2}{c||}{75.32}&\multicolumn{2}{c|}{76.12}&\multicolumn{2}{c|}{75.14}&\multicolumn{2}{c||}{76.97}&\multicolumn{2}{c|}{36.81}&\multicolumn{2}{c|}{33.07}&\multicolumn{2}{c||}{42.43}&\multicolumn{2}{c|}{40.85}&\multicolumn{2}{c|}{32.17}&\multicolumn{2}{c}{44.60}\\

\bottomrule
&DNN&\multicolumn{6}{c}{ROBERTA}&\multicolumn{6}{c}{XLNet}&\multicolumn{6}{c}{BERT}&\multicolumn{6}{c}{DistilBERT}&\multicolumn{6}{c}{Electra} \\ \cline{3-32}
\huge{Text}& \multirow{2}{*}{Similarity} & &&&&&&&&&&&&&&&&&&&&&&&&&&&&& \\
&&\multicolumn{2}{c|}{Median}&\multicolumn{2}{c|}{$1^{st}$-Q}&\multicolumn{2}{c||}{$3^{rd}$-Q}&\multicolumn{2}{c|}{Median}&\multicolumn{2}{c|}{$1^{st}$-Q}&\multicolumn{2}{c||}{$3^{rd}$-Q}&\multicolumn{2}{c|}{Median}&\multicolumn{2}{c|}{$1^{st}$-Q}&\multicolumn{2}{c||}{$3^{rd}$-Q}&\multicolumn{2}{c|}{Median}&\multicolumn{2}{c|}{$1^{st}$-Q}&\multicolumn{2}{c||}{$3^{rd}$-Q}&\multicolumn{2}{c|}{Median}&\multicolumn{2}{c|}{$1^{st}$-Q}&\multicolumn{2}{c}{$3^{rd}$-Q}\\
\midrule
\midrule
\multirow{9}{*}{\rotatebox[origin=c]{90}{Selection Heuristic}}&EBF 2 $\uparrow$&\multicolumn{2}{c|}{\textbf{33.57}}&\multicolumn{2}{c|}{30.98}&\multicolumn{2}{c||}{34.19}&\multicolumn{2}{c|}{35.37}&\multicolumn{2}{c|}{34.26}&\multicolumn{2}{c||}{35.89}&\multicolumn{2}{c|}{\textbf{42.25}}&\multicolumn{2}{c|}{\textbf{33.74}}&\multicolumn{2}{c||}{\textbf{46.96}}&\multicolumn{2}{c|}{\textbf{50.50}}&\multicolumn{2}{c|}{\textbf{48.60}}&\multicolumn{2}{c||}{\textbf{52.14}}&\multicolumn{2}{c|}{37.07}&\multicolumn{2}{c|}{36.49}&\multicolumn{2}{c}{\textbf{41.29}}\\

&OBF 2 $\uparrow$&\multicolumn{2}{c|}{33.06}&\multicolumn{2}{c|}{\textbf{31.72}}&\multicolumn{2}{c||}{\textbf{34.70}}&\multicolumn{2}{c|}{\textbf{37.38}}&\multicolumn{2}{c|}{\textbf{36.84}}&\multicolumn{2}{c||}{\textbf{37.66}}&\multicolumn{2}{c|}{41.90}&\multicolumn{2}{c|}{33.15}&\multicolumn{2}{c||}{46.00}&\multicolumn{2}{c|}{47.65}&\multicolumn{2}{c|}{45.46}&\multicolumn{2}{c||}{49.86}&\multicolumn{2}{c|}{\textbf{37.37}}&\multicolumn{2}{c|}{\textbf{36.77}}&\multicolumn{2}{c}{39.54}\\

&rand 2 $\uparrow$&\multicolumn{2}{c|}{27.42}&\multicolumn{2}{c|}{24.85}&\multicolumn{2}{c||}{31.54}&\multicolumn{2}{c|}{31.60}&\multicolumn{2}{c|}{29.27}&\multicolumn{2}{c||}{33.96}&\multicolumn{2}{c|}{38.84}&\multicolumn{2}{c|}{27.42}&\multicolumn{2}{c||}{45.26}&\multicolumn{2}{c|}{46.56}&\multicolumn{2}{c|}{43.50}&\multicolumn{2}{c||}{49.10}&\multicolumn{2}{c|}{35.04}&\multicolumn{2}{c|}{33.04}&\multicolumn{2}{c}{37.71}\\ \cline{2-32}

&EBF 3 $\uparrow$&\multicolumn{2}{c|}{40.50}&\multicolumn{2}{c|}{39.29}&\multicolumn{2}{c||}{41.93}&\multicolumn{2}{c|}{44.87}&\multicolumn{2}{c|}{44.28}&\multicolumn{2}{c||}{45.66}&\multicolumn{2}{c|}{\textbf{52.65}}&\multicolumn{2}{c|}{40.75}&\multicolumn{2}{c||}{\textbf{57.88}}&\multicolumn{2}{c|}{\textbf{60.23}}&\multicolumn{2}{c|}{\textbf{59.54}}&\multicolumn{2}{c||}{\textbf{61.58}}&\multicolumn{2}{c|}{\textbf{47.28}}&\multicolumn{2}{c|}{\textbf{46.86}}&\multicolumn{2}{c}{\textbf{50.64}}\\

&OBF 3 $\uparrow$&\multicolumn{2}{c|}{\textbf{42.17}}&\multicolumn{2}{c|}{\textbf{40.61}}&\multicolumn{2}{c||}{\textbf{43.94}}&\multicolumn{2}{c|}{\textbf{46.32}}&\multicolumn{2}{c|}{\textbf{44.73}}&\multicolumn{2}{c||}{\textbf{47.47}}&\multicolumn{2}{c|}{50.62}&\multicolumn{2}{c|}{\textbf{42.20}}&\multicolumn{2}{c||}{54.97}&\multicolumn{2}{c|}{56.51}&\multicolumn{2}{c|}{54.37}&\multicolumn{2}{c||}{59.42}&\multicolumn{2}{c|}{46.36}&\multicolumn{2}{c|}{45.24}&\multicolumn{2}{c}{48.90}\\

&rand 3 $\uparrow$&\multicolumn{2}{c|}{35.96}&\multicolumn{2}{c|}{33.23}&\multicolumn{2}{c||}{40.48}&\multicolumn{2}{c|}{40.68}&\multicolumn{2}{c|}{38.33}&\multicolumn{2}{c||}{43.81}&\multicolumn{2}{c|}{49.37}&\multicolumn{2}{c|}{35.96}&\multicolumn{2}{c||}{56.90}&\multicolumn{2}{c|}{59.37}&\multicolumn{2}{c|}{57.11}&\multicolumn{2}{c||}{61.42}&\multicolumn{2}{c|}{46.33}&\multicolumn{2}{c|}{44.48}&\multicolumn{2}{c}{48.98}\\ \cline{2-32}

&EBF 4 $\uparrow$&\multicolumn{2}{c|}{47.50}&\multicolumn{2}{c|}{46.38}&\multicolumn{2}{c||}{49.33}&\multicolumn{2}{c|}{53.17}&\multicolumn{2}{c|}{51.98}&\multicolumn{2}{c||}{53.66}&\multicolumn{2}{c|}{\textbf{61.34}}&\multicolumn{2}{c|}{\textbf{47.86}}&\multicolumn{2}{c||}{\textbf{67.14}}&\multicolumn{2}{c|}{\textbf{68.45}}&\multicolumn{2}{c|}{\textbf{67.57}}&\multicolumn{2}{c||}{\textbf{69.75}}&\multicolumn{2}{c|}{\textbf{55.96}}&\multicolumn{2}{c|}{\textbf{55.40}}&\multicolumn{2}{c}{\textbf{59.18}}\\

&OBF 4 $\uparrow$&\multicolumn{2}{c|}{\textbf{48.62}}&\multicolumn{2}{c|}{\textbf{47.67}}&\multicolumn{2}{c||}{\textbf{50.69}}&\multicolumn{2}{c|}{\textbf{53.50}}&\multicolumn{2}{c|}{\textbf{52.57}}&\multicolumn{2}{c||}{\textbf{53.84}}&\multicolumn{2}{c|}{57.95}&\multicolumn{2}{c|}{48.66}&\multicolumn{2}{c||}{61.78}&\multicolumn{2}{c|}{63.45}&\multicolumn{2}{c|}{61.33}&\multicolumn{2}{c||}{66.20}&\multicolumn{2}{c|}{53.47}&\multicolumn{2}{c|}{51.97}&\multicolumn{2}{c}{57.18}\\

&rand 4 $\uparrow$&\multicolumn{2}{c|}{46.72}&\multicolumn{2}{c|}{43.54}&\multicolumn{2}{c||}{49.01}&\multicolumn{2}{c|}{51.19}&\multicolumn{2}{c|}{47.57}&\multicolumn{2}{c||}{53.36}&\multicolumn{2}{c|}{58.26}&\multicolumn{2}{c|}{46.72}&\multicolumn{2}{c||}{65.48}&\multicolumn{2}{c|}{67.78}&\multicolumn{2}{c|}{65.78}&\multicolumn{2}{c||}{69.40}&\multicolumn{2}{c|}{54.58}&\multicolumn{2}{c|}{52.72}&\multicolumn{2}{c}{56.12}\\

\bottomrule
\end{tabular}}
\label{tab:crit_comp_kmnc}
\end{table}

The dendrogram represents, from the perspective of a DNN model under test, the distance between each test set in terms of fault types. Remark that however, it does not mean that those test sets cover more or less fault types than the other test sets on the DNN model under test. This simply quantifies the distance between test sets. So, for instance, the top dendrogram can be read as: \enquote{from the point of view of the given \textit{RoBERTa} DNN model under test, the reference test sets from \textit{XLNet} (DNN model seed 4) and \textit{XLNet} (DNN model seed 6) have a difference in terms of fault types of 8}. In the \textit{Text}-based generation technique, one can see that reference test sets for a given DNN model types (\eg \textit{XLnet}, \textit{BERT}...) are closer in terms of fault types compared to test sets of other DNN model types. So they would lead to more similar fault types from the point of the DNN model under test \textit{RoBERTa}. In that case, using \textit{EBF} as the selection criteria is the sound choice to improve coverage. On the other hand, in the \textit{GAN}-based generation technique, reference test set clusters are not reliant on the DNN model types. In that case, one should rely on the absolute coverage of each of the test sets for the combination to maximize the coverage, \ie the \textit{OBF} criteria.

To measure the effectiveness of our criteria, for each generation technique, we evaluate both OBF and EBF criteria on each DNN model under test. We do so by taking the $k$-th best DNN models in terms of each criterion following the ranking of the chosen similarity. We varied $k$ from 2 to 4 to see if the coverage increased or stagnated as we added more test sets. To put it in perspective, we also compare the criteria to randomly sampling the same number of reference test sets. To do so, we varied $k$ from 2 to 4 and sampled 30 test sets randomly for each $k$. For the choice of similarity, we picked \textit{PWCCA} and \textit{J$_{Div}$} as those similarities worked reasonably well both in terms of correlations and $Top-1$/$Top-5$ metrics in previous research questions. Results are presented in Table \ref{tab:crit_comp} for the property based on Fault type coverage and in Table \ref{tab:crit_comp_kmnc} for the property based on neuron coverage.

Results show that, for the chosen similarities, with a fixed generation technique and property, there is a criterion better than the other across most DNN models under test. For both neuron and fault-type coverage properties, \textit{EBF} is better for the \textit{Text}-based generation technique and \textit{OBF} is better for \textit{GAN}-based and \textit{Fuzz}-based generation techniques. The reason why one criterion works best on a particular generation technique is as shown with the dendrogram examples: depending on the DNN model under test, certain reference tests have more overlapping faults while belonging to the same DNN model type. For the fault-type property, coverage reaches $\sim$60-80\% of the fault-type of $T_O$ depending on the DNN model type and generation techniques. For the neuron property, coverage reached $\sim$40-80\% but has much more variation across generation techniques and DNN model types. Notably, while more than doubling, coverage reaches only $\sim$40\% on \textit{VGG} DNN model types and $\sim$50\% on \textit{RoBERTa} and \textit{XLNet}. More importantly, one can see that randomly sampling test sets is not a good approach as it rarely leads to more coverage compared to using the proposed heuristics criterion and the similarity. Thus, selecting test sets based on DNN model similarity can bring benefits in increasing the coverage when transferring test sets.

\begin{tcolorbox}[colback=blue!5,colframe=blue!40!black]
\textbf{Findings 4:} At fixed similarity, generation technique and property, one heuristic selection criterion for test sets combination is best overall. In our case, \textit{EBF} works better on \textit{Text}-based generation technique while \textit{OBF} works better on \textit{Fuzz}-based and \textit{GAN}-based generation techniques. Moreover, not only does combining test sets using ranks of similarity help in increasing property coverage, but it does so better than randomly sampling them. Thus, carefully selecting test sets leveraging defined similarity is beneficial for transferring test sets.
\end{tcolorbox}

\subsection{RQ4: \rqfour}\label{sec:rq4}

To demonstrate the effectiveness of \textit{GIST}, we first measure the execution time of each step of the process (\textit{Offline} and \textit{Online}) as well as the range of execution time of each of the test case generation procedures in our experiments. Results are presented in Table \ref{table:time}. We used a computer with the following configuration to compute the execution time of experiments: Ubuntu 22.04, GPU (8 GB VRAM), 16 Go RAM and CPU with 8 cores.

\begin{table}[h!]
\resizebox{\textwidth}{!}{\begin{tabular}{cc||c|cc|cc||c}
\toprule
\multicolumn{2}{c||}{Generation technique} & \multicolumn{5}{c||}{Offline}                                               & \multicolumn{1}{c}{Online} \\
\midrule
\midrule
\multicolumn{1}{c}{Type}&\multicolumn{1}{c||}{}&\multicolumn{1}{c|}{Features} & \multicolumn{2}{c|}{Property (Fault / neuron)} & \multicolumn{2}{c||}{Correlation (Fault / neuron)} & \multicolumn{1}{c}{Selection}\\
\midrule
     Fuzz  & $\sim$ 2,000 - 8,000 & $\sim$ 1,100 & $\sim$ 7,000   & $\sim$ 6    & $\sim$ 1,700  &  $\sim$ 2,000  &  $\sim$ 1-20  \\
     GAN   &  $\sim$ 3,000 - 5,000  &  $\sim$ 1,100 & $\sim$ 8,800   & $\sim$ 6    & $\sim$ 1,800  & $\sim$ 2,000 & $\sim$ 1-20  \\
     Text   &  $\sim$ 12,000 - 14,000  & $\sim$ 4,500 & $\sim$ 6,265  & $\sim$ 10    & $\sim$ 2,000  & $\sim$ 2,500 & $\sim$ 1-30  \\
\bottomrule
\end{tabular}}
\caption{Execution time of each part of \textit{GIST} (\textit{Offline} and \textit{Online}) in seconds. We detail each part of the \textit{Offline} procedure and give the execution time of the test generation technique as a comparison.}
\label{table:time}
\end{table}

We observe that the different phases have different impacts on the overall process: feature extraction is proportional to the DNN model's size and so larger DNN models such as is the case in the text base dataset (\ie RoBERTa, XLNet...) will take a longer time to run. On the contrary, the correlation computation is proportional to the number of comparisons to do and the number of similarities to use. For the similarities, the execution time varies a lot depending on which one is used: representational similarities will take a longer time than functional ones. For instance, functional similarities take on average 1-2 seconds while similarities such as PWCCA can take up to 30 seconds per DNN model under test as highlighted in the \textit{Online} part. For those representational metrics, we used a PyTorch version of the implementation as provided by Ding et al. \cite{Ding21} to facilitate replication. Lastly, the execution time on the property drastically varies: while the fault type property using UMAP/HDBSCAN takes around 7,000 sec, the neuron coverage-based one is much less expensive to compute (around 6 seconds). As such, the property chosen can have a high impact on the execution time of \textit{GIST}. Note that for the fault type, we relied on the same kind of libraries and procedures used by Aghababaeyan et al. \cite{Aghababaeyan23}. So, for instance, the UMAP/HDBSCAN used in our experiments is the original CPU-only implementation\footnote{\url{https://umap-learn.readthedocs.io/en/latest/} and \url{https://hdbscan.readthedocs.io/en/latest/}} to allow for replication, which impacts execution time. However, GPU-based implementations such as the one in the cuML library \cite{Raschka20} could drastically improve execution time.

In the \textit{Generation technique} column, we show the average execution time range for running the generation technique. Execution time varies depending on the DNN model and the technique. For instance, for the fuzzing method applied on a smaller less robust DNN model such as \textit{PreResnet20} will take on average around 2,300 seconds to obtain the 1,000 generated inputs requested. On the contrary, applying on a more robust and bigger DNN model such as \textit{PreResnet110} will take on average around 8,000 seconds. Finally, the selection of relevant test sets in the \textit{Online} part has a very small cost to compute the most similar DNN model to the DNN model under test given the similarity (1-2 seconds for functional similarity and up to 30 seconds for representational). The follow-up inference cost is dependent on the number of inputs being transferred and the DNN model size but will inevitably be smaller than the \textit{Generation} cost which generally requires additional computation (such as gradient backpropagation, multiple inference passes...).

\begin{figure}
    \centering
    
\includegraphics[width=0.8\textwidth]{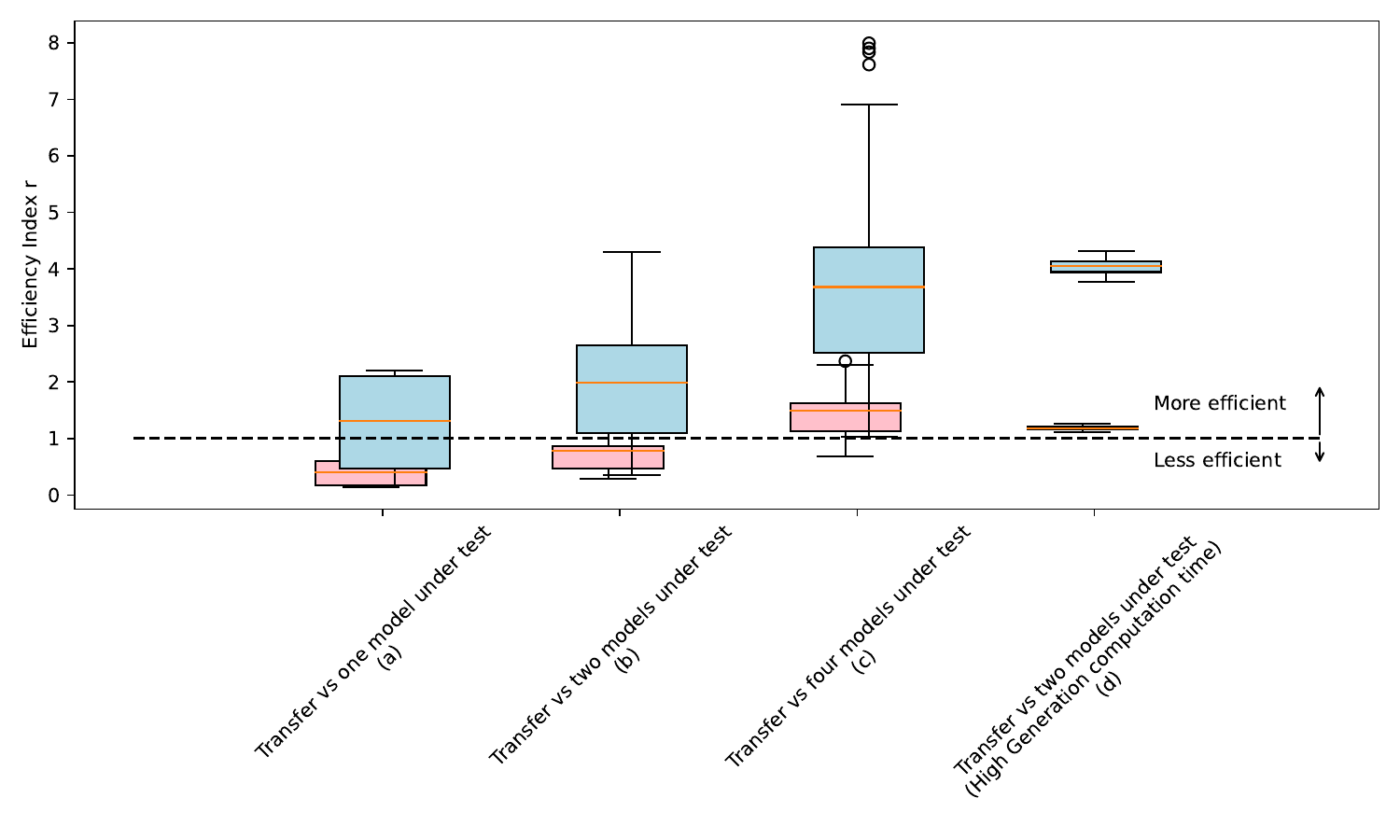}
\caption{Efficiency index $r$ for different comparison scenarios of applying \textit{GIST} vs applying the generation techniques (Fuzzing) on \textit{n} DNN models under test. Red boxplots are results for \textit{fault-type} based coverage, blue boxplots are results for \textit{neuron} based coverage.}
\label{fig:efficiency_fuzz}
\end{figure}

\begin{figure}
    \centering
    
\includegraphics[width=0.8\textwidth]{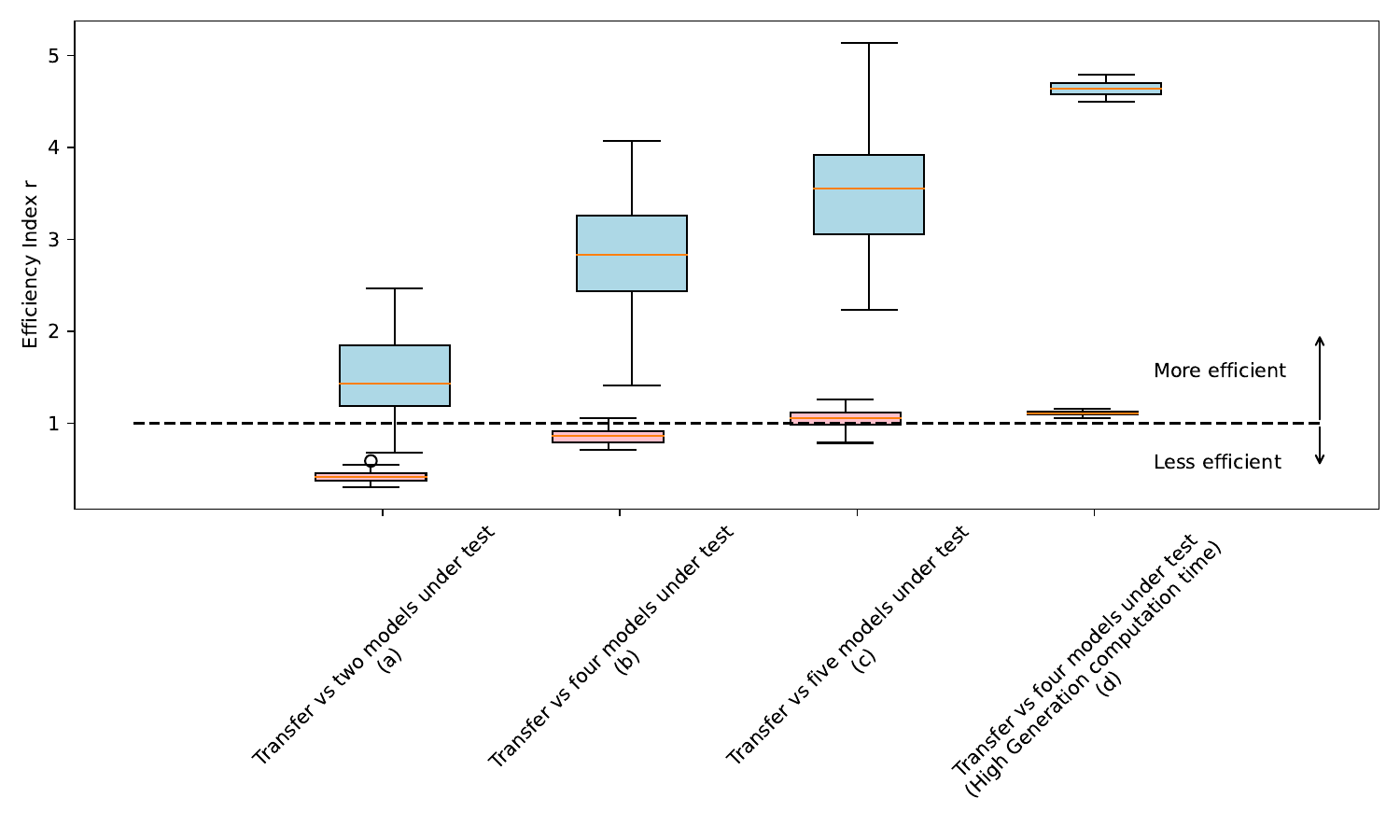}
\caption{Efficiency index $r$ for different comparison scenarios of applying \textit{GIST} vs applying the generation techniques (GAN) on \textit{n} DNN models under test. Red boxplots are results for \textit{fault-type} based coverage, blue boxplots are results for \textit{neuron} based coverage.}
\label{fig:efficiency_gen}
\end{figure}

\begin{figure}
    \centering
    
\includegraphics[width=0.8\textwidth]{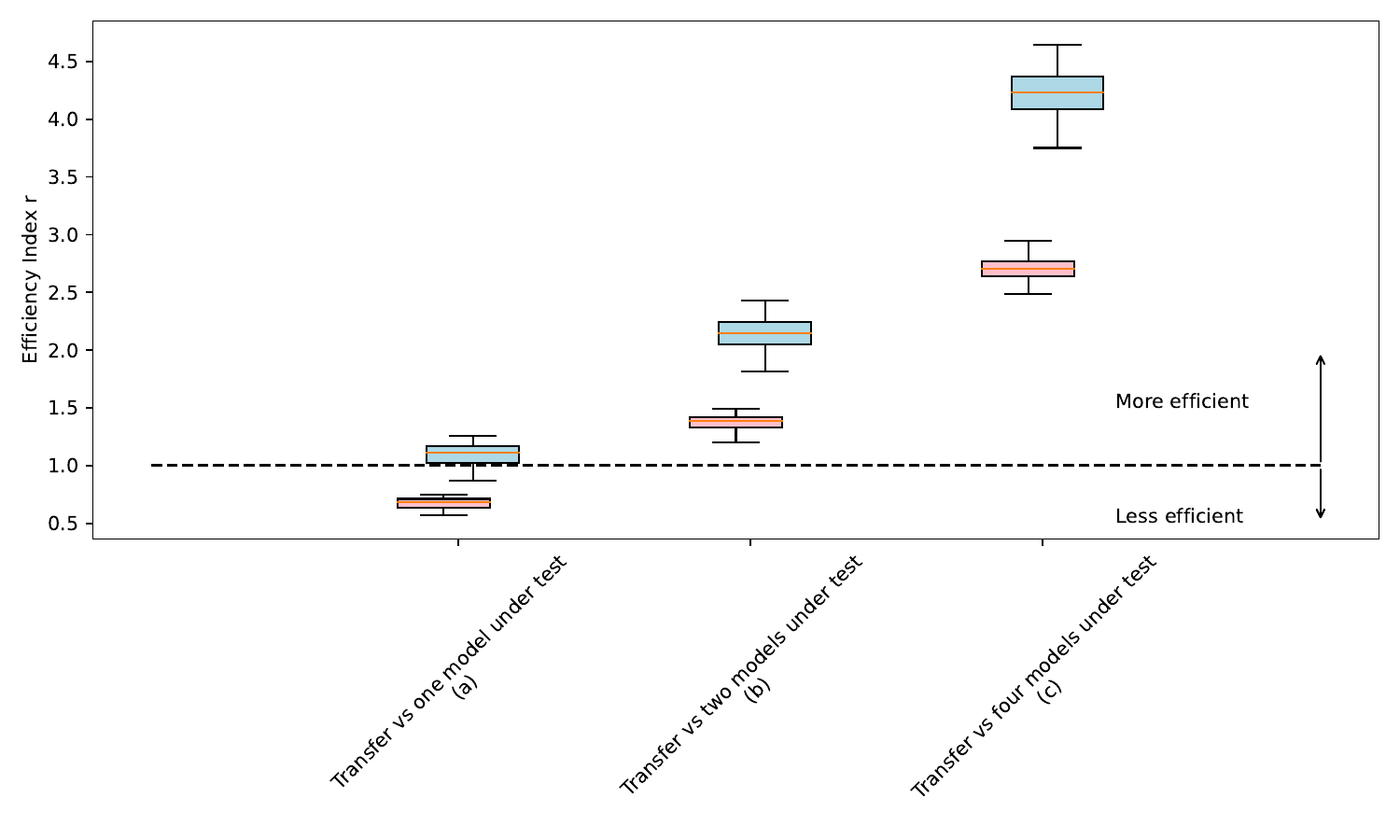}
\caption{Efficiency index $r$ for different comparison scenarios of applying \textit{GIST} vs applying the generation techniques (\textit{Text}-based) on \textit{n} DNN models under test. Red boxplots are results for \textit{fault-type} based coverage, blue boxplots are results for \textit{neuron} based coverage.}
\label{fig:efficiency_text}
\end{figure}

Once we have measured the execution time, we can define an index $r$ to measure the trade-offs of the transferability (\textit{Offline} + \textit{Online}) against reapplying the generation techniques. This $r$ is inspired by existing performance metrics \cite{Zhang23}. We randomly sample $100$ execution times for different DNN model seeds and collect their property coverage obtained in RQ3 (through EBF/OBF criteria, with 4 test sets). We equivalently get the execution time taken by the generation of tests for this DNN model seed and the execution time it would take to apply \textit{GIST} using data from Table \ref{table:time}. We then obtained the index $r$ by dividing the property coverage obtained by applying GIST and a time ratio \textit{t}. This time ratio \textit{t} is itself obtained by dividing the execution time of applying GIST for transferring \textit{n} times and the execution time of applying the test cases generation technique on \textit{n} DNN model under test. The rationale is to show what is gained (in terms of property coverage) compared to what is lost (in terms of execution time) against just applying the traditional generation technique (ratio \textit{Offline}/\textit{Online} and generation technique execution time). For instance, $r = 1$ means that the property coverage was $100\%$ and the \textit{Offline}/\textit{Online} execution took as much time as applying the generation techniques on $n$ DNN models under test. It can also mean that the property coverage was $50\%$ but the \textit{Offline}/\textit{Online} execution took half the time of the generation techniques on $n$ DNN models under test. Generally, a $r$ value above 1 means that the transfer is more efficient and a $r$ value below 1 means that it's less efficient than reapplying the generation techniques on the $n$ DNN models under test. Results are given for the fuzzing generation technique in Figure \ref{fig:efficiency_fuzz}, for the GAN generation technique in Figure \ref{fig:efficiency_gen} and for the text-based generation technique in Figure \ref{fig:efficiency_text}.

For the fuzzing generation technique, in scenario (a) (transferring only on one DNN model), \textit{fault-types} based coverage property transfer is not more efficient than reapplying the generation technique. However, \textit{neuron} based coverage property is already more efficient (median $r \sim 1.3$), even though there is a high variability depending on the DNN models under test. The further we increase the number of generation techniques required (i.e., if we had to apply the technique on multiple DNN models under test), the more efficient the transfer becomes as the \textit{Offline} cost is paid once with only subsequent cheap \textit{Online} execution. For four DNN models under test, both property transfers are more efficient than actually applying the generation techniques from scratch on all of them (scenario (c)). For just two DNN models under test, if the execution time of applying the test cases generation technique is high ($\sim$ 8,000 s such as a \textit{densenet100bc} or \textit{preresnet110} DNN model types), then both property transfers are already more efficient (scenario (d)). We obtain similar results for the text-based generation technique, with the transfer already being more efficient starting from two DNN models under test. The results for the GAN generation technique follow a similar pattern, even though it does require a higher number of DNN models under test to transfer on for the fault types property based transfer to become efficient. The results for the neuron-based property are not affected. We explain the difference because of the added execution time in calculating the fault type clusters due to different hyperparameters. As we explained at the beginning of this section, the execution time for the fault type property is CPU-based only due to the libraries used. And so, even in that case, the transfer shows some benefit that would likely be improved by using more optimized calculations. Note that $r$ accounts both for coverage and execution time. If only execution time is of importance, transferring can already be more efficient with a lower number of generation techniques applied.

\begin{tcolorbox}[colback=blue!5,colframe=blue!40!black]
\textbf{Findings 5:} While the upfront cost of \textit{GIST} (\textit{Offline} part) can be high, it is paid only once. Thus, \textit{GIST} transfer becomes profitable as more test case generation techniques (and so DNN model under test) are required. This is mainly because the subsequent \textit{Online} parts are fast. In particular, it takes two to four reapplication of the generation techniques on DNN models under test for the transfer to be effective in terms of property covered / execution time ratio. In terms of execution time only, transferring can be even faster compared to using a generation technique from scratch on a DNN model under test.
\end{tcolorbox}

\section{Discussion}\label{sec:discussion}

The goal of this paper is the development of a theoretical and practical framework, \textit{GIST}, for the transfer of test inputs compared to regenerating them every time through test case generation techniques. Section \ref{sec:approach} describes the methodology and explains that the success of the envisioned transfer is dependant on both $\mathcal{P}_O$ and $\mathcal{P}$. While we investigated transferability in terms of fault types and neuron coverage, $\mathcal{P}_O$ could be any property we would want to transfer. The first requirement would be to find a proxy $\mathcal{P}$ that only relies on data available during the online phase, \ie DNN models' weights or train data for instance, but not the actual test sets. The second requirement would be to show that there exists a correlation and that some empirical criterion can be leveraged to transfer a test set. Good candidates for $\mathcal{P}$, are the representational/functional similarities metrics we used in our experiments. While other proxies $\mathcal{P}$ could be explored, we believe those similarity metrics should be checked first. 

In practice, \textit{GIST} can be used as shown in Figure \ref{fig:gen_framework}: after investigating the correlation between $\mathcal{P}_O$ and all $\mathcal{P}$ available on the benchmark of available DNN models and test sets, one determines which $\mathcal{P}$ is best. This is the offline phase, similar to what we did in our experiments. At the online phase, a new DNN model under test is trained without having access to its test set through the generation technique \textbf{G}. The user then calculates the similarity with the chosen metric between the DNN model under test and all benchmark DNN models. Then, the test sets being determined the best according to a defined criterion are transferred to the DNN model under test and we have empirical evidence that the chosen test sets are on average better than other available test sets in terms of the metric $\mathcal{P}_O$. That is to say, the chosen test sets can effectively cover similar properties as to what we would have obtained by generating the test set from scratch. There are however some trade-offs within the framework that one must keep in mind and that we discuss in the following:

\subsection{Time complexity}\label{tim_compl}

While in the \textit{Online} phase, not much information is needed on the DNN model under test, there is some extensive computation being made on the benchmark during the \textit{Offline} phase.

As such, when it comes to trade-offs, \textit{GIST} would shine in several cases:
 \begin{itemize}
     \item[1)] If the \textit{Offline} process has already been computed, similar to having access to a DNN model to transfer in transfer learning, then the cost of the transfer procedure is solely reliant on the \textit{Online} part which is drastically inferior to the generation technique used.
     \item[2)] If the \textit{Offline} part hasn't been computed, then it should translate to planning on using the transfer for multiple DNN models. Indeed, while the upfront cost can be similar or more expensive than applying the generation technique once, it is paid only once. As such, any further application of the generation technique on a new DNN model seed or a new architecture will require new computation whereas \textit{GIST} only requires a one-time computation. This is what we demonstrated in RQ4 (Section \ref{sec:rq4}). In particular, note that the \textit{Offline} part can be done anytime as it does not require the DNN model under test contrary to the \textit{Online}/generation process. This is analogous to how one would likely not train a DNN model for transfer learning if there is only one task on which the DNN model will be used.
     \item[3)] Alternatively, still in the case of the \textit{Offline} part not being computed, the transfer procedure could be used (even if it's not the intended usage) on one DNN model when the criteria cost is reduced. Indeed, while the fault-type based property might not be the most cost-effective for one DNN model under test, using the neuron coverage-based (or similar) property drastically reduces the cost and makes the process less costly than applying the generation technique, even for one DNN model under test. As such, \textit{GIST} might still be useful even when the user only has one DNN model, but only for certain properties.
 \end{itemize} 

Finally, while one could certainly run all test sets available without selection, this incurs an inference cost proportional to the number of inputs and the number of test sets. Since we saw in RQ3 that carefully selecting the test sets yields good results which are better than randomly selecting them, leveraging ranking similarity as \textit{GIST} is a valid option. This is especially true when inference cost becomes prohibitive and because \textit{Online} execution time is very low. Moreover, with the aim of potentially improving the DNN model by retraining with the test sets to alleviate faulty inputs, fewer inputs will lead to a lower training time.

\subsection{Number of DNN model seeds and DNN model types}

While the transfer paradigm could be applied in any situation, \textit{GIST} requires a certain number of DNN model seeds/types to validate the proxy. This can limit its usage in situations where such DNN model seeds/types are harder to come by. Thus, aiming to reduce the number of DNN model seeds and/or types needed to select a similarity could help in making \textit{GIST} more practical in those specific DNN model-limited tasks. However, this limitation can be mitigated by choosing smaller/simpler DNN models to build up the available DNN models and apply them to much more complex DNN models. All the publicly available DNN models and test sets of our replication package \cite{ReplicationPackage} can constitute a good starting point to further expand the available DNN models. As an example of this, we trained a \textit{PreResnet1202} which is roughly $\sim$10$\times$ the computational complexity compared to $PreResnet110$ \cite{He16}. Applying the fuzzing generation technique on this DNN model took $40,000$ seconds to generate a test set of $1,000$ inputs following the same technique as we did in our experiments. Leveraging \textit{GIST}, we could select appropriate test sets for transfer. In practice, this would for instance result in more than $70\%$ of both properties being covered for the hypothetical $T_O$ of this DNN model using the $EBF/OBF$ criteria with 4 test sets transferred. As such, we can cover a reasonable amount of the property while saving a large amount of time.

\subsection{Potential improvements venues}

Having shown \textit{GIST}'s setup theoretically and its practical effectiveness, there are still several improvements that can be designed in future works. The main goal of our study was to introduce the idea that transferring test sets could be another avenue for DNN model testing. To show our approach worked, we focused on two properties (neuron and fault-types based), two modalities (image and text) along with test cases generation methods as well as several DNN model types (VGG, DenseNet or ResNet-like for images and BERT-like transformers for text) and showed that \textit{GIST} could be leveraged to transfer test set. Nevertheless, future works could expand on other modalities, DNN models and properties. In particular, as we mentioned at the beginning of this section, this framework is general in the property to be transferred. While we only studied two properties, it would be interesting to compare the same DNN models and test sets on other properties, either using the similarity metrics as a proxy $\mathcal{P}$ or coming up with a new one. Secondly, on the topic of similarity metrics, it could be worthwhile to investigate other types of metrics we did not study here. The ones we chose are good samples of available ones, yet some such as Model Stitching  \cite{Csiszarik21} (which requires additional computation and information from the DNN models compared to the metrics we used) could also show good performance. Moreover, it could even be possible to leverage Deep Metric Learning \cite{Kaya19} to come up with a specific metric for a given property. This however would need to account for the constraint that in the online phase, \ie we do not have access to the test set that would have been generated on the DNN model under test. Then, while \textit{EBF} and \textit{OBF} criteria are simple and straightforward heuristics to leverage similarities to combine test sets and improve coverage, they increase the number of test inputs. When inference computation is inexpensive, or even in our case where individual test sets do not have more than 1,000 individual inputs, this might not be a problem. However, this can become one for larger DNN models or even when using larger test sets. Moreover, combining test sets as a block will inevitably lead to multiple inputs covering the same fault type making them redundant. As such, leveraging test seed selection methods \cite{Zhi23, Xie19, Wei21} on top of the similarity metrics could allow the sampling of an optimal test subset for this property, \ie each input would for instance only cover one fault-type or neuron band. However, just as for the choice of the metric, this needs to take into account the constraint that, during the online phase, we do not have access to the test set of the DNN model under test. Finally, another future work could consider how to include the transfer property as an objective in the generation technique. In that sense, transferred test sets could serve as a basis of easily accessible test inputs and the generation technique could aim to complete this basis with additional test inputs. Those additional test inputs would cover the remaining part of the property not covered by the original basis. This could be a middle point between purely transferring test sets and plainly generating test inputs via the generation technique.

\section{Related Works}\label{sec:related_works}

\textbf{Transferability.} As an idea, transferability has been widely studied in different fields though the way of applying it might differ. In style transfer methods, the style of an image is transferred to another one using the neural style transfer technique \cite{Jing20} while preserving the content of the target image. Transferability even led to a new sub-field of machine learning with the transfer learning paradigm \cite{Pan10}\cite{Farahani21} which aims to improve the learning of a DNN model on a new task by transferring some knowledge from another DNN model. %similar one said DNN model was trained on. 
While the idea is similar, here we focus on transferring test sets based on some desirable properties using similarity metrics as a proxy. Multiple studies have analyzed which adversarial attacks transfer best across different DNN models \cite{Wang22, Tramer17, Inkawhich19} and how they affect the feature space of each DNN model. In our case, we leverage test input generation methods that are not necessarily traditional adversarial attacks. Moreover, where they mainly focus on the attack success rate (\ie the percentage of inputs wrongly classified after transfer), we instead focused on properties of the test sets (fault type/neuron coverage), which from a testing point of view is more desirable. Even traditional software engineering can benefit from transferability, with Lin et al. \cite{Lin19} showing that assertion tests can be transferred between Android apps. The closest idea to our work is a part of the work of Zhi et al. \cite{Zhi23} which showed that their test seed selection strategy could transfer between two DNN models of either different DNN model seeds or different architecture. They however only focused on image dataset and \textit{Fuzzing}-based generation techniques while tracking the number of failures and some coverage criteria based only on neural activation. In our case, we focused on both neuron-based, which is widely used, and fault-types based properties as neuron coverage can be limited \cite{Harel-Canada20, Li19, Aghababaeyan23}. Moreover, their transfer aimed to verify that coverage on a new DNN model under test was still relevant. In a different direction, we aimed to show that transfer can be used to prevent regenerating test cases from scratch by verifying we can reach the same properties. Finally, we conceived a complete framework that considers how different DNN models are related to each other through similarity which allows us to measure existing correlation and pick a relevant test set. Nonetheless, their test seed selection method could be leveraged as we mentioned in the discussion select a subset from the combination of test sets, to obtain an optimal set in terms of the property covered and size.

\textbf{Representational and functional similarity.} Klabunde et al.'s survey \cite{Klabunde23} details existing representational and functional similarities, and we only chose a subset of them in our work among the most popular. Several works have investigated and compared existing representational and functional similarities \cite{Kornblith19, Ding21, Morcos18} or leveraged them to compare or analyze DNN models \cite{Sridhar20, Raghu21, Vulic20, Ollerenshaw22}. For instance, Raghu et al. \cite{Raghu21} used representational similarity to compare Vision Transformers and Convolutional Neural Networks, showing that there exists a difference in features and internal structures. Ding et al. \cite{Ding21} compared \textit{PWCCA}, \textit{CKA} and \textit{Orthogonal Procrustes} on different properties such as specificity to seed choice or sensitivity to principal components. In RQ2, we leveraged the approach that they used to evaluate their metrics, by computing similarity between DNN models on a given layer, measuring a certain property, and making use of Kendall's $\tau$ to assess possible correlation. However, several other similarity metrics are being developed, that we did not experiment with in this study, such as DNN model stitching \cite{Csiszarik21} or adversarial-based \cite{Li21}. Those metrics generally require extra information/learning, which adds further computation, and the metrics we leveraged were enough in our case. Yet, it could be useful depending on the property $\mathcal{P}_O$ used.

\textbf{Coverage in DNN.} Multiple criteria have been proposed to test neural network knowledge which are based on the idea of evaluating neural activation. Neuron Coverage \cite{Pei19}, K-Multisection neuron Coverage (KMNC) \cite{Ma18}, MC/DC based \cite{Sun19} or surprise adequacy based \cite{Kim19} are all criteria leveraging activations of neurons of particular layers in order to evaluate the coverage. They can be combined with some Search-Based Software Testing (SBST) techniques to generate new inputs that will maximize such coverage. Techniques such as DLFuzz \cite{Guo18} or DeepEvolution \cite{Braiek19} use this approach. Activations-based criteria showed some pros and cons but several recent research argued that they could be limited concerning their DNN testing ability \cite{Harel-Canada20, Li19}. As such, criteria relying on alternative coverage testing were devised. For instance, DeepGini \cite{Feng20} leverages an uncertainty-based index to measure the likelihood of misclassification. DeepJanus \cite{Riccio20} uses a set of pairs of inputs that are on the opposite side of the classification boundaries and in different regions of the input space to better probe the DNN model. Finally, Aghababaeyan et al. \cite{Aghababaeyan23} used the idea of clustering fault types as well as a Geometric Diversity (GD) criterion using the determinant of the feature matrix. In our case, the property to be transferred could be any of the above criteria, as \textit{GIST} is agnostic to the property to be used. However, no matter the property to transfer, one would then need to find a proxy $\mathcal{P}$ as we did in our case with representational and functional similarity metrics. To demonstrate transferability, in our case, we used both fault type and neuron coverage. The neuron coverage property is based on the previously mentioned coverage criteria and the fault type property uses a similar approach to Aghababaeyan et al. \cite{Aghababaeyan23}. Note that, contrary to them, our criterion is DNN model-dependent contrary to their DNN model-independent approach since the precise goal was to gauge the extent to which reference test sets would reveal the same fault type as the hypothetical test set on the DNN model under test. As such, we have to rely on the DNN model under test in the clustering strategy. 

\section{Threats to Validity}\label{sec:threats}

\textit{Construct validity:} We provided a general framework with proper constraints and a clear frame: to find $\mathcal{P}_O$ and $\mathcal{P}$ with a clear correlation relation between them. This approach could be generalized to any case in which the above problem can be defined, independently of the $\mathcal{P}_O$ and $\mathcal{P}$. To establish such a relation, we used Kendall-$tau$ as prescribed in the literature. We based our property $\mathcal{P}_O$ on existing metrics: either fault-type or neuron coverage-based. Each of those metrics has its existing limitations. However, the point of the framework is to show that a relation can be leveraged independently of the $\mathcal{P}_O$. We chose those metrics that were used in the literature. For the proxy, we used several similarities defined in the literature. As we can not know in advance what proxy can work for the correlation, several similarity metrics need to be used, which can be a threat to validity. Nonetheless, with the diverse pool we selected, we showed that a correlation exists with both our properties (fault types/neuron coverage). Finally, we showed that it could be leveraged to be of practical use. The comparison against random samples could be a threat to validity. We used $30$ independent random samples as it is common in the literature \cite{Arcuri11}. Therefore, we believe our empirical evaluations validate and motivate \textit{GIST}.

\textit{Internal validity:} There are several internal validity in our study, namely the clustering approach for the fault types, the choice of DNN models, the test inputs data as well as the impact of the seed over the result. To mitigate all of them, we applied several strategies: for the clustering, we followed similar approaches leveraging dimensionality reduction before clustering and verified that clusters represent different fault types following previous papers methodology \cite{Aghababaeyan23, Attaoui23}. For the choice of DNN models, we picked different DNN model types that, while reaching similar accuracy on the dataset, have some differences and resemblances to investigate how our results would be affected. For the test input generation, we used different existing approaches. We made sure to use the same original test inputs for the generation to reduce potential bias when comparing DNN models. Finally, we made sure to use different DNN model seeds, to compare how our results would differ, limiting potential seed bias, and evaluating the robustness of the method against this effect.

\textit{External validity:} To further the generalization of our study, we made sure to experiment using two datasets of different modalities as well as three different test input generation techniques. For each of those techniques, we leveraged five different DNN model types with ten DNN model seeds, each used to generate a test set. In all the cases, we showed that one similarity could be chosen which worked reasonably well across all DNN model seeds of all DNN models under tests. As such, we expect our results to generalize to other modalities, datasets, DNN models and test input generation techniques. Finally, we studied the transferability of two properties (fault types and neuron coverage), which further comforted us that the problem setting of \textit{GIST} is sufficiently general that it could be leveraged in other situations.

\textit{Reliability validity:} To increase the reliability of our study, we detailed our methodology and provided a replication package \cite{ReplicationPackage} with our artifacts to allow for replication of our experiments.

\section{Conclusion}\label{sec:conclusion}

In this paper, we studied the transferability of generated test sets through the development of a framework named \textit{GIST}. Our goal was to promote an alternative way compared to regenerating test cases from scratch for every DNN model under test. After giving a theoretical frame for the work, we studied in detail the transferability of test sets in terms of two properties: fault type and neuron coverage. To do so, \textit{GIST} requires a proxy that relates to the chosen property to transfer. We thus made use of existing works on representational/functional similarities, leveraging existing metrics. We showed that there exists some relations between reference DNN models and DNN models under tests that are visible in the way that the reference-generated test sets cover the properties. As such, not only representational/functional similarities are warranted as a proxy for the chosen property coverage, but we also show that one similarity for each generation method shows positive significant correlations with either fault type or neuron coverage. Then, we show that using simple heuristic criteria leveraging similarity and observations on the property coverage, we can pick relatively good candidates to transfer. Those test sets cover the selected property similarly to that of the test set generated from scratch would. Moreover, they do so better than randomly sampling test sets. Finally, we showed that transferring with \textit{GIST} on \textit{n} DNN models under test was more effective, in terms of the covered property/execution time trade-off, than applying the test case generation techniques on each of them.  We make \textit{GIST} replication package publicly available as a first building block for a possible expansion of a transferability benchmark as we described in the discussion \cite{ReplicationPackage}.

While \textit{GIST} is a first step in improving the transferability of generated test sets, we discuss potential improvement of the method. Besides investigating more DNN models and test generation techniques, some improvements such as reducing test set size through seed selection of the generated test sets or finding a similarity metric that is more tailored for the property to transfer would constitute future works worth investigating. Also interesting would be to study the possibility of transferring between datasets to increase the potential of available DNN models and test sets.

\section*{Acknowledgments}

This work was supported by: Fonds de Recherche du Québec (FRQ), the Canadian Institute for Advanced Research (CIFAR) as well as the DEEL project CRDPJ 537462-18 funded by the Natural Sciences and Engineering Research Council of Canada (NSERC) and the Consortium for Research and Innovation in Aerospace in Québec (CRIAQ), together with its industrial partners Thales Canada inc, Bell Textron Canada Limited, CAE inc and Bombardier inc.

\bibliographystyle{ACM-Reference-Format}
\bibliography{sample-base}
\end{document}